\documentclass[11pt]{article}

\usepackage[final]{acl}

\usepackage[T1]{fontenc}
\usepackage[utf8]{inputenc}
\usepackage{times}
\usepackage{latexsym}
\usepackage{microtype}

\usepackage{amsmath,amssymb}

\usepackage{graphicx}
\usepackage{algorithm}
\usepackage{algorithmic}
\usepackage{booktabs}
\usepackage{array}
\usepackage{multirow}
\usepackage{tabularx}
\usepackage{makecell}
\usepackage{threeparttable}
\usepackage{adjustbox}
\usepackage{xspace}
\usepackage{enumitem}
\usepackage{tikz}
\usetikzlibrary{arrows.meta, positioning, fit, backgrounds}
\usepackage{siunitx}
\usepackage{placeins}
\usepackage{afterpage}
\usepackage{url}
\usepackage{listings}
\usepackage[table]{xcolor}

\newcolumntype{Y}{>{\raggedright\arraybackslash}X}
\newcolumntype{C}{>{\centering\arraybackslash}X}

\lstdefinestyle{jsonstyle}{
    basicstyle=\ttfamily\small,
    breaklines=true,
    frame=single,
    columns=fullflexible,
    showstringspaces=false
}

\definecolor{SectBestDark}{HTML}{4A4A4A}
\definecolor{OverallBestDark}{HTML}{1F2A44}

\definecolor{navy}{HTML}{07143D}
\definecolor{purple}{HTML}{5B2DB5}
\definecolor{blue}{HTML}{0C4FD1}
\definecolor{teal}{HTML}{07808B}
\definecolor{orange}{HTML}{C98A00}
\definecolor{pink}{HTML}{D91B6C}
\definecolor{green}{HTML}{136C12}
\definecolor{red}{HTML}{D71920}

\definecolor{softpurple}{HTML}{F7F2FF}
\definecolor{softblue}{HTML}{F4F8FF}
\definecolor{softteal}{HTML}{F1FBFC}
\definecolor{softorange}{HTML}{FFF9EA}
\definecolor{softpink}{HTML}{FFF2F7}
\definecolor{softgreen}{HTML}{F4FBF1}

\newcommand{\sectbest}[1]{\textit{\textsc{#1}}}
\newcommand{\overallbest}[1]{\underline{\textbf{#1}}}

\title{CondRelBench: A Mixed-Domain Benchmark for Condition-Scoped Relation Understanding}

\author{
Sydney Anuyah$^{1}$ \quad
Bofu Dong$^{2}$ \quad
Sudhasmita Swain$^{3}$ \quad
Sunandan Chakraborty$^{1}$ \\
$^{1}$Luddy School of Informatics, Indiana University Indianapolis \\
$^{2}$Information School, University of Wisconsin--Madison \\
$^{3}$Luddy School of Informatics, Indiana University Bloomington \\
\texttt{sanuyah@iu.edu} \quad
\texttt{bdong55@wisc.edu} \quad
\texttt{sudswain@iu.edu} \quad
\texttt{sunchak@iu.edu}
}

\begin{document}
\maketitle

\begin{abstract}
Relation extraction systems often identify what a relation is, but not the condition under which it holds. This can collapse condition-bound claims into unsupported generalizations, a failure we call {relation overgeneralization}. We introduce CondRelBench, a mixed-domain benchmark for evaluating condition-scoped relation understanding. Each claim is represented as $(E_1, R, E_2 \mid C)$, where $C$ is the condition licensing the relation. Using 3,000 synthetic paragraphs, 200 PubMed full-text papers, and 695 Reddit posts, we construct five probe variants per tuple that preserve, replace, remove, or contradict the condition, or contradict the relation itself. Models classify each probe as \textit{Supported}, \textit{Not Supported}, or \textit{Not Enough Evidence}. We also introduce Overgeneralization Rate (OGR) and Scope Preservation Score (SPS) to measure false support for altered claims and consistency across probe families.
Across prompted LLMs, fine-tuned LLMs, and encoder models, we find that Accuracy and Macro F1 often hide scope failures. Models frequently support the original scoped claim but fail when its condition is removed or altered. Domain-specific encoders achieve the strongest scope preservation, while instruction-tuned LLMs often show higher overgeneralization. Synthetic supervision helps but does not close the gap to real-world PubMed and Reddit performance. Code is available at \url{https://github.com/sydneyanuyah/ConditionalCausality} and Data is available at
\url{huggingface.co/datasets/sanuyah/CondRelBench}

\end{abstract}

\section{Introduction}

Relationship extraction is a fundamental task in natural language processing that enables the automatic identification of semantic connections between entities in text. It plays a critical role across domains such as biomedicine (e.g., drug–disease and protein–protein interactions), finance (e.g., company–acquisition relationships), and social networks (e.g., person–organization affiliations). Accurate relationship extraction supports downstream applications including knowledge graph construction, question answering, recommendation systems, and scientific discovery. In many fields, such as precision medicine and public health, relation extraction systems must distinguish when a relation is supported only under specific textual conditions. Extracting a relation without its validity conditions can turn a scoped claim into an unsupported generalization, which is especially problematic in domains where conclusions depend on population, subtype, exposure, intervention, or clinical context \cite{glass2013causal, ge2026review, cao2026prediction}. For example, a text may state that a gene--disease association holds only for a particular population, molecular subtype, environmental exposure, or clinical setting.  \cite{antoniou2006parity} states that \textit{``... two or more births conferred significantly lower breast cancer risk only among BRCA1 mutation carriers."}. 

This information should not be interpreted as evidence that two or more births lower breast cancer risk for all groups. The condition \textit{``only among BRCA1 mutation carriers''} limits where the relation is valid. Outside that condition, the broader claim is unsupported by the available evidence \cite{parekh2026condition, abdaljalil2026knowing}, and has to be either validated or contradicted. As LLMs are increasingly used to answer questions over scientific and clinical text, contradicted or unsupported answers can misrepresent the evidence available in the source text. Prior work on abstention-aware reasoning shows that LLMs may generate plausible answers even when the available evidence is insufficient \cite{abdaljalil2026knowing}. However, prior work has largely focused on whether a relation or answer is supported by the text, rather than explicitly extracting the conditions that define when the relation is valid. This leaves a gap in modeling scoped relations, where the same relation may be correct under one condition but unsupported or misleading outside that condition. This motivates our first research question: \textbf{RQ1: Do models extract the correct relation but overgeneralize it beyond the stated condition?}


\begin{figure*}[t]
\centering
\resizebox{\textwidth}{!}{%
\begin{tikzpicture}[x=1cm,y=1cm,>=Latex, font=\sffamily]
\path[use as bounding box] (0.12,3.08) rectangle (17.78,10.20);
\fill[white] (0.12,3.08) rectangle (17.78,10.20);

\tikzset{
  panel/.style={rounded corners=4pt, line width=.75pt, fill=white},
  softbox/.style={rounded corners=4pt, line width=.55pt, fill=white},
  flowarrow/.style={-{Stealth[length=1.7mm,width=1.25mm]}, line width=1.65pt, draw=navy, fill=navy, line cap=round},
  itemtext/.style={font=\sffamily\fontsize{4.9}{5.6}\selectfont, text=navy, align=left},
  tinytext/.style={font=\sffamily\fontsize{4.25}{4.85}\selectfont, text=navy, align=left},
  smalltext/.style={font=\sffamily\fontsize{4.65}{5.35}\selectfont, text=navy, align=left},
  titletext/.style={font=\sffamily\fontsize{5.65}{6.4}\selectfont\bfseries, align=center},
  num/.style={circle, text=white, font=\sffamily\bfseries\small, inner sep=2.0pt, minimum size=.38cm}
}

\node[font=\sffamily\fontsize{13.7}{15.0}\selectfont\bfseries, text=navy, align=center, text width=17.3cm] at (8.96,9.82)
  {CondRelBench: Benchmark Construction and Evaluation Pipeline};
\node[font=\sffamily\fontsize{7.1}{8.3}\selectfont, text=navy, align=center, text width=12.5cm] at (8.96,9.35)
  {};

\def\topY{3.22}
\def\topTop{8.90}

\foreach \xa/\xb in {3.26/3.92,6.84/7.50,10.42/11.08,14.00/14.66}{
  \draw[flowarrow] ({\xa+0.06},6.16) -- ({\xb-0.06},6.16);
}

\draw[panel, draw=purple, fill=softpurple] (0.34,\topY) rectangle (3.26,\topTop);
\node[num, fill=purple] at (0.82,8.55) {1};
\node[titletext, text=purple, text width=2.05cm] at (2.02,8.53)
  {Data \& Scope};
\node[smalltext, text=black, align=center, text width=2.42cm] at (1.80,7.56)
  {Relations are annotated with the condition C that licenses them.};

\draw[softbox, draw=purple!55, fill=white] (0.56,6.72) rectangle (3.04,7.18);
\node[itemtext, text width=2.18cm, align=center] at (1.80,6.95) {$(E1, R, E2 \mid C)$};

\draw[softbox, draw=blue!55, fill=white] (0.56,5.55) rectangle (3.04,6.42);
\node[itemtext, anchor=west, text=blue, font=\sffamily\fontsize{5.1}{5.7}\selectfont\bfseries] at (0.68,6.22) {Synthetic $\rightarrow$ Handcrafted \\+Rule--Based};
\node[tinytext, anchor=west, text=black, text width=1.85cm] at (0.78,5.80) {3,000 paragraph rows;\\ 7,765 tuples};

\draw[softbox, draw=green!50, fill=white] (0.56,4.65) rectangle (3.04,5.32);
\node[itemtext, anchor=west, text=green, font=\sffamily\fontsize{5.1}{5.7}\selectfont\bfseries] at (0.78,5.12) {PubMed $\rightarrow$ 200 Papers };
\node[tinytext, anchor=west, text=black, text width=1.95cm] at (0.78,4.85) {681 paragraph rows;\\ 1,397 tuples };

\draw[softbox, draw=orange!65, fill=white] (0.56,3.75) rectangle (3.04,4.42);
\node[itemtext, anchor=west, text=orange!85!red, font=\sffamily\fontsize{5.1}{5.7}\selectfont\bfseries] at (0.78,4.22) {Reddit $\rightarrow$ $4$ Subreddits};
\node[tinytext, anchor=west, text=black, text width=1.95cm] at (0.78,3.95) {695 posts;\\1,586 tuples};

\draw[panel, draw=teal, fill=softteal] (3.92,\topY) rectangle (6.84,\topTop);
\node[num, fill=teal] at (4.40,8.55) {2};
\node[titletext, text=teal, text width=1.80cm] at (5.44,8.50)
  {Relation\\Extraction\\{\fontsize{5.05}{5.7}\selectfont from LLMs}};

\node[smalltext, text=black, align=center, text width=2.35cm] at (5.38,7.35)
  {Models extract scoped relations from text.};

\draw[softbox, draw=teal!55, fill=white] (4.18,6.42) rectangle (6.58,6.98);
\node[itemtext, text=teal, anchor=west, font=\sffamily\fontsize{5.0}{5.7}\selectfont\bfseries] at (4.42,6.73) {Prompted LLMs};
\node[tinytext, text=black, anchor=east] at (6.34,6.49) {};

\draw[softbox, draw=teal!55, fill=white] (4.18,5.62) rectangle (6.58,6.18);
\node[itemtext, text=teal, anchor=west, font=\sffamily\fontsize{5.0}{5.7}\selectfont\bfseries] at (4.42,5.93) {Fine-tuned LLMs};
\node[tinytext, text=black, anchor=east] at (6.34,5.69) {};

\draw[softbox, draw=teal!55, fill=white] (4.18,4.82) rectangle (6.58,5.38);
\node[itemtext, text=teal, anchor=west, font=\sffamily\fontsize{5.0}{5.7}\selectfont\bfseries] at (4.42,5.13) {Other LMs};
\node[tinytext, text=black, anchor=east] at (6.34,4.89) {encoder/encoder-decoder};

\draw[panel, draw=orange, fill=softorange] (7.50,\topY) rectangle (10.42,\topTop);
\node[num, fill=orange] at (7.98,8.55) {3};
\node[titletext, text=orange!85!black, text width=1.85cm] at (9.17,8.50)
  {Probe\\Construction\\{\fontsize{5.05}{5.7}\selectfont 5 probe types}};

\node[smalltext, text=black, align=center, text width=2.32cm] at (8.96,7.40)
  {Five variants test sensitivity to scope and relation changes.};

\newcommand{\simpleProbeRow}[5]{%
  \draw[softbox, draw=#1!62!black, fill=white] (7.75,#2) rectangle (10.18,#3);
  \node[tinytext, text=#1!72!black, font=\sffamily\fontsize{4.75}{5.35}\selectfont\bfseries, anchor=west] at (7.96,{(#2+#3)/2+.08}) {#4};
  \node[tinytext, text=black, anchor=west] at (7.96,{(#2+#3)/2-.17}) {#5};
}

\simpleProbeRow{purple}{6.55}{7.05}{Contradict R}{relation contradicted}{Keep C}{original}
\simpleProbeRow{blue}{5.90}{6.40}{Replace C}{condition swapped}
\simpleProbeRow{orange}{5.25}{5.75}{Remove C}{no scope given}
\simpleProbeRow{green}{4.60}{5.10}{Keep C}{original}
\simpleProbeRow{red}{3.95}{4.45}{Contradict C}{opposite scope}

\draw[panel, draw=pink, fill=softpink] (11.08,\topY) rectangle (14.00,\topTop);
\node[num, fill=pink] at (11.56,8.55) {4};
\node[titletext, text=pink, text width=1.95cm] at (12.73,8.50)
  {Probe Queries\\{\fontsize{5.05}{5.7}\selectfont 3-way label}};

\draw[softbox, draw=orange!55, fill=white] (11.62,6.86) rectangle (13.46,7.56);
\draw[draw=orange!55, fill=white] (12.12,6.86) -- (12.12,6.63) -- (12.38,6.86) -- cycle;
\node[smalltext, text=black, align=center, text width=1.62cm] at (12.54,7.22)
  {Is the probe claim supported by the source context?};

\draw[softbox, draw=green!55, fill=white] (11.42,5.86) rectangle (12.24,6.42);
\node[font=\sffamily\fontsize{5.2}{5.8}\selectfont\bfseries, text=green!55!black, align=center] 
  at (11.83,6.14) {Supported};

\draw[softbox, draw=red!55, fill=white] (12.83,5.86) rectangle (13.70,6.42);
\node[font=\sffamily\fontsize{4.75}{5.25}\selectfont\bfseries, text=red!70!black, align=center, text width=0.72cm] 
  at (13.265, 6.14) {Not\\Supported};

\draw[softbox, draw=blue!45, fill=white] (11.72,5.05) rectangle (13.36,5.60);
\node[font=\sffamily\fontsize{4.65}{5.15}\selectfont\bfseries, text=blue!65!black, align=center, text width=1.42cm] 
  at (12.54,5.325) {Not Enough\\Evidence};

\node[smalltext, text=black, align=center, text width=2.18cm] at (12.54,4.28)
  {Each probe is converted into a natural-language claim using the original context.};

\draw[panel, draw=green, fill=softgreen] (14.66,\topY) rectangle (17.58,\topTop);
\node[num, fill=green] at (15.14,8.55) {5};
\node[titletext, text=green, text width=1.90cm] at (16.42,8.50)
  {Evaluation \&\\Metrics};

\node[smalltext, text=black, align=center, text width=2.35cm] at (16.12,7.45)
  {Evaluate classification quality and whether models preserve scope.};

\draw[softbox, draw=green!48, fill=white] (15.00,4.12) rectangle (17.24,6.70);
\node[font=\sffamily\fontsize{5.0}{5.8}\selectfont\bfseries, text=black, align=left, text width=1.80cm, anchor=north west] at (15.20,6.42)
  {$\bullet$ Accuracy\\[2pt]
   $\bullet$ Macro F1\\[2pt]
   $\bullet$ SPS: scope preservation\\[2pt]
   $\bullet$ OGR: overgeneralization};

\end{tikzpicture}
}
\caption{CondRelBench benchmark construction and evaluation pipeline.}
\label{fig:condrelbench-pipeline}
\end{figure*}
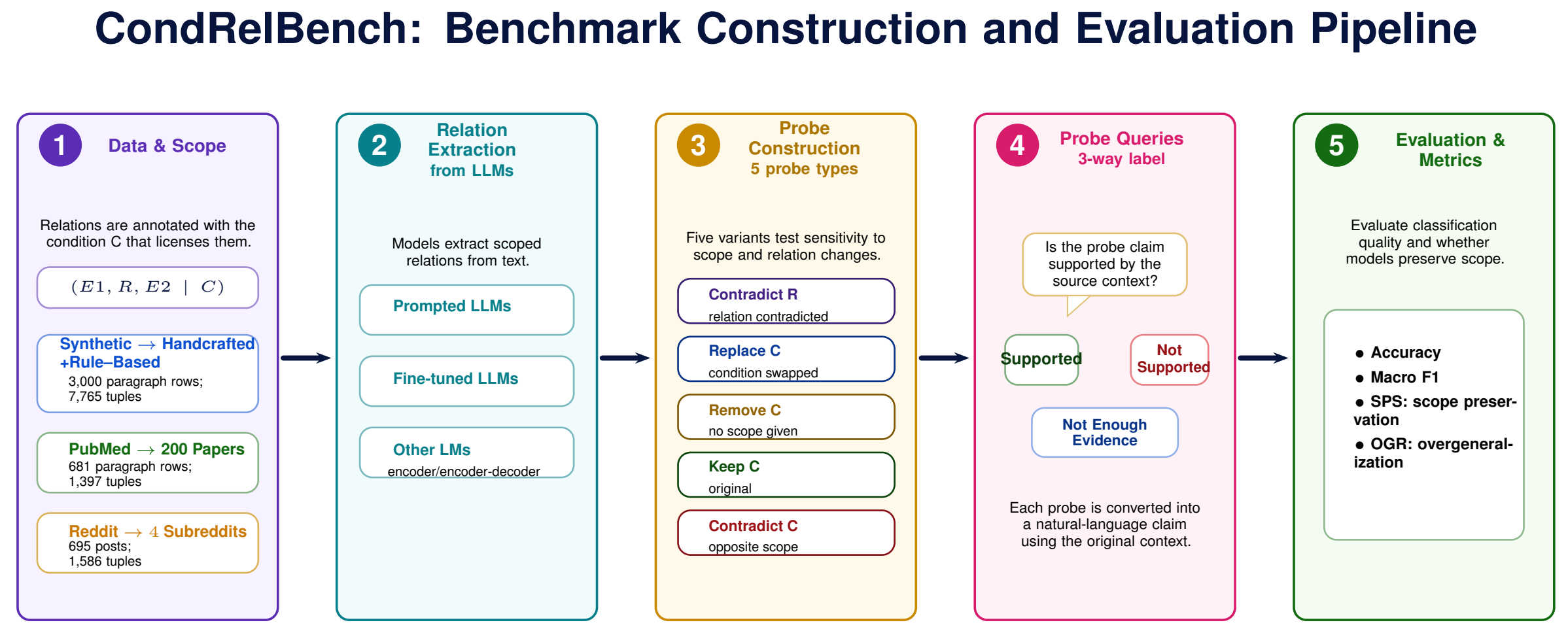

This failure is not directly tested by standard relation extraction (RE) formulations \cite{zhao2024comprehensive, ding2024improving}, which typically represent extracted facts as triples rather than condition-scoped claims \cite{mesinovic2025causal, saini2025relational}. In a conventional triple $(E_1, R, E_2)$, $E_1$ and $E_2$ are relation arguments and $R$ is the relation expressed between them \cite{xu2024context, bou2024two}. Such triples capture the relation but not the textual condition under which the relation is supported \cite{gerner2012biocontext, sosa2023associating}. A model may therefore identify the correct arguments and relation label while still making an unsupported bounded inference \cite{kim2023conditional}. We call this failure mode \textit{relation overgeneralization}: predicting that a relation holds outside the condition under which the text supports it.


We thus represent each scoped relation as $(E_1, R, E_2 \mid C)$, where $C$ is the condition that licenses the relation in the source text. To evaluate relation overgeneralization, we introduce a probe-based framework for condition-aware RE. Given a paragraph that supports a relation only under an explicit or inferable condition, we create queries that preserve the original condition, replace it with a different compatible condition, contradict it, or omit it while holding the argument pair and relation fixed. We also construct relation-perturbation probes that replace the supported relation with an incompatible relation while keeping the arguments and condition fixed. \textbf{RQ2: How robust are models when the supporting condition is preserved, modified, contradicted, or removed?} Models must classify each queried claim as \textit{Supported}, \textit{Not-Supported}, or \textit{Not-Enough-Evidence} with respect to the source paragraph. This leads to our third research question: \textbf{RQ3: Which condition perturbations most often expose relation overgeneralization?}

We evaluate on three data sources: synthetic paragraphs, PubMed full-text papers, and Reddit posts. To test diverse relation and condition structures, we create 3,000 synthetic examples using rule-based templates that vary paragraph structure, argument structure, relation type, condition type, and distractor noise, followed by human verification and correction. For the biomedical setting, linguistic experts review 200 PubMed full-text papers and extract 1,397 unique tuples from 681 condition-bearing paragraphs, which are then converted into controlled probe queries. For the social media setting, we extract 1,586 unique tuples from 695 publicly available Reddit posts drawn from r/ChangeMyView, r/Relationships, r/ExplainLikeImFive, and r/ProstateCancer, using preprocessing and annotation procedures designed to address privacy and ethical considerations \cite{gliniecka2023ethics}.  In summary, this work makes the following contributions: 
\begin{itemize}[noitemsep]
    \item We evaluate condition-aware RE as a scoped claim problem, where relations are valid only under textually supported conditions.
    \item We identify \textit{relation overgeneralization} as a distinct failure mode in which models recover a relation but assign it an unsupported validity scope.
    \item We construct 7,765 Synthetic, 1,397 PubMed, and 1,586 Reddit annotated scoped tuples. After probe validation, the released benchmark contains 37,268 Synthetic, 6,961 PubMed, and 7,883 Reddit probe rows across 7,466, 1,397, and 1,586 retained tuple families, respectively.
    
    \item We introduce a controlled probe-based evaluation framework that tests whether models preserve conditional scope when classifying perturbed claims. The resulting probe set contains {37,268} synthetic queries, {6,961} PubMed queries, and {7,883} Reddit queries.
    \item We evaluate current models on these probes and analyze when they overgeneralize condition-bound relations, especially when the queried condition is omitted, replaced, or contradicted.
\end{itemize} 
Our goal is not to introduce a new model for condition-aware relation extraction, but to evaluate where existing models currently succeed and fail in preserving the conditions under which relations are stated. We discuss the motivation and background in Section~\ref{Sec:Background}, the methodology in Section~\ref{Sec:Method}, and the experiments and results in Section~\ref{Sec:Experiments}. Analysis, discussion, and conclusion are provided in Sections~\ref{sec:Analysis}, \ref{sec:Discussion}, and \ref{Sec:Conclusion} respectively. We acknowledge the limitation of our work, in the Limitation section. Annotation details are provided in Appendix~\ref{sec:annotation_protocol}; extraction and probe evaluation details are provided in Appendices~\ref{app:exp1-evaluation}--\ref{app:exp2-evaluation}; transfer and distributional validation details are provided in Appendices~\ref{app:finetuning_transfer_effects}--\ref{app:data_validation}; probe validation and responsible research disclosures are provided in Appendices~\ref{app:probe_validation}--\ref{app:responsible_research_artifacts}.

\section{Background and Related Work}
\label{Sec:Background}
In this section, we discuss how our task intersects RE, structured representation, textual inference, and causal understanding, which provides the basis for evaluating conditionality in text.


\subsection{Condition-Aware Relation Extraction}
RE includes both sentence-level (SL) and document-level (DL) RE tasks. In SL--RE, prior work assumes that the evidence for a relation is localized within a sentence, while DL--RE allows the relation to be distributed across a broader context \cite{jia2019document,zheng2024survey}. This context can be defined at a paragraph level or span the entire document. In this work, we annotate using a paragraph level span \cite{zheng2024survey} for PubMed and a full-document span for Reddit. 

Previous research has attempted to extend RE beyond the standard triple-based approach to include n-ary relations, event extraction, slot filling, template-based extraction, and hyper-relational or qualifier-aware extraction \cite{luo2022biored, choudhary2024qaevent, schmidt2025grammar, egami2025comparison}. These results form the foundation of our RE tasks, but note that the current triple extraction method still lacks contextual information \cite{saini2025relational, ding2024temporal, panda2024holmes}, as the solutions they provided are targeted towards solving extraction problems, which focus on identifying the relation, argument, event role, qualifier, or attribute \cite{saini2025relational, choudhary2024qaevent, schmidt2025grammar}, rather than reasoning. In our work, we focus on identifying whether LLMs can identify if a condition affects the ground truth relational triple. We do this by extracting the triple and generating probes of comparable length to the paragraphs to test how they can perform inference based on the tuple, helping us to identify to what degree LLMs overgeneralize by testing their inference based on a boundary. Our benchmark is designed to evaluate this inferential use of conditions rather than only their extraction capabilities \cite{arzt2024beyond, gardner2020contrast}.

\subsection{Motivation of This Work}
One key limitation that we have seen from RE literature is the absence of contexts or extra attributes that capture the nuances of text \cite{diazgarcia2025survey, ding2024improving, saini2025relational}. As a subset of natural language inference (NLI), verifying whether models preserve the conditions helps us to trust the factual evidence \cite{dmonte2024claim}. Our work is at the intersection of factual evaluation of LLMs, NLI, and behavioural probing. We evaluate whether models can keep the scope of condition-bearing relational claims when the condition is kept, omitted, contradicted, or replaced. This helps us target the gap between extraction and inference. Models may extract the relation from the text, but still fail to keep the condition that decides what the text actually supports \cite{saini2025relational,modarressi2024consistent,tang2023evaluating}.

 \section{Benchmark Construction}
\label{Sec:Method}
\begin{table*}[t]
\centering
\small
\caption{Conditional tuple extraction across domains and training settings. $S$, $P$, and $R$ denote Synthetic, PubMed, and Reddit. Italic small caps mark the best score within a training block; bold underlining marks the best score within a domain.}
\label{tab:tuple-extraction-comparison}
\resizebox{\textwidth}{!}{
\begin{tabular}{lllrrrrrrrrr}
\toprule
\textbf{Type} & \textbf{Model} & \textbf{Prompt/Setting} &
$\boldsymbol{R_{\mathrm{tuple}}}$ &
$\boldsymbol{F_{1,\mathrm{tuple}}}$ &
$\boldsymbol{R_{E_1}}$ &
$\boldsymbol{R_R}$ &
$\boldsymbol{R_{E_2}}$ &
$\boldsymbol{R_C}$ &
$\boldsymbol{R_{\mathrm{RE}}}$ &
$\boldsymbol{R(\mathrm{RE}\mid C)}$ &
$\boldsymbol{R(C\mid \mathrm{RE})}$ \\
\midrule

\multicolumn{12}{l}{\textit{\textbf{PubMed}}} \\
\midrule
\multicolumn{12}{l}{\hspace{1em}\textbf{Base LLMs}} \\
Base-LLM & Qwen2.5-32B-Instruct & Two Shot + CoT & 0.0456 & 0.0531 & 0.1630 & 0.1753 & 0.1442 & 0.1517 & 0.0713 & 0.3004 & 0.6391 \\
Base-LLM & Kimi-K2 & Two Shot & 0.2327 & 0.1325 & 0.6917 & 0.5598 & 0.5432 & 0.4729 & 0.3383 & \sectbest{0.4921} & 0.6878 \\
Base-LLM & GPT-5.4-mini & Two Shot & \sectbest{0.2487} & 0.1193 & \overallbest{0.7016} & \sectbest{0.5826} & \sectbest{0.5968} & \sectbest{0.5400} & \sectbest{0.3428} & 0.4605 & 0.7254 \\
Base-LLM & Llama-3.3-70B-Instruct & One Shot & 0.2263 & \sectbest{0.1429} & 0.6231 & 0.4890 & 0.4751 & 0.5008 & 0.2804 & 0.4518 & 0.8069 \\
Base-LLM & Phi-4-mini-instruct & Two Shot & 0.0643 & 0.0643 & 0.2890 & 0.2343 & 0.1989 & 0.2097 & 0.1029 & 0.3069 & 0.6250 \\

\addlinespace[2pt]
\multicolumn{12}{l}{\hspace{1em}\textbf{Fine-tuned Models: Synthetic $\rightarrow$ PubMed $(S \rightarrow P)$}} \\
FT-LLM & Llama-3.3-70B-Instruct & One Shot & \sectbest{0.2220} & \sectbest{0.1453} & 0.3197 & 0.3393 & 0.3037 & \sectbest{0.5222} & 0.2877 & 0.4252 & \sectbest{0.7716} \\
FT-LLM & Phi-4-mini-instruct & Two Shot & 0.0568 & 0.0628 & 0.2078 & 0.1829 & 0.1563 & 0.1563 & 0.0906 & 0.3636 & 0.6275 \\
FT-LM & FLAN-T5 & Seq2Seq LM & 0.0558 & 0.0475 & 0.4858 & 0.3416 & 0.2525 & 0.1394 & 0.1137 & 0.4000 & 0.4906 \\
FT-LM & SpanBERT & Encoder LM & 0.0000 & 0.0000 & 0.0311 & 0.0236 & 0.0606 & 0.2231 & 0.0000 & 0.0000 & 0.0000 \\
FT-LLM & Qwen2.5-32B-Instruct & Two Shot + CoT & 0.1776 & 0.1195 & \sectbest{0.5968} & \sectbest{0.5311} & \sectbest{0.4760} & 0.3464 & \sectbest{0.2931} & \sectbest{0.5128} & 0.6061 \\

\addlinespace[2pt]
\multicolumn{12}{l}{\hspace{1em}\textbf{Fine-tuned Models: PubMed $\rightarrow$ PubMed $(P \rightarrow P)$}} \\
FT-LLM & Phi-4-mini-instruct & Two Shot & 0.1279 & 0.0633 & \sectbest{0.6590} & 0.5204 & 0.4654 & 0.5044 & 0.2274 & 0.2535 & 0.5625 \\
FT-LLM & Qwen2.5-32B-Instruct & Two Shot + CoT & 0.1847 & 0.1374 & 0.5968 & 0.5329 & 0.4902 & 0.3446 & 0.3091 & 0.5361 & 0.5977 \\
FT-LLM & Llama-3.3-70B-Instruct & One Shot & 0.1794 & 0.1565 & 0.4441 & 0.3925 & 0.4032 & 0.3606 & 0.2416 & 0.4975 & 0.7426 \\
FT-LM & FLAN-T5-Base & Seq2Seq LM & \overallbest{0.3440} & \overallbest{0.2276} & 0.6418 & 0.6046 & 0.5496 & 0.5053 & \overallbest{0.4060} & \overallbest{0.6807} & \overallbest{0.8472} \\
FT-LM & SpanBERT & Encoder LM & 0.2819 & 0.0147 & 0.6277 & \overallbest{0.6188} & \overallbest{0.6578} & \overallbest{0.6401} & 0.3652 & 0.4404 & 0.7718 \\

\midrule
\multicolumn{12}{l}{\textit{\textbf{Reddit}}} \\
\midrule
\multicolumn{12}{l}{\hspace{1em}\textbf{Base LLMs}} \\
Base-LLM & Qwen2.5-32B-Instruct & Two Shot + CoT & 0.0058 & 0.0072 & 0.0511 & 0.1224 & 0.0664 & 0.0969 & 0.0081 & 0.0602 & \sectbest{0.7222} \\
Base-LLM & Kimi-K2 & Two Shot & 0.0287 & 0.0276 & 0.1395 & 0.1704 & 0.1812 & 0.1650 & 0.0444 & \sectbest{0.1739} & 0.6465 \\
Base-LLM & GPT-5.4-mini & Two Shot & \sectbest{0.0673} & \sectbest{0.0285} & \sectbest{0.3303} & \sectbest{0.3647} & \sectbest{0.3797} & 0.4111 & \sectbest{0.0942} & 0.1636 & 0.7143 \\
Base-LLM & Llama-3.3-70B-Instruct & One Shot & 0.0354 & 0.0234 & 0.2574 & 0.3058 & 0.3498 & \sectbest{0.4242} & 0.0511 & 0.0835 & 0.6930 \\
Base-LLM & Phi-4-mini-instruct & Two Shot & 0.0117 & 0.0123 & 0.1215 & 0.1368 & 0.1404 & 0.1691 & 0.0206 & 0.0690 & 0.5652 \\

\addlinespace[2pt]
\multicolumn{12}{l}{\hspace{1em}\textbf{Fine-tuned Models: Synthetic $\rightarrow$ Reddit $(S \rightarrow R)$}} \\
FT-LLM & Llama-3.3-70B-Instruct & One Shot & \sectbest{0.0463} & \sectbest{0.0416} & 0.0867 & 0.0747 & 0.0762 & \sectbest{0.2167} & \sectbest{0.0658} & 0.2138 & 0.7045 \\
FT-LLM & Phi-4-mini-instruct & Two Shot & 0.0105 & 0.0145 & 0.0433 & 0.0762 & 0.0419 & 0.0568 & 0.0105 & 0.1842 & \overallbest{1.0000} \\
FT-LM & FLAN-T5 & Seq2Seq LM & 0.0000 & 0.0000 & 0.0000 & 0.0000 & 0.0000 & 0.0000 & 0.0000 & 0.0000 & 0.0000 \\
FT-LM & SpanBERT & Encoder LM & 0.0000 & 0.0000 & 0.0000 & 0.0000 & 0.0000 & 0.0000 & 0.0000 & 0.0000 & 0.0000 \\
FT-LLM & Qwen2.5-32B-Instruct & Two Shot + CoT & 0.0269 & 0.0356 & \sectbest{0.0912} & \sectbest{0.1121} & \sectbest{0.1076} & 0.0807 & 0.0404 & \sectbest{0.3333} & 0.6667 \\

\addlinespace[2pt]
\multicolumn{12}{l}{\hspace{1em}\textbf{Fine-tuned Models: Reddit $\rightarrow$ Reddit $(R \rightarrow R)$}} \\
FT-LLM & Phi-4-mini-instruct & Two Shot & 0.0284 & 0.0131 & 0.2242 & 0.3064 & 0.2272 & 0.3573 & 0.0389 & 0.0795 & 0.7308 \\
FT-LLM & Qwen2.5-32B-Instruct & Two Shot + CoT & 0.0105 & 0.0167 & 0.0433 & 0.0628 & 0.0583 & 0.0389 & 0.0269 & 0.2692 & 0.3889 \\
FT-LLM & Llama-3.3-70B-Instruct & One Shot & 0.0030 & 0.0057 & 0.0209 & 0.0164 & 0.0209 & 0.0149 & 0.0045 & 0.2000 & 0.6667 \\
FT-LM & FLAN-T5-Base & Seq2Seq LM & \overallbest{0.1267} & \overallbest{0.0816} & \overallbest{0.4536} & \overallbest{0.4227} & 0.3343 & 0.3387 & \overallbest{0.1767} & \overallbest{0.3739} & 0.7167 \\
FT-LM & SpanBERT & Encoder LM & 0.0722 & 0.0037 & 0.3549 & 0.3932 & \overallbest{0.4389} & \overallbest{0.4919} & 0.0987 & 0.1467 & \sectbest{0.7313} \\

\midrule
\multicolumn{12}{l}{\textit{\textbf{Synthetic}}} \\
\midrule
\multicolumn{12}{l}{\hspace{1em}\textbf{Base LLMs}} \\
Base-LLM & Qwen2.5-32B-Instruct & Two Shot + CoT & 0.8321 & \sectbest{0.8744} & 0.8845 & 0.8576 & 0.8849 & 0.8555 & 0.8542 & \sectbest{0.9726} & \sectbest{0.9741} \\
Base-LLM & Kimi-K2 & Two Shot & \sectbest{0.8713} & 0.8659 & \sectbest{0.9565} & \sectbest{0.9152} & \sectbest{0.9578} & \sectbest{0.9096} & \sectbest{0.9118} & 0.9578 & 0.9556 \\
Base-LLM & GPT-5.4-mini & Two Shot & 0.7045 & 0.7018 & 0.9497 & 0.7417 & 0.9528 & 0.9032 & 0.7365 & 0.7800 & 0.9565 \\
Base-LLM & Llama-3.3-70B-Instruct & One Shot & 0.7992 & 0.8047 & 0.8973 & 0.8487 & 0.8994 & 0.8474 & 0.8419 & 0.9432 & 0.9494 \\
Base-LLM & Phi-4-mini-instruct & Two Shot & 0.7042 & 0.7504 & 0.8014 & 0.7677 & 0.7835 & 0.7481 & 0.7455 & 0.9413 & 0.9445 \\

\addlinespace[2pt]
\multicolumn{12}{l}{\hspace{1em}\textbf{Fine-tuned Models: Synthetic $\rightarrow$ Synthetic $(S \rightarrow S)$}} \\
FT-LLM & Phi-4-mini-instruct & Two Shot & 0.7225 & 0.7589 & 0.7992 & 0.7886 & 0.7801 & 0.7741 & 0.7472 & 0.9334 & 0.9669 \\
FT-LM & FLAN-T5 & Seq2Seq LM & \overallbest{0.9808} & \overallbest{0.9840} & \overallbest{0.9910} & \overallbest{0.9872} & 0.9906 & \overallbest{0.9842} & \overallbest{0.9868} & \overallbest{0.9965} & \overallbest{0.9940} \\
FT-LM & SpanBERT & Encoder LM & 0.9314 & 0.7691 & 0.9830 & 0.9680 & \overallbest{0.9962} & 0.9650 & 0.9540 & 0.9651 & 0.9763 \\
FT-LLM & Llama-3.3-70B-Instruct & One Shot & 0.8713 & 0.8975 & 0.9211 & 0.9207 & 0.8887 & 0.9122 & 0.8824 & 0.9551 & 0.9874 \\
FT-LLM & Qwen2.5-32B-Instruct & One Shot & 0.9037 & 0.9242 & 0.9433 & 0.9344 & 0.9437 & 0.9156 & 0.9318 & 0.9870 & 0.9698 \\

\bottomrule
\end{tabular}
}
\end{table*}

We define a condition $C$ as any explicit textual scope that restricts a relation to a population, setting, state, time, dosage, or comparable context. Each valid claim is represented as a scoped tuple $t=(E_1,R,E_2\mid C)$, where $E_1,E_2$ are arguments, $R$ is the relation, and $C$ is the condition under which $R(E_1,E_2)$ holds. A tuple is retained if all components in $\{E_1,R,E_2,C\}$ are recoverable from the source text; thus, $C$ is part of the claim, not optional context. Conditions that were vague, non-restrictive, or not supported by the source text were not treated as valid scope. A tuple was rejected if any component was missing, incorrect, or wrongly scoped, even when the remaining components were valid. 

\subsection{Data}
The benchmark has three subsets: Synthetic, PubMed, and Reddit. Synthetic contains controlled template-derived examples rewritten to vary condition position, entity structure, relation type, and distractors. PubMed contains paragraph-level biomedical text from prostate cancer genomics articles, providing a focused setting where relations are often bounded by genes, biomarkers and clinical context. Reddit contains post-level informal text from \textit{ChangeMyView}, \textit{Relationships}, \textit{ExplainLikeImFive}, and \textit{ProstateCancer}, selected to introduce topical and stylistic diversity and to stress-test scope preservation in open-ended user-generated discourse, with privacy filtering for identifying details. The final benchmark contains $37{,}268$ Synthetic probes, $6{,}961$ PubMed probes, and $7{,}883$ Reddit probes. Probe validation may remove individual probes from a family. The released data contains 7,466 Synthetic, 1,397 PubMed, and 1,586 Reddit families. Of these, 7,429 Synthetic, 1,374 PubMed, and 1,539 Reddit families retain all five P1–P5 probes. Further dataset details are provided in Appendix~\ref{subsec:annotation_datasets}, probe-retention statistics are provided in Appendix~\ref{app:probe_validation}, and distributional checks are provided in Appendix~\ref{app:data_validation}.


\subsection{Annotation and Validation}
Annotation followed two stages. Stage 1 identified candidate tuples $t=(E_1,R,E_2\mid C)$ by marking the evidence span, entity pair, relation, and condition. For PubMed and Reddit, validation subsets (15\% of the entire dataset) were independently annotated by two annotators and scored with token-level IoU at threshold $\tau=0.5$. Stage 1 showed strong span-level agreement, with Full-Tuple F1 of 0.9164 for PubMed and 0.6593 for Reddit.

Stage 2 reviewed PubMed and Reddit sources from stage 1 with a binary decision $v(t)\in\{\textsc{Yes},\textsc{No}\}$, where $v(t)=\textsc{Yes}$ only if the tuple was valid. Any missing, unsupported, or wrongly scoped component implied rejection. In this stage, we also had 2 annotators. Disagreements were discarded rather than adjudicated, prioritizing high-confidence annotations over exhaustive coverage. The raw agreement scores were 0.9346 for PubMed and 0.9208 for Reddit. Additional details on annotation roles, validation subsets, condition acceptance criteria, disagreement handling, invalid-row rates, and dataset-level annotation statistics are provided in Appendix~\ref{sec:annotation_protocol}. 

For Synthetic, tuples were first rule-generated from templates and then manually rewritten for more natural linguistic variation. Exact-match span annotation between two independent annotators yielded $\mathrm{EM}=1.0$ for both component and full-tuple alignments.

\subsection{Probe Construction}
For each retained tuple $t_i=(E_{1i},R_i,E_{2i}\mid C_i)$, we construct five probes $p_{ik}$: $k=1$ relation contradiction, $k=2$ condition replacement, $k=3$ condition removal, $k=4$ original tuple, and $k=5$ condition contradiction. Their gold labels are
$y_{i1}=y_{i5}=\textsc{NotSupported}$,
$y_{i2}=y_{i3}=\textsc{NotEnoughEvidence}$, and
$y_{i4}=\textsc{Supported}$.
Thus, contradiction and unsupported scope expansion are separated: replacing or removing $C_i$ does not contradict the text, but it exceeds what the evidence licenses. Probes were generated with a fixed JSON-only prompt and manually filtered to remove malformed cases, wrong-component edits, weak alternate conditions, or removals that still preserved $C_i$. 

\subsection{Evaluation}
Extraction is evaluated over component recovery for $E_1,R,E_2,C$ and full-tuple recovery for $(E_1,R,E_2\mid C)$. Probe classification is evaluated with accuracy, macro-F1, label-level F1, and probe-type accuracy. We also report condition-specific errors using the Overgeneralization Rate (OGR), defined over the condition probes $\mathcal{K}=\{2,3,5\}$, along with the Scope Preservation Score (SPS):

$
\mathrm{OGR}
=
\frac{\sum_{k\in\mathcal{K}}\sum_i
\mathbf{1}[\hat{y}_{ik}=\textsc{Supported}]}
{\sum_{k\in\mathcal{K}}N_k}
$

$\mathrm{SPS}=N^{-1}\sum_i\mathbf{1}[\hat{\mathbf{y}}_i=\mathbf{y}_i]$

OGR measures false support for scope-modified probes, while SPS measures whether all probes derived from a base tuple are classified correctly.
For SPS, \(N\) is the number of complete retained tuple families. A family is complete when it contains at least one retained instance of every probe type P1–P5. Incomplete families are excluded from the SPS denominator. Accuracy and Macro F1 are computed over all retained probe rows, while OGR is computed over all retained P2, P3, and P5 rows.

 \section{Experiments and Results}
\label{Sec:Experiments}
\begin{table*}[t]
\centering
\small
\caption{Scope-preservation probe classification across domains and training settings. Lower OGR is better. $S$, $P$, and $R$ denote Synthetic, PubMed, and Reddit. Italic small caps mark the best score within a training block; bold underlining marks the best score within a domain. NB: SPS is computed only over complete P1–P5 probe families.}
\label{tab:exp2-master-comparison}
\resizebox{\textwidth}{!}{%
\begin{tabular}{lllrrrrr||rrrrrrr}
\toprule
\textbf{Type} & \textbf{Model} & \textbf{Prompt/Setting} &
$\boldsymbol{\mathrm{Acc.}}$ &
$\boldsymbol{\mathrm{Macro\ F1}}$ &
$\boldsymbol{\mathrm{Sup.\ F1}}$ &
$\boldsymbol{\mathrm{NS\ F1}}$ &
$\boldsymbol{\mathrm{NEE\ F1}}$ &
$\boldsymbol{\mathrm{P1\ Acc.}}$ &
$\boldsymbol{\mathrm{P2\ Acc.}}$ &
$\boldsymbol{\mathrm{P3\ Acc.}}$ &
$\boldsymbol{\mathrm{P4\ Acc.}}$ &
$\boldsymbol{\mathrm{P5\ Acc.}}$ &
$\boldsymbol{\mathrm{OGR}}$ &
$\boldsymbol{\mathrm{SPS}}$ \\
\midrule

\multicolumn{8}{l||}{\textit{\textbf{PubMed}}} & \multicolumn{7}{l}{} \\
\midrule
\multicolumn{8}{l||}{\hspace{1em}\textbf{Base LLMs}} & \multicolumn{7}{l}{} \\
Base-LLM & Kimi-K2 & Two Shot + CoT & 0.5416 & 0.5310 & 0.5396 & 0.6311 & 0.4222 & 0.5559 & 0.4324 & 0.1453 & \sectbest{0.9294} & 0.6514 & 0.3589 & 0.0306 \\
Base-LLM & Llama-3.3-70B-Instruct & Two Shot + CoT & \sectbest{0.6558} & \sectbest{0.6539} & \sectbest{0.5678} & \sectbest{0.7188} & 0.6749 & \sectbest{0.5695} & 0.5863 & 0.5268 & 0.9199 & 0.6807 & 0.2990 & \sectbest{0.1972} \\
Base-LLM & Phi-4-mini-instruct & Zero Shot & 0.6525 & 0.6440 & 0.5400 & 0.6851 & \sectbest{0.7070} & 0.5086 & \sectbest{0.6535} & \sectbest{0.6779} & 0.7707 & 0.6535 & \sectbest{0.2209} & 0.1689 \\
Base-LLM & Qwen2.5-32B-Instruct & Two Shot + CoT & 0.6381 & 0.6356 & 0.5590 & 0.7078 & 0.6399 & 0.5480 & 0.5412 & 0.4689 & 0.9170 & \sectbest{0.7201} & 0.3061 & 0.1536 \\

\addlinespace[2pt]
\multicolumn{8}{l||}{\hspace{1em}\textbf{Fine-tuned Models: Synthetic $\rightarrow$ PubMed}} & \multicolumn{7}{l}{} \\
FT-LLM & Phi-4-mini-instruct & Zero Shot & 0.7539 & 0.7577 & 0.7712 & 0.7440 & 0.7579 & 0.7901 & 0.5333 & 0.7180 & 0.7627 & 0.9656 & 0.0057 & 0.2620 \\
FT-LLM & Llama-3.3-70B-Instruct & Two Shot + CoT & 0.8887 & 0.8811 & 0.8527 & 0.8489 & 0.9418 & 0.6604 & \sectbest{0.9882} & 0.9764 & \overallbest{0.9831} & 0.8373 & 0.0149 & 0.5755 \\
FT-LLM & Qwen2.5-32B-Instruct & Two Shot + CoT & 0.5069 & 0.4424 & 0.7048 & 0.6154 & 0.0071 & \overallbest{0.9693} & 0.0000 & 0.0071 & 0.5639 & \sectbest{0.9953} & \sectbest{0.0016} & 0.0000 \\
FT-Encoder & BERT & Encoder & 0.8917 & 0.8830 & 0.8419 & 0.8865 & 0.9208 & 0.9176 & 0.8812 & 0.9155 & 0.7868 & 0.9556 & 0.0131 & 0.6521 \\
FT-Encoder & PubMedBERT & Encoder & \sectbest{0.9392} & \sectbest{0.9359} & \sectbest{0.9208} & \sectbest{0.9307} & 0.9561 & 0.9169 & 0.9778 & \sectbest{0.9807} & 0.9090 & 0.9112 & 0.0084 & 0.7666 \\
FT-Encoder & BioBERT & Encoder & 0.8914 & 0.8845 & 0.8513 & 0.8833 & 0.9189 & 0.9248 & 0.8933 & 0.9155 & 0.8042 & 0.9177 & 0.0167 & 0.6722 \\
FT-Encoder & SciBERT & Encoder & 0.9358 & 0.9311 & 0.9088 & 0.9266 & \sectbest{0.9578} & 0.9377 & 0.9549 & 0.9628 & 0.8741 & 0.9485 & 0.0067 & \sectbest{0.7717} \\

\addlinespace[2pt]
\multicolumn{8}{l||}{\hspace{1em}\textbf{Fine-tuned Models: PubMed $\rightarrow$ PubMed}} & \multicolumn{7}{l}{} \\
FT-LLM & Phi-4-mini-instruct & Zero Shot & 0.6698 & 0.6558 & 0.5877 & 0.6223 & 0.7574 & 0.5377 & 0.7193 & 0.7972 & 0.5855 & 0.7075 & 0.0338 & 0.1392 \\
FT-LLM & Llama-3.3-70B-Instruct & Two Shot + CoT & 0.8441 & 0.8714 & 0.8527 & 0.8503 & 0.9112 & 0.6368 & 0.8797 & 0.8396 & 0.9904 & 0.8774 & 0.0134 & 0.4434 \\

FT-LLM & Qwen2.5-32B-Instruct & Two Shot + CoT & 0.7432 & 0.7603 & 0.9215 & 0.7472 & 0.6122 & 0.8915 & 0.4009 & 0.4835 & 0.9470 & \overallbest{0.9976} & \overallbest{0.0008} & 0.1651 \\
FT-Encoder & BERT & Encoder & 0.9209 & 0.9138 & 0.8817 & 0.9236 & 0.9362 & 0.8962 & 0.9670 & 0.9717 & 0.8530 & 0.9151 & 0.0134 & 0.7571 \\
FT-Encoder & PubMedBERT & Encoder & \overallbest{0.9792} & \overallbest{0.9767} & \overallbest{0.9649} & \overallbest{0.9758} & \overallbest{0.9894} & \sectbest{0.9670} & \overallbest{0.9929} & \overallbest{0.9953} & 0.9614 & 0.9788 & 0.0016 & \overallbest{0.9080} \\
FT-Encoder & BioBERT & Encoder & 0.8944 & 0.8829 & 0.8118 & 0.9080 & 0.9290 & 0.8656 & 0.8750 & 0.8986 & 0.9301 & 0.9033 & 0.0786 & 0.7500 \\
FT-Encoder & SciBERT & Encoder & 0.9687 & 0.9654 & 0.9493 & 0.9663 & 0.9806 & 0.9481 & 0.9906 & 0.9788 & 0.9470 & 0.9788 & 0.0039 & 0.8679 \\

\midrule
\multicolumn{8}{l||}{\textit{\textbf{Reddit}}} & \multicolumn{7}{l}{} \\
\midrule
\multicolumn{8}{l||}{\hspace{1em}\textbf{Base LLMs}} & \multicolumn{7}{l}{} \\
Base-LLM & Kimi-K2 & Two Shot + CoT & 0.5448 & 0.5340 & 0.6257 & 0.6029 & 0.3735 & 0.6583 & 0.3594 & 0.1583 & \sectbest{0.8603} & 0.6974 & 0.1976 & 0.0247 \\
Base-LLM & Llama-3.3-70B-Instruct & Two Shot + CoT & \sectbest{0.6999} & \sectbest{0.6964} & 0.6778 & \sectbest{0.7173} & \sectbest{0.6940} & 0.7011 & \sectbest{0.6696} & 0.5088 & 0.8291 & 0.7945 & 0.1171 & \sectbest{0.1988} \\
Base-LLM & Phi-4-mini-instruct & Zero Shot & 0.6492 & 0.6455 & 0.6173 & 0.6614 & 0.6578 & 0.6236 & 0.4994 & \sectbest{0.6356} & 0.7765 & 0.7150 & 0.1412 & 0.1235 \\
Base-LLM & Qwen2.5-32B-Instruct & Two Shot + CoT & 0.6222 & 0.6180 & \sectbest{0.6919} & 0.6552 & 0.5069 & \sectbest{0.7119} & 0.3323 & 0.3739 & 0.8389 & \sectbest{0.8607} & \sectbest{0.1019} & 0.0819 \\

\addlinespace[2pt]
\multicolumn{8}{l||}{\hspace{1em}\textbf{Fine-tuned Models: Synthetic $\rightarrow$ Reddit}} & \multicolumn{7}{l}{} \\
FT-LLM & Phi-4-mini-instruct & Zero Shot & 0.5889 & 0.5944 & 0.6473 & 0.6302 & 0.5058 & 0.7125 & 0.2276 & 0.4615 & 0.6959 & 0.8499 & 0.0679 & 0.0757 \\
FT-LLM & Llama-3.3-70B-Instruct & Two Shot + CoT & \sectbest{0.6757} & \sectbest{0.7612} & \sectbest{0.7652} & \sectbest{0.7179} & \sectbest{0.8004} & 0.5475 & \sectbest{0.7231} & \sectbest{0.7479} & \sectbest{0.7411} & 0.6198 & 0.0138 & \sectbest{0.4483} \\
FT-LLM & Qwen2.5-32B-Instruct & Two Shot + CoT & 0.4803 & 0.4032 & 0.4860 & 0.6050 & 0.1186 & 0.9607 & 0.0475 & 0.0785 & 0.3284 & 0.9835 & \overallbest{0.0000} & 0.0000 \\
FT-Encoder & BERT & Encoder & 0.5722 & 0.4848 & 0.2870 & 0.6526 & 0.5149 & 0.9868 & 0.2989 & 0.4023 & 0.1715 & 0.9899 & 0.0057 & 0.1362 \\
FT-Encoder & PubMedBERT & Encoder & 0.6019 & 0.5361 & 0.3805 & 0.6687 & 0.5590 & 0.9823 & 0.3739 & 0.4073 & 0.2430 & \overallbest{0.9924} & 0.0069 & 0.2144 \\
FT-Encoder & BioBERT & Encoder & 0.6041 & 0.5307 & 0.3432 & 0.6695 & 0.5795 & 0.9792 & 0.3903 & 0.4439 & 0.2118 & 0.9836 & 0.0055 & 0.1879 \\
FT-Encoder & SciBERT & Encoder & 0.6411 & 0.5911 & 0.4585 & 0.6914 & 0.6235 & \overallbest{0.9874} & 0.4199 & 0.4931 & 0.3054 & 0.9899 & 0.0061 & 0.2560 \\

\addlinespace[2pt]
\multicolumn{8}{l||}{\hspace{1em}\textbf{Fine-tuned Models: Reddit $\rightarrow$ Reddit}} & \multicolumn{7}{l}{} \\
FT-LLM & Phi-4-mini-instruct & Zero Shot & \overallbest{0.9063} & \overallbest{0.9145} & \overallbest{0.9180} & \overallbest{0.8951} & \overallbest{0.9305} & 0.8450 & \overallbest{0.8884} & \overallbest{0.9504} & \overallbest{0.9305} & 0.9174 & 0.0048 & \overallbest{0.6488} \\
FT-LLM & Llama-3.3-70B-Instruct & Two Shot + CoT & 0.5960 & 0.6770 & 0.6780 & 0.6491 & 0.7038 & 0.5021 & 0.6033 & 0.5331 & 0.7537 & 0.5909 & 0.0751 & 0.2541 \\
FT-LLM & Qwen2.5-32B-Instruct & Zero Shot & 0.5508 & 0.5275 & 0.6328 & 0.6342 & 0.3155 & 0.9070 & 0.1921 & 0.1921 & 0.4989 & 0.9628 & 0.0048 & 0.0248 \\
FT-Encoder & BERT & Encoder & 0.6777 & 0.6647 & 0.6205 & 0.6947 & 0.6789 & 0.8244 & 0.6281 & 0.6302 & 0.4716 & 0.8306 & 0.0069 & 0.3864 \\
FT-Encoder & PubMedBERT & Encoder & 0.6682 & 0.6451 & 0.5882 & 0.7021 & 0.6449 & \sectbest{0.9607} & 0.4897 & 0.4917 & 0.4316 & 0.9628 & 0.0048 & 0.3781 \\
FT-Encoder & BioBERT & Encoder & 0.6632 & 0.6366 & 0.5710 & 0.7008 & 0.6379 & \sectbest{0.9607} & 0.4731 & 0.4876 & 0.4147 & \sectbest{0.9752} & \sectbest{0.0041} & 0.3533 \\
FT-Encoder & SciBERT & Encoder & 0.7051 & 0.6941 & 0.6551 & 0.7185 & 0.7088 & 0.8574 & 0.6426 & 0.6446 & 0.5179 & 0.8595 & 0.0103 & 0.4360 \\

\midrule
\multicolumn{8}{l||}{\textit{\textbf{Synthetic-Test}}} & \multicolumn{7}{l}{} \\
\midrule
\multicolumn{8}{l||}{\hspace{1em}\textbf{Base LLMs}} & \multicolumn{7}{l}{} \\
Base-LLM & Kimi-K2 & Two Shot + CoT & 0.5481 & 0.5427 & 0.5928 & 0.5820 & 0.4532 & 0.7621 & 0.5354 & 0.1154 & 0.8839 & 0.4445 & 0.2924 & 0.0227 \\
Base-LLM & Llama-3.3-70B-Instruct & Two Shot + CoT & \sectbest{0.7417} & \sectbest{0.7374} & \sectbest{0.7142} & 0.7134 & \sectbest{0.7847} & 0.7693 & \sectbest{0.8851} & 0.6606 & \sectbest{0.8969} & 0.4971 & 0.1366 & \sectbest{0.2365} \\
Base-LLM & Phi-4-mini-instruct & Zero Shot & 0.5930 & 0.5353 & 0.3253 & 0.6651 & 0.6155 & 0.6927 & 0.7327 & \sectbest{0.6739} & 0.2366 & 0.6285 & \sectbest{0.0383} & 0.0486 \\
Base-LLM & Qwen2.5-32B-Instruct & Two Shot + CoT & 0.6890 & 0.6791 & 0.6543 & \sectbest{0.7393} & 0.6438 & \sectbest{0.7915} & 0.7029 & 0.3715 & 0.7076 & \sectbest{0.8717} & 0.0928 & 0.1216 \\

\addlinespace[2pt]
\multicolumn{8}{l||}{\hspace{1em}\textbf{Fine-tuned Models: Synthetic $\rightarrow$ Synthetic}} & \multicolumn{7}{l}{} \\
FT-LLM & Phi-4-mini-instruct & Zero Shot & 0.8948 & 0.8959 & 0.8962 & 0.8751 & 0.9165 & 0.8445 & 0.7737 & 0.9225 & 0.9402 & 0.9933 & 0.0009 & 0.5728 \\
FT-LLM & Llama-3.3-70B-Instruct & Two Shot + CoT & 0.9261 & 0.9212 & 0.9007 & 0.9013 & 0.9616 & 0.8004 & 0.9973 & 0.9924 & \overallbest{0.9960} & 0.8445 & 0.0068 & 0.6971 \\
FT-LLM & Qwen2.5-32B-Instruct & Two Shot + CoT & 0.6037 & 0.5833 & 0.9105 & 0.6543 & 0.1851 & 0.8748 & 0.0882 & 0.1158 & 0.9420 & \overallbest{0.9987} & 0.0006 & 0.0089 \\
FT-Encoder & BERT & Encoder & 0.9907 & 0.9892 & 0.9817 & 0.9887 & 0.9973 & 0.9835 & 0.9942 & 0.9982 & 0.9804 & 0.9973 & 0.0003 & 0.9546 \\
FT-Encoder & PubMedBERT & Encoder & 0.9933 & 0.9924 & 0.9876 & 0.9916 & 0.9979 & 0.9804 & 0.9964 & 0.9991 & 0.9951 & 0.9955 & 0.0003 & 0.9666 \\
FT-Encoder & BioBERT & Encoder & \overallbest{0.9949} & \overallbest{0.9942} & \overallbest{0.9909} & \overallbest{0.9936} & \overallbest{0.9982} & \overallbest{0.9871} & \overallbest{0.9978} & \overallbest{0.9996} & 0.9942 & 0.9960 & \overallbest{0.0001} & \overallbest{0.9746} \\
FT-Encoder & SciBERT & Encoder & 0.9932 & 0.9923 & 0.9876 & 0.9916 & 0.9977 & 0.9840 & 0.9955 & 0.9982 & 0.9929 & 0.9955 & 0.0007 & 0.9661 \\

\bottomrule
\end{tabular}%
}
\end{table*}

We evaluate condition-scoped relation understanding with two tasks: (1) How well can a model extract the full tuple $(E_1, R, E_2 \mid C)$ from data. (2) Does a model preserve the scope when the relation or condition is perturbed in a probe claim. We also evaluate the cross-domain transfer for each task.

\subsection{Experiment 1: Conditional Tuple Extraction}
\label{sec:tuple_extraction_results}

Given a source text, models output condition-scoped tuples of the form $(E_1, R, E_2 \mid C)$. This task tests whether models can recover not only the entity pair and relation, but also the condition that licenses the relation. Table~\ref{tab:tuple-extraction-comparison} reports the partial-match results across Synthetic, PubMed, and Reddit, while additional Synthetic exact-match results are reported in Appendix~\ref{app:summary-synthetic-extraction}, and more details on Experiment 1 extraction results are reported in Appendix~\ref{app:exp1-extraction-results}.


\subsection{Experiment 2: Scope-Preservation Probe Classification}
\label{sec:probe_classification_results}

Given a source text and a probe claim, models predict whether the claim is \textit{Supported}, \textit{Not Supported}, or \textit{Not Enough Evidence}. The probes hold the base relation fixed while preserving, replacing, removing, or contradicting the condition, and also include relation-contradiction cases. This task isolates scope preservation from tuple discovery: the model must decide whether the queried claim remains licensed by the evidence.

\subsection{Model Categories}
The tested models can be broadly classified into (1) Base LLMs, (2) Finetuned LLMs and (3) Fine-tuned smaller Language Models(LMs). We used a variety of models to capture the different sizes and weights. The full list of the models are in the Model column of Table \ref{tab:tuple-extraction-comparison} and Table~\ref{tab:exp2-master-comparison}.


\subsection{Training Notation.}
For fine-tuned rows, $X\rightarrow Y$ denotes training on source $X$ and evaluation on target $Y$, where $S$, $P$, and $R$ denote Synthetic, PubMed, and Reddit. In-domain rows such as $P\rightarrow P$ and $R\rightarrow R$ estimate the effect of target-domain supervision. Transfer rows such as $S\rightarrow P$ and $S\rightarrow R$ test whether supervision from controlled synthetic examples generalizes to natural biomedical and social-media text. 

\subsection{Metrics}
For tuple extraction, we report full-tuple recall and F1, component recall for $E_1$, $R$, $E_2$, and $C$, relation-only recall over $(E_1,R,E_2)$, and conditional recovery measures $R(\mathrm{RE}\mid C)$ and $R(C\mid \mathrm{RE})$. For probe classification, we report accuracy, Macro F1, label-level F1, probe-type accuracy, Overgeneralization Rate (OGR), and Scope Preservation Score (SPS).

\section{Analysis}
\label{sec:Analysis}

\subsection{Experiment 1 Analysis}
Table~\ref{tab:tuple-extraction-comparison} shows a clear-cut difference between synthetic and real-world data. On Synthetic, prompted LLMs can extract much of the scoped structure: Qwen2.5-32B-Instruct reaches $F_{1,\mathrm{tuple}}=0.8744$, and Kimi-K2 obtains the highest base-model tuple recall. Fine-tuning further improves these results with FLAN-T5 reaching $F_{1,\mathrm{tuple}}=0.9840$ under $S\rightarrow S$.

The real-world subsets are substantially harder. On PubMed, the best prompted model reaches only $F_{1,\mathrm{tuple}}=0.1429$, while in-domain fine-tuned FLAN-T5 again dominates this task with $F_{1,\mathrm{tuple}}=0.2276$. Reddit proved even more difficult, as all LLMs have low tuple F1, and FLAN-T5 still performed best with an in-domain fine-tuned approach, reaching $0.1267$ on $F_1$. Under relation-only recall, in-domain fine-tuned FLAN-T5 reaches up to 0.4060 and 0.1767 on PubMed and Reddit. 

We also noted that the synthetic data did not generalize well to real-world data. The results are not definitive; however, we generally notice an improvement in the LLMs compared to using the prompt-only technique. For PubMed, $S\rightarrow P$ improves some component recalls but does not match $P\rightarrow P$ fine-tuning. For Reddit, $S\rightarrow R$ remains weak across model families. 
\subsection{Experiment 2 Analysis}
Table~\ref{tab:exp2-master-comparison} shows that probe classification is easier than open tuple extraction, but this does not make it reliable. Among prompted LLMs, Llama-3.3-70B-Instruct is the strongest overall base model, with the best Macro F1 on PubMed, Reddit, and Synthetic-Test. However, its SPS remains low across domains, showing that good aggregate classification does not imply consistent behavior across the full probe family for a scoped tuple.

Fine-tuning substantially improves probe classification, especially for encoders. On PubMed, in-domain PubMedBERT obtains Macro F1 $0.9767$ and SPS $0.9080$, while SciBERT and other biomedical encoders also perform strongly. On Synthetic-Test, fine-tuned encoders approach ceiling performance, with BioBERT reaching Macro F1 $0.9942$ and SPS $0.9746$. Reddit remains harder: the best in-domain fine-tuned result is Phi-4-mini-instruct with Macro F1 $0.9145$ and SPS $0.6488$, but smaller encoders still lag behind, reflecting the difficulty of informal and discourse-dependent conditions. The transfer rows again show that synthetic supervision does not fully solve real-world scope preservation. $S\rightarrow P$ and $S\rightarrow R$ models often reduce OGR, but low OGR alone can be misleading when models over-predict non-supported labels or fail to recover the complete probe pattern.

\section{Discussion}
\label{sec:Discussion}

\subsection{Scope Preservation Beyond Relation Recognition}
\label{sec:scope_preservation_discussion}

We see that condition-aware relation understanding is not captured by ordinary relation recognition alone. Models can often recognize the original scoped claim, but fail when the condition is removed or altered. This is most visible in the gap between P4, which preserves the condition and P3, which removes the condition. For example, Kimi-K2 reaches P4 accuracy $0.9294$ on PubMed, but only $0.1453$ on P3, with OGR $0.3589$ and SPS $0.0306$, indicating that the model can confirm the fully scoped claim, but often fails to reject the broader unscoped version.

These failures motivate reporting OGR and SPS alongside Accuracy and Macro F1. OGR measures whether a model over-supports altered or unscoped claims, while SPS measures whether the full probe family derived from the same tuple is classified consistently. Low OGR alone is not sufficient, since a model can avoid overgeneralization by under-predicting the \textit{Supported} label. Similarly, high P4 accuracy is not sufficient, since a model may support the original claim while still failing on condition replacement or removal probes. A reliable condition-aware model should combine high Macro F1, high P4 accuracy, low OGR, and high SPS.

The synthetic-to-real gap further shows that controlled data is useful for diagnosis but does not replace real-domain evaluation. Encoder models reach near-ceiling Macro F1 on Synthetic-Test, but drop on PubMed and especially Reddit. For instance, SciBERT drops from Macro F1 $0.9923$ on Synthetic-Test to $0.9311$ on PubMed and $0.5911$ on Reddit. This suggests that Reddit's informal and discourse-dependent conditions remain difficult even when models perform well on controlled perturbations.

\subsection{Effect of Tuple Supervision}
\label{sec:tuple_supervision_ablation}

We also compare probes-only fine-tuning with probes-plus-tuples fine-tuning. The ablation shows that adding tuple supervision does not uniformly improve scope-sensitive classification. For Phi-4-mini-instruct, it substantially reduces the performance across all three domains: Macro F1 drops by $0.5513$ on PubMed, $0.3812$ on Reddit, and $0.6879$ on Synthetic-Test. In contrast, encoder models show much smaller changes, with near-zero differences on Synthetic-Test and mixed effects on Reddit.

\begin{table}[t]
\centering
\tiny
\caption{Ablation results comparing probes-plus-tuples (P+T) with probes-only (P) fine-tuning for scope-sensitive classification. $\Delta$ is $F1_{P+T}-F1_P$; negative values indicate that adding tuple supervision lowers Macro F1.}
\label{tab:macro-f1-ablation-pt-vs-p}
\begin{tabular}{llrrr}
\toprule
\textbf{Dataset} & \textbf{Model} &
$\boldsymbol{\mathrm{F1}_{P+T}}$ &
$\boldsymbol{\mathrm{F1}_{P}}$ &
$\boldsymbol{\Delta_{\mathrm{Macro\ F1}}}$ \\
\midrule
\multicolumn{5}{l}{\textit{\textbf{Fine-tuned Large Language Models}}} \\
\midrule
PubMed & Phi & 0.2064 & 0.7577 & -0.5513 \\
Reddit & Phi & 0.2132 & 0.5944 & -0.3812 \\
Synthetic & Phi & 0.2080 & 0.8959 & -0.6879 \\

\midrule
\multicolumn{5}{l}{\textit{\textbf{Other Fine-tuned Language Models}}} \\
\midrule
PubMed & BERT & 0.8440 & 0.8830 & -0.0390 \\
PubMed & PubMedBERT & 0.9175 & 0.9359 & -0.0184 \\
PubMed & BioBERT & 0.8506 & 0.8845 & -0.0339 \\
PubMed & SciBERT & 0.9134 & 0.9311 & -0.0177 \\

\midrule
Reddit & BERT & 0.5025 & 0.4848 & +0.0177 \\
Reddit & PubMedBERT & 0.5019 & 0.5361 & -0.0342 \\
Reddit & BioBERT & 0.5134 & 0.5307 & -0.0173 \\
Reddit & SciBERT & 0.5603 & 0.5911 & -0.0308 \\

\midrule
Synthetic & BERT & 0.9897 & 0.9892 & +0.0005 \\
Synthetic & PubMedBERT & 0.9925 & 0.9924 & +0.0001 \\
Synthetic & BioBERT & 0.9934 & 0.9942 & -0.0008 \\
Synthetic & SciBERT & 0.9927 & 0.9923 & +0.0004 \\
\bottomrule
\end{tabular}
\end{table}

This supports treating tuple extraction and scope-sensitive verification as connected but distinct objectives. Tuple supervision may help represent relation structure, but the ablation does not show that it improves the downstream inference of whether a condition-altered claim is \textit{Supported}, \textit{Not Supported}, or \textit{Not Enough Evidence}. A more suitable design may therefore use a two-stage approach: first extract the scoped tuple $(E_1, R, E_2 \mid C)$, then optimize a separate verification module for condition-sensitive probe classification.

\section{Conclusion}
\label{Sec:Conclusion}

This work shows that recognizing a relation is not the same as understanding when that relation holds. Standard relation extraction represents claims as triples, but many real-world claims are valid only under a stated condition. By representing claims as $(E_1, R, E_2 \mid C)$ and evaluating models with controlled probes, CondRelBench isolates relation overgeneralization: the failure to preserve the condition under which a relation is supported.

Across prompted LLMs, fine-tuned LLMs, and encoder models, we find that aggregate accuracy and Macro F1 can hide scope-specific failures. Models often support the original scoped claim but fail when the condition is removed, replaced, or contradicted. OGR and SPS expose these failures by measuring over-support for altered claims and consistency across the full probe family. The results also show that synthetic performance does not fully transfer to PubMed and Reddit, especially when conditions are informal, implicit, or distributed across broader discourse.

Finally, the ablation results suggest that tuple extraction and scope-sensitive verification should not be collapsed into a single training objective without care. Probes-only training substantially outperforms probes-plus-tuples training for Phi-4-mini-instruct, indicating that learning to recover tuples does not automatically teach a model to judge condition-altered claims. Future work should explore two-stage extraction-and-verification systems, explicit scope-preservation losses, and training objectives that penalize unsupported generalization. 

\section*{Limitations}
\label{Sec:Limitations}

This study has several limitations. First, the synthetic dataset provides strong control over relations, conditions, and perturbations, but it may overestimate real-world condition-aware performance. Models perform substantially better on Synthetic-Test than on PubMed and Reddit, especially encoder models. For example, BioBERT reaches Macro F1 $0.9942$ on Synthetic-Test, but drops to $0.8845$ on PubMed and $0.5307$ on Reddit. This suggests that synthetic examples are useful for controlled diagnosis, but they should not be treated as a direct substitute for naturally occurring text.

Second, the controlled probe design simplifies the range of ways that conditions appear in real discourse. The probes are useful because they isolate specific scope failures by preserving, removing, replacing, or contradicting the condition. However, real-world condition expressions may be more implicit, distributed across multiple sentences, or dependent on broader document context. As a result, the benchmark may not capture every form of condition-sensitive reasoning required in practical relation extraction systems.

Third, the real-world annotations are selective rather than exhaustive. For PubMed and Reddit, the gold tuples are intended to represent valid condition-bearing relations found in the source, but they do not necessarily cover every possible valid relation in each paragraph or document. This makes the benchmark suitable for targeted probing and transfer evaluation, but less suitable for exhaustive recall-based relation extraction evaluation.

Fourth, the results do not fully explain why probes-plus-tuples training fails for Phi-4-mini-instruct. The ablation shows a large performance drop from probes-only to probes-plus-tuples training, especially on Synthetic-Test, where Macro F1 falls from $0.8959$ to $0.2080$. However, the current tables do not directly test whether this collapse is caused by objective mismatch, tuple-format overfitting, loss imbalance, training-order effects, or the model treating tuple reconstruction as more important than scope-sensitive classification.

Fifth, the study does not fully separate extraction errors from reasoning errors in an end-to-end setting. Experiment 2 gives the model ground-truth relational information and tests whether it can classify controlled probes. This design is useful because it isolates condition-aware inference, but a deployed system would also need to extract the correct entities, relation, and condition before performing scope-sensitive verification. Future work should therefore evaluate full pipelines that combine tuple extraction with condition-aware claim classification.

Finally, the benchmark focuses on three datasets: Synthetic, PubMed, and Reddit. These sources provide useful coverage of controlled, biomedical, and informal user-generated text, but they do not cover all domains where scoped claims matter. Future work should extend the evaluation to legal, financial, policy, and clinical guideline texts, where applying a relation outside its valid condition can also lead to unsupported or harmful conclusions.

Responsible research, artifact, privacy, compute, annotation, and AI-assistance disclosures are provided in Appendix~\ref{app:responsible_research_artifacts}.

\section*{Ethical Concern}
The PubMed subset is based on scientific text from published biomedical papers. The Reddit subset is based on publicly available posts, but we applied additional filtering because Reddit posts may contain personal, identifying, political, or potentially offensive content. We removed any Reddit document that mentioned the name of a person, family member, pet, or similar identifying detail. After this filtering step, the retained Reddit rows did not require further anonymization.

The goal of this filtering was to preserve the linguistic properties of informal discourse while reducing the risk of exposing private identities or personal relationships. Nevertheless, retained Reddit examples may still contain sensitive opinions, political references, or language that some readers may find offensive.

This benchmark should therefore be used only for research on condition-aware relation understanding. It should not be used to identify, profile, contact, or make decisions about individuals, or to infer personal or political attributes of users.
\bibliography{main.bib}
\providecommand{\sectbest}[1]{\textbf{#1}}
\providecommand{\overallbest}[1]{\textbf{#1}}

\appendix

\section{Annotation Protocol}
\label{sec:annotation_protocol}

This appendix describes the annotation protocol used to construct the condition-aware relation extraction benchmark. The task targets relation-bearing claims whose validity is restricted by an explicit or recoverable textual condition. Instead of annotating an ordinary triple $(E_1, R, E_2)$, each accepted claim is represented as a scoped tuple:

\[
(E_1, R, E_2 \mid C),
\]

where $E_1$ and $E_2$ are the relation arguments, $R$ is the relation, and $C$ is the condition under which the relation is valid. The condition is treated as part of the claim meaning, not as an optional modifier. A tuple is incomplete if any of $E_1$, $R$, $E_2$, or $C$ is missing. Relations without a condition are not retained for this benchmark, even when they are otherwise valid relation extraction examples. For example, ``Drug A increases complications in elderly patients'' is not annotated as an unconditional claim about all patients; it is annotated as a relation licensed only under the condition ``in elderly patients.''

Each accepted annotation therefore contains four fields:

\begin{enumerate}
    \item $E_1$: the first relation argument;
    \item $R$: the relation expressed in the text;
    \item $E_2$: the second relation argument;
    \item $C$: the condition under which the relation is valid.
\end{enumerate}

\subsection{Definition of a Condition}
\label{subsec:condition_definition}

A condition is defined as a textual element that restricts the validity of a relation to a particular scope. This scope may be expressed through a population, setting, state, time window, intervention, dosage, study cohort, disease subtype, environmental background, or other contextual boundary. Examples include phrases such as:

\begin{itemize}
    \item ``in elderly patients'';
    \item ``under fasting conditions'';
    \item ``in a Caucasian population'';
    \item ``in this distinct ethnic group'';
    \item ``if the plastic house collapses or gets severely damaged'';
    \item ``in a subset of settings'';
    \item ``when the context changes to state theta.''
\end{itemize}

The key requirement is that the condition must restrict when the relation is supported. A phrase is not accepted as a condition merely because it appears near the relation. It must delimit the scope of the relation. In other words, if the condition is removed, replaced, or contradicted, the original relation should no longer be safely inferred from the source text.

\subsection{Datasets}
\label{subsec:annotation_datasets}

The annotation protocol was applied to three data sources: Synthetic, PubMed, and Reddit. These sources were selected to cover controlled diagnostic examples, formal biomedical text, and informal user-generated text.

\subsubsection{Synthetic Dataset}
\label{subsubsec:synthetic_annotation}

The synthetic dataset contains 3,000 multi-sentence examples. Each synthetic example contains at least three sentences. The synthetic data was generated using a rule-based template system and then rewritten by four linguists to make the text more natural before annotation. This allowed the dataset to maintain controlled structural properties while avoiding overly rigid template language.

The synthetic dataset was constructed to be balanced across the main controlled categories. This balance is important because it allows the benchmark to test condition-aware relation understanding across different structural and semantic forms rather than overrepresenting one type of example.

\begin{table}[h]
\centering
\small
\begin{tabular}{lrr}
\hline
\textbf{Structure type} & \textbf{Count} & \textbf{Percent} \\
\hline
contrastive & 500 & 16.67\% \\
embedded clause & 500 & 16.67\% \\
implicit condition & 500 & 16.67\% \\
inline condition & 500 & 16.67\% \\
leading condition & 500 & 16.67\% \\
restriction & 500 & 16.67\% \\
\hline
\textbf{Total} & \textbf{3,000} & \textbf{100.00\%} \\
\hline
\end{tabular}
\caption{Structure balance in the synthetic dataset.}
\label{tab:synthetic_structure_balance}
\end{table}

The condition types are also balanced. Each condition type appears 600 times, corresponding to 20\% of the synthetic dataset.

\begin{table}[h]
\centering
\small
\begin{tabular}{lrr}
\hline
\textbf{Condition type} & \textbf{Count} & \textbf{Percent} \\
\hline
contrastive & 600 & 20.00\% \\
explicit & 600 & 20.00\% \\
multi-condition & 600 & 20.00\% \\
natural & 600 & 20.00\% \\
weak implicit & 600 & 20.00\% \\
\hline
\textbf{Total} & \textbf{3,000} & \textbf{100.00\%} \\
\hline
\end{tabular}
\caption{Condition type balance in the synthetic dataset.}
\label{tab:synthetic_condition_balance}
\end{table}

The remaining controlled categories are also balanced. Entity type has four values with 750 examples each, while relation type, condition type, and noise type each have five values with 600 examples each.

\begin{table}[h]
\centering
\small
\begin{tabular}{llr}
\hline
\textbf{Category} & \textbf{Value} & \textbf{Count} \\
\hline
entity type & alternative & 750 \\
entity type & conjoined & 750 \\
entity type & distractor & 750 \\
entity type & nested & 750 \\
\hline
relation type & ambiguous & 600 \\
relation type & association & 600 \\
relation type & decrease & 600 \\
relation type & increase & 600 \\
relation type & mixed & 600 \\
\hline
noise type & conflicting signal & 600 \\
noise type & coreference noise & 600 \\
noise type & distractor relation & 600 \\
noise type & partial information & 600 \\
noise type & sentence distance & 600 \\
\hline
\end{tabular}
\caption{Controlled category counts in the synthetic dataset.}
\label{tab:synthetic_category_counts}
\end{table}

The synthetic dataset is therefore a controlled diagnostic subset. It is suitable for testing whether models can recognize condition-bound relations across balanced combinations of relation structure, condition type, entity structure, and noise. However, because the examples are controlled, the synthetic dataset is not treated as a substitute for naturally occurring biomedical or social discourse.

\subsubsection{PubMed Dataset}
\label{subsubsec:pubmed_annotation}

The PubMed dataset was curated from prostate cancer articles, with a focus on genomics studies. This topic was selected because, among three randomly examined biomedical topic areas, prostate cancer genomics showed the highest tendency to contain condition-bearing relations. The PubMed search was based on the following query:

{\scriptsize
\begin{verbatim}
("Prostatic Neoplasms"[MeSH] OR "prostate cancer")
AND
("Neoplasm Recurrence, Local"[MeSH] OR recurrence
 OR "biochemical recurrence" OR relapse)
AND
("Genomics"[MeSH] OR genomic OR transcriptomic
 OR "gene expression profiling" OR "genomic classifier")
AND
(prognosis OR prediction OR risk stratification)
\end{verbatim}
}

The data was sampled on March 12, 2026, and the sampled collection was frozen before annotation. From this frozen set, 200 full-text papers were randomly selected for annotation. Annotation was performed over the full paper, but the retained annotation unit was the paragraph. Each paper produced a different number of candidate paragraphs. Across the 200 papers, 2,892 rows were annotated during the first candidate annotation stage.

The final PubMed set contains 681 examples and 1,397 unique tuples. The PubMed subset represents formal biomedical text, where conditions often appear as study populations, disease subtypes, clinical cohorts, ethnic groups, sample restrictions, treatment settings, genomic contexts, or experimental conditions.

\subsubsection{Reddit Dataset}
\label{subsubsec:reddit_annotation}

The Reddit dataset was created on April 1, 2026. Unlike PubMed, which is formal biomedical writing, Reddit contains informal user-generated discourse. This makes it useful for evaluating whether condition-aware relation understanding transfers to noisier, less standardized language.

The candidate Reddit data was collected from four sources:

\begin{table}[h]
\centering
\small
\begin{tabular}{lr}
\hline
\textbf{Subreddit} & \textbf{Candidate posts} \\
\hline
ChangeMyView & 484 \\
Relationships & 447 \\
ExplainLikeImFive & 336 \\
ProstateCancer & 479 \\
\hline
\end{tabular}
\caption{Candidate Reddit posts before final validation.}
\label{tab:reddit_candidate_sources}
\end{table}

After the second-stage validation, 695 Reddit examples were retained. The final distribution is shown in Table~\ref{tab:reddit_final_distribution}.

\begin{table}[h]
\centering
\small
\begin{tabular}{lrr}
\hline
\textbf{Subreddit} & \textbf{Row count} & \textbf{Percent} \\
\hline
changemyview & 313 & 45.04\% \\
relationships & 230 & 33.09\% \\
ProstateCancer & 101 & 14.53\% \\
explainlikeimfive & 51 & 7.34\% \\
\hline
\textbf{Total} & \textbf{695} & \textbf{100.00\%} \\
\hline
\end{tabular}
\caption{Final Reddit distribution after validation.}
\label{tab:reddit_final_distribution}
\end{table}

For Reddit, the annotation unit is the full post/document rather than a paragraph. This is because relation-bearing claims in Reddit posts may depend on broader local context, especially when users build an argument over several sentences. All retrieved posts were eligible for screening, but posts were retained only when they contained valid condition-bearing relational claims.

\subsection{Privacy and Ethical Filtering for Reddit}
\label{subsec:reddit_privacy}

The Reddit subset was filtered with privacy considerations in mind. Any document that mentioned a name of a person, family member, pet, or similar identifying personal detail was dropped. No retained row required further anonymization because the retained rows already satisfied this principle. The goal was to avoid including posts that could expose personal identity or private relationships while still preserving the linguistic properties of informal discourse.

\subsection{Annotation Unit and Tuple Fields}
\label{subsec:annotation_unit_tuple_fields}

The annotation unit differs slightly across datasets because the sources have different discourse structures. Synthetic examples are multi-sentence paragraphs. PubMed examples are paragraph-level spans extracted from full papers. Reddit examples are full posts/documents.

For each valid unit, annotators identify a tuple of the form:

\[
(E_1, R, E_2 \mid C).
\]

Each field has a specific annotation function.

\paragraph{$E_1$: First relation argument.}
$E_1$ is the first argument participating in the relation. It may be a named entity, noun phrase, biomedical concept, pronoun, or discourse referent, depending on the dataset. In PubMed, $E_1$ is often a biomedical entity, study group, mutation, molecular marker, intervention, or measured factor. In Reddit, $E_1$ may be a person, object, situation, action, or informal referent. In Synthetic, $E_1$ is controlled by the generation template.

\paragraph{$R$: Relation.}
$R$ is the relation expressed between $E_1$ and $E_2$. In the synthetic data, relation families are controlled and include increase, decrease, association, ambiguous effect, and mixed relations. In PubMed and Reddit, the relation is extracted from the naturally occurring text and is not restricted to the synthetic label categories. Therefore, real-world relations are treated as free-text relation spans.

\paragraph{$E_2$: Second relation argument.}
$E_2$ is the second argument participating in the relation. Like $E_1$, it may be a named entity, noun phrase, biomedical concept, pronoun, or discourse referent. A relation cannot be accepted if $E_2$ is missing. This is especially important for Reddit, where some surface-level claims may be incomplete, elliptical, or conversationally implied.

\paragraph{$C$: Condition.}
$C$ is the condition that restricts when the relation holds. It may appear before the relation, after the relation, inside an embedded clause, in an ``if'' clause, as a population qualifier, as a contextual state, or as a multi-part condition. A tuple is accepted only when the condition is sufficiently recoverable from the text. If the condition is missing, vague, or does not actually license the relation, the candidate tuple is rejected.

\subsection{Span Matching and Boundary Rules}
\label{subsec:span_matching}

Span matching was used to validate agreement over the annotated tuple fields. For the annotation agreement stage, a token-level intersection-over-union score was used. Given a candidate span $A$ and a reference span $B$, token-level IOU is computed as:

\[
\text{IOU}(A,B) = \frac{|A \cap B|}{|A \cup B|}.
\]

A token overlap threshold of 0.5 was used for span agreement in the annotation setting. This threshold was applied to long noun phrases, abbreviations, pronouns, nested entities, relation spans, and condition spans. The threshold allows minor boundary differences while still requiring substantial overlap between annotators.

For the synthetic extraction evaluation in Table 1, exact matching was used for $E_1$ and $E_2$, while token overlap with IOU of 0.5 was used for $R$ and $C$. For the relaxed cross-domain extraction evaluation in Table 2, all three datasets used a relaxed token overlap threshold of 0.3. This lower threshold was used because naturally occurring PubMed and Reddit text can contain longer, noisier, and less standardized spans.

\subsection{Two-Stage Annotation Process}
\label{subsec:two_stage_annotation}

The annotation process followed two stages: candidate annotation and validation annotation.

\subsubsection{Stage 1: Candidate Annotation}
\label{subsubsec:stage1_candidate_annotation}

The first stage was a candidate annotation stage. Its purpose was to identify candidate scoped relation tuples of the form $(E_1, R, E_2 \mid C)$ from each eligible text. In this stage, one main annotator annotated the full candidate set. A second annotator independently annotated a validation subset of the PubMed and Reddit data. For the Synthetic dataset, the full span was validated because the examples were controlled and generated from known templates.

The first-stage annotator was responsible for identifying:

\begin{enumerate}
    \item the relevant paragraph or document;
    \item the first relation argument $E_1$;
    \item the relation $R$;
    \item the second relation argument $E_2$;
    \item the condition $C$ that scopes the relation.
\end{enumerate}

The first-stage annotation was not designed to be exhaustive for real-world text. Instead, it was designed to identify high-confidence condition-bearing relations. This is important because PubMed papers and Reddit posts may contain several possible relations, but not every relation is relevant to the condition-aware benchmark. The goal was therefore to collect reliable scoped relation tuples, not to annotate every possible relation in each document.

Annotators were trained using instructional videos. The annotators had biomedical degrees or biomedical minors, which was particularly important for the PubMed subset because the biomedical text often includes study populations, molecular concepts, recurrence outcomes, genomic classifiers, clinical risk factors, and cohort-specific claims.

For PubMed, the first-stage process produced 2,892 candidate annotation rows from the 200 sampled full-text papers. A validation check was then performed using span matching. This check showed that the main annotator annotated approximately 5\% more candidate tuples than the validator.

\begin{table*}[ht]
\centering
\small
\begin{tabular}{p{2.4cm}p{3.0cm}p{3.7cm}p{4.2cm}}
\hline
\textbf{Stage} & \textbf{Dataset} & \textbf{Annotators} & \textbf{Role} \\
\hline
Stage 1: Candidate annotation & PubMed and Reddit & One main annotator for the full set; one second annotator for the validation subset & Identify candidate scoped tuples by marking $E_1$, $R$, $E_2$, and $C$. \\
\hline
Stage 1: Candidate annotation & Synthetic & Full span validated because the data was controlled & Verify controlled synthetic tuples generated from known templates. \\
\hline
Stage 2: Validation annotation & PubMed and Reddit & Two validators different from the Stage 1 annotators & Independently mark each candidate tuple as accepted or rejected. \\
\hline
\end{tabular}
\caption{Annotator roles across the two annotation stages.}
\label{tab:annotator_roles}
\end{table*}

Agreement in Stage 1 was computed using token-level span matching. A candidate span was treated as matching the validator span when it reached at least 50\% token overlap. This threshold allowed minor boundary differences while still requiring substantial agreement on the annotated span. The span-matching audit was reported for the condition component and for the full tuple. We did not treat this Stage 1 audit as a full adjudication process; it was used to check whether the main annotator's candidate tuples were sufficiently aligned with an independent validator before moving to the stricter validation stage.

\begin{table}[ht]
\centering
\small
\begin{tabular}{lcc}
\hline
\textbf{Metric} & \textbf{PubMed} & \textbf{Reddit} \\
\hline
\texttt{Condition\_EM} & 0.8750 & 0.8500 \\
\texttt{Condition\_F1} & 0.9370 & 0.9621 \\
\hline
\texttt{Full\_Tuple\_Precision} & 0.9373 & 0.6663 \\
\texttt{Full\_Tuple\_Recall} & 0.9183 & 0.6734 \\
\texttt{Full\_Tuple\_F1} & 0.9164 & 0.6593 \\
\hline
\end{tabular}
\caption{Stage 1 span matching evaluation for PubMed and Reddit using a 50\% token-overlap matching threshold.}
\label{tab:stage1_span_matching}
\end{table}

The PubMed results show strong alignment between the main annotator and the validation annotator, both for condition matching and full-tuple matching. The Reddit results are lower at the full-tuple level, which reflects the difficulty of annotating informal user-generated text. Reddit posts often contain pronouns, implicit arguments, missing relation arguments, loose causal or explanatory language, and conditions that are spread across longer stretches of text.

\subsubsection{Stage 2: Validation Annotation and Disagreement Handling}
\label{subsubsec:stage2_validation_annotation}

The second stage was used for PubMed and Reddit only. In this stage, two validators different from the Stage 1 annotators independently reviewed the candidate tuples. Each candidate tuple was marked as either \textit{Yes} or \textit{No}. A \textit{Yes} decision means that the validator fully agreed that the candidate tuple contained a valid $E_1$, $R$, $E_2$, and $C$. A \textit{No} decision means that the validator rejected the tuple because one or more of the required components was missing, incorrect, unsupported, or not properly scoped.

Partial agreement was not allowed. This means that a validator could not accept only part of a tuple. For example, if the validator agreed with $E_1$, $R$, and $E_2$ but disagreed with the condition $C$, the candidate was marked as \textit{No}. This strict decision rule was used because the benchmark depends on complete scoped tuples. A relation without the correct condition is not a valid annotation for this task.

Disagreements were not adjudicated. Instead, they were discarded. Only rows where both validators marked \textit{Yes} were retained in the final real-world datasets. This conservative decision was made because the goal of the real-world annotation was to build a high-confidence set of condition-bearing relation tuples for targeted probing and transfer evaluation, not to exhaustively annotate every possible relation in each paragraph or post.

\begin{table*}[ht]
\centering
\small
\begin{tabular}{llcc}
\hline
\textbf{Validator 1} & \textbf{Validator 2} & \textbf{PubMed Count} & \textbf{Reddit Count} \\
\hline
Yes & Yes & 681 & 695 \\
No & Yes & 14 & 26 \\
Yes & No & 34 & 34 \\
No & No & 5 & 3 \\
Invalid & Invalid & 3 & 8 \\
\hline
\textbf{Total} & & \textbf{737} & \textbf{766} \\
\hline
\end{tabular}
\caption{Stage 2 validation outcomes for PubMed and Reddit. Only Yes/Yes rows were retained in the final datasets.}
\label{tab:stage2_validation_counts}
\end{table*}

The final retained PubMed set contains 681 examples, and the final retained Reddit set contains 695 examples. Rows with disagreement were removed, and invalid rows were also removed. The invalid rate was 0.41\% for PubMed, computed as $3/737$, and 1.04\% for Reddit, computed as $8/766$.

Raw agreement was used as the main validation statistic because the Stage 2 validation set was already biased toward positive candidates after Stage 1. In other words, the validation task was not a balanced classification task over arbitrary text; it was a review of candidate tuples that had already been selected as likely condition-bearing relations. For this reason, raw agreement was more directly interpretable than chance-corrected agreement for this validation step.

\subsubsection{Probe Generation and Manual Filtering}
\label{subsubsec:probe_generation_filtering}

After tuple validation, each retained scoped tuple was converted into five controlled probes. The probes were created to test whether a model preserves the scope of the relation when the condition is preserved, replaced, removed, or contradicted, and when the relation itself is contradicted.

Probe generation was handcrafted. We generated five probes for each tuple. Each generated probe contained a probe number, probe type, evidence paragraph, claim, and gold answer. The allowed gold answers were standardized as:

\begin{itemize}
    \item \textit{Supported};
    \item \textit{Not Supported};
    \item \textit{Not Enough Evidence}.
\end{itemize}

The label \textit{Not Supported} was used for both relation contradiction and condition contradiction. The label \textit{Not Enough Evidence} was used when the condition was replaced with an unsupported alternate condition or removed entirely. This distinction is important because a condition-removed claim is not always contradicted by the evidence; rather, it is broader than what the evidence supports.

\begin{table*}[ht]
\centering
\small
\begin{tabular}{p{1.3cm}p{3.3cm}p{6.4cm}p{2.5cm}}
\hline
\textbf{Probe} & \textbf{Type} & \textbf{Modification} & \textbf{Gold Label} \\
\hline
P1 & Relation contradiction & Contradicts the original relation while preserving $E_1$, $E_2$, and $C$. & Not Supported \\
\hline
P2 & Condition replacement & Replaces the original condition with a semantically and pragmatically different unsupported condition. & Not Enough Evidence \\
\hline
P3 & Condition removal & Removes the condition entirely, making the relation broader than the evidence supports. & Not Enough Evidence \\
\hline
P4 & Supported original tuple & Preserves the original $E_1$, $R$, $E_2$, and $C$. & Supported \\
\hline
P5 & Condition contradiction & Contradicts or negates the original condition while preserving the relation. & Not Supported \\
\hline
\end{tabular}
\caption{Probe types generated from each scoped tuple.}
\label{tab:probe_types}
\end{table*}

The generation prompt required the model to keep $E_1$ and $E_2$ exactly as written, avoid yes/no questions, write each claim as a declarative statement, avoid repetitive wording, and return valid JSON only. The evidence was required to be written as a natural paragraph rather than as a repeated template. The prompt also specified that relation contradiction and condition contradiction must not be confused: P1 must contradict the relation, while P5 must contradict the condition.

Generated probes were then manually inspected before inclusion in the final probe set. A probe was rejected if it failed to match its intended probe type. For example, a P1 probe was rejected if it contradicted the condition instead of the relation. A P2 probe was rejected if the alternate condition was not meaningfully different from the original condition. A P3 probe was rejected if it still mentioned the original condition. A P4 probe was rejected if it failed to preserve the original scoped tuple. A P5 probe was rejected if it contradicted the relation rather than the condition.

This manual filtering step was necessary because the benchmark depends on controlled differences between the five probe types. Without this filtering, a model could be evaluated on noisy probes where the intended label does not match the actual semantic change.

\subsubsection{Selective Annotation and Interpretation of Real-World Results}
\label{subsubsec:selective_annotation_interpretation}

The PubMed and Reddit annotations are selective rather than exhaustive. This means that the retained tuples are high-confidence condition-bearing relations, but they do not necessarily cover every valid relation in each PubMed paragraph or Reddit post. Therefore, the real-world datasets should not be interpreted as exhaustive relation extraction benchmarks.

This distinction affects how the extraction results should be read. The real-world extraction results are best interpreted as targeted recovery of validated condition-bearing tuples, not as complete measurement of all possible relations in the text. In contrast, the probe evaluation is the central scope-preservation test because each probe is derived from a validated scoped tuple and has a controlled gold label.

Accordingly, the real-world data is used mainly for targeted probing and transfer evaluation. This allows the benchmark to test whether models preserve conditional scope in naturally occurring text while avoiding the stronger claim that every relation in the source documents has been exhaustively annotated.

\subsection{Accepted and Rejected Annotation Examples}
\label{subsec:accepted_rejected_examples}

This subsection provides concrete examples from the annotation process. These examples show what counts as a valid scoped tuple and what is rejected.

\subsubsection{PubMed Example}
\label{subsubsec:pubmed_example}

\paragraph{Source paragraph.}
\begin{quote}
\small
However, one subsequent study failed to find evidence for a frequent mutation in a Caucasian population, possibly due to the difference in the composition of the study cohort. Thus, in this study, we investigated the mutation status of MED12 in 102 radical prostatectomy specimens from Korean men to explore its relevance in this distinct ethnic group, testing the possibility that MED12 may represent another point of molecular difference between the Caucasian and Asian population.
\end{quote}

\paragraph{Accepted tuples.}
\begin{enumerate}
    \item
\[
\begin{aligned}
E_1 &= \begin{minipage}[t]{0.85\linewidth}
``the mutation status of MED12 in 102 radical prostatectomy specimens from Korean men''
\end{minipage} \\
R &= \text{``to explore its''} \\
E_2 &= \text{``relevance''} \\
C &= \text{``in this distinct ethnic group''}
\end{aligned}
\]

    This tuple is accepted because the paragraph explicitly states that a subsequent study failed to find evidence for a frequent mutation under the population condition ``in a Caucasian population.'' The condition is not incidental; it restricts the scope of the finding.

\item
\[
\begin{aligned}
E_1 &= 
\begin{minipage}[t]{0.85\linewidth}
``the mutation status of MED12 in 102 radical prostatectomy specimens from Korean men''
\end{minipage}\\
R &= \text{``to explore its''} \\
E_2 &= \text{``relevance''} \\
C &= \text{``in this distinct ethnic group''}
\end{aligned}
\]
    This tuple is accepted because the investigation of MED12 relevance is scoped to a distinct ethnic group. The condition provides the population boundary for the relation.
\end{enumerate}

\paragraph{Rejected tuple.}
\[
\begin{aligned}
E_1 &= \text{``MED12''}\\
R &= \text{``may represent''} \\
E_2 &= 
\begin{minipage}[t]{0.85\linewidth}
``another point of molecular difference between the Caucasian and Asian population''
\end{minipage} \\
C &= \text{``testing the possibility''}
\end{aligned}
\]

This tuple is rejected because it is based on a speculative nested clause. The phrase ``testing the possibility'' does not function as a substantive condition that licenses the relation. It signals speculation or research intent rather than a scope under which the relation is established. Therefore, this row is not retained as a valid condition-bearing tuple.

\subsubsection{Reddit Example}
\label{subsubsec:reddit_example}

\paragraph{Source post.}
\begin{quote}
\small
You probably seen those houses made of plastic bottles that claim to be eco friendly, right? They dont. The reason behind this is because if you just upcycle plastic to make houses, you will create more microplastics to float around the air and make us unhealthier. In other words, if the plastic house collapses or gets severely damaged, it is almost impossible to repair it and therefore all your house becomes useless trash that goes in the landfill. Plastic houses also arent really safe, so they wont actually provide protection against bullets unless its just lots of plastic huddled together to make tons of bulletproof layers, which makes more microplastics to appear.

It is better to use leftover pallets than plastic waste because pallets are also quite wasted a lot, and theyre at least biodegradable because theyre usually made of wood.

In conclusion, despite all the benefits plastic homes can offer, they still harm the environment like big toxic beams generating a lot of waste. After all, wood is better for building eco friendly homes because they also have a high quantity left and theyre biodegradable too.
\end{quote}

\newcommand{\tupletext}[1]{%
\begin{minipage}[t]{0.85\linewidth}
\raggedright ``#1''
\end{minipage}%
}

\paragraph{Accepted tuples.}
\begin{enumerate}
    \item
    \[
    \begin{aligned}
    E_1 &= \text{``you''} \\
    R &= \text{``will create''} \\
    E_2 &= \tupletext{more microplastics to float around the air and make us unhealthier} \\
    C &= \tupletext{if you just upcycle plastic to make houses}
    \end{aligned}
    \]

    This tuple is accepted because the relation is explicitly conditional. The claim is not that the reader always creates microplastics. The claim is that this effect occurs under the condition of upcycling plastic to make houses.

    \item
    \[
    \begin{aligned}
    E_1 &= \text{``it''} \\
    R &= \text{``is almost impossible to repair''} \\
    E_2 &= \text{``it''} \\
    C &= \tupletext{if the plastic house collapses or gets severely damaged}
    \end{aligned}
    \]

    This tuple is accepted because the repair difficulty is explicitly scoped to the condition that the plastic house collapses or is severely damaged.
\end{enumerate}

\paragraph{Rejected tuples.}
The following candidates are rejected:

\[
\begin{aligned}
E_1 &= \tupletext{those houses made of plastic bottles} \\
R &= \text{``claim to be eco friendly''} \\
E_2 &= \text{``''} \\
C &= \text{``''}
\end{aligned}
\]

\[
\begin{aligned}
E_1 &= \text{``the plastic house''} \\
R &= \text{``collapses''} \\
E_2 &= \text{``''} \\
C &= \tupletext{if the plastic house collapses or gets severely damaged}
\end{aligned}
\]

\[
\begin{aligned}
E_1 &= \text{``the plastic house''} \\
R &= \text{``gets severely damaged''} \\
E_2 &= \text{``''} \\
C &= \tupletext{if the plastic house collapses or gets severely damaged}
\end{aligned}
\]

\[
\begin{aligned}
E_1 &= \text{``Plastic houses''} \\
R &= \text{``arent really safe''} \\
E_2 &= \text{``''} \\
C &= \text{``''}
\end{aligned}
\]

These tuples are rejected because they are incomplete. Some have no $E_2$, while others have no valid condition. A relation may be linguistically meaningful in the paragraph, but it is not accepted for this benchmark unless it can be represented as a complete scoped tuple with $E_1$, $R$, $E_2$, and $C$.

\subsubsection{Synthetic Example}
\label{subsubsec:synthetic_example}

\paragraph{Source paragraph.}
\begin{quote}
\small
The text begins by presenting the entities before restricting the relevant environment. In a subset of settings, Protein Z amplifies Signal Kappa. Conversely, when the context changes to state theta, Protein Z suppresses Signal Kappa. Consequently, the passage demands that the reader monitor which environment corresponds to which assertion. No generalized conclusion is permitted beyond the specific background mentioned. A secondary comment notes that the documentation is unfinished, yet the primary data pair remains sufficiently clear for labeling. The immediate focus for the extraction stays fixed on the circumstances related to setting upsilon.
\end{quote}

\paragraph{Accepted tuples.}
\begin{enumerate}
    \item
    \[
    \begin{aligned}
    E_1 &= \text{``Protein Z''} \\
    R &= \text{``amplifies''} \\
    E_2 &= \text{``Signal Kappa''} \\
    C &= \text{``setting upsilon''}
    \end{aligned}
    \]

    \item
    \[
    \begin{aligned}
    E_1 &= \text{``Protein Z''} \\
    R &= \text{``suppresses''} \\
    E_2 &= \text{``Signal Kappa''} \\
    C &= \text{``state theta''}
    \end{aligned}
    \]
\end{enumerate}

This example illustrates why condition-aware annotation is necessary. The same entity pair, Protein Z and Signal Kappa, appears with two different relations under two different conditions. Protein Z amplifies Signal Kappa under one condition and suppresses Signal Kappa under another. If the condition is ignored, the example becomes contradictory or ambiguous. With the condition included, both tuples are valid because each relation is scoped to a different context.

\subsection{Probe Construction from Annotated Tuples}
\label{subsec:probe_construction_annotation}

After the scoped tuples were annotated and validated, each tuple was converted into a set of five classification probes. The probes are designed to test whether a model preserves the condition under which the relation is valid. Each probe is derived from the same base tuple $(E_1, R, E_2 \mid C)$ but modifies the relation or condition in a controlled way.

The five probe types are:

\begin{enumerate}
    \item \textbf{Probe 1: Relation Contradiction.}
    The relation is contradicted while $E_1$, $E_2$, and $C$ are preserved. The gold label is \textit{Not Supported}.

    \[
    p_{i1}: (E_{1i}, \neg R_i, E_{2i} \mid C_i)
    \]

    \item \textbf{Probe 2: Condition Replacement.}
    The original condition is replaced with an unsupported alternate condition. The relation itself is not contradicted. The gold label is \textit{Not Enough Evidence}.

    \[
    p_{i2}: (E_{1i}, R_i, E_{2i} \mid C'_i)
    \]

    Alternate conditions are selected from a bag of conditions that are semantically and pragmatically different from the original condition.

    \item \textbf{Probe 3: Condition Removal.}
    The condition is removed entirely, making the relation broader than the evidence supports. The gold label is \textit{Not Enough Evidence}.

    \[
    p_{i3}: (E_{1i}, R_i, E_{2i})
    \]

    This probe is especially important because it tests whether the model treats the relation as generally true once it recognizes the entity--relation pattern.

    \item \textbf{Probe 4: Supported Original Tuple.}
    The original scoped tuple is preserved. The gold label is \textit{Supported}.

    \[
    p_{i4}: (E_{1i}, R_i, E_{2i} \mid C_i)
    \]

    \item \textbf{Probe 5: Condition Contradiction.}
    The condition is contradicted while the original relation is not contradicted. The gold label is \textit{Not Supported}.

    \[
    p_{i5}: (E_{1i}, R_i, E_{2i} \mid \neg C_i)
    \]

    The contradictory condition is generated through semantic opposition or negation.
\end{enumerate}

The final probe set contains 37,268 Synthetic queries, 6,961 PubMed queries, and 7,883 Reddit queries.

\subsection{Probe Label Semantics}
\label{subsec:probe_label_semantics}

The probe labels are defined as follows:

\paragraph{Supported.}
A probe is labeled \textit{Supported} only when the claim preserves the original $E_1$, $R$, $E_2$, and $C$, and the source evidence supports the full scoped tuple.

\paragraph{Not Supported.}
A probe is labeled \textit{Not Supported} when the claim contradicts the evidence. This occurs in two probe types: relation contradiction and condition contradiction. In relation contradiction, the relation itself is reversed or opposed while the original condition is preserved. In condition contradiction, the relation is kept but the condition is contradicted.

\paragraph{Not Enough Evidence.}
A probe is labeled \textit{Not Enough Evidence} when the source does not provide enough support for the modified claim. This occurs when the original condition is replaced with another unsupported condition or when the condition is removed entirely. In these cases, the relation may look familiar, but the claim is broader or differently scoped than the source evidence allows.

This label design separates contradiction from unsupported generalization. A condition-removed claim is not necessarily contradicted by the text. Instead, it is not sufficiently supported because the text only licenses the relation under a specific condition.

\subsection{Annotation-to-Evaluation Mapping}
\label{subsec:annotation_evaluation_mapping}

Each annotation field is connected to one or more evaluation metrics. This mapping ensures that the evaluation reflects the annotation design.

\begin{table*}[h]
\centering
\small
\begin{tabular}{lll}
\hline
\textbf{Annotation item} & \textbf{Meaning} & \textbf{Evaluation metric} \\
\hline
$E_1$ & First relation argument & $F1_{E1}$, $R_{E1}$ \\
$R$ & Relation span or label & $F1_R$, $R_R$ \\
$E_2$ & Second relation argument & $F1_{E2}$, $R_{E2}$ \\
$C$ & Condition span & $F1_C$, $R_C$ \\
$(E_1,R,E_2,C)$ & Full scoped tuple & $F1_{tuple}$, $R_{tuple}$ \\
$(E_1,R,E_2)$ & Base relation without condition & $F1_{RE}$, $R_{RE}$ \\
Probe label & Scope-sensitive judgment & Accuracy, Macro F1, OGR, SPS \\
\hline
\end{tabular}
\caption{Mapping between annotation fields and evaluation metrics.}
\label{tab:annotation_evaluation_mapping}
\end{table*}

This mapping is important because ordinary relation extraction metrics can hide condition failures. A model may correctly identify $E_1$, $R$, and $E_2$, but still fail to identify $C$. Such a model would appear successful under ordinary relation extraction but would fail the condition-aware task. For this reason, the benchmark reports both ordinary relation extraction scores and condition-aware scores.

\subsection{Scope-Sensitive Evaluation}
\label{subsec:scope_sensitive_evaluation}

The probe classification task is evaluated using accuracy, Macro F1, label-level F1, probe-level accuracy, Overgeneralization Rate (OGR), and Scope Preservation Score (SPS).

OGR measures how often a model incorrectly labels modified-condition or no-condition probes as \textit{Supported}. Lower OGR is better. A high OGR means that the model tends to accept a relation even when the condition has been changed or removed.

\[
\text{OGR}
=
\frac{
\sum_{k \in \{2,3,5\}} \sum_i \mathbb{1}[\hat{y}_{ik} = \text{Supported}]
}{
N_2 + N_3 + N_5
}.
\]

SPS measures whether a model correctly classifies the full family of probes derived from the same base tuple. Let $S_i = \{q_{i1}, \ldots, q_{ik}\}$ denote the probe family for base claim $i$, with gold label vector $Y_i$ and predicted label vector $\hat{Y}_i$. SPS is defined as:

\[
\text{SPS}
=
\frac{1}{N}
\sum_{i=1}^{N}
\mathbb{1}[\hat{Y}_i = Y_i].
\]

SPS is strict because a model receives credit only when it preserves the correct judgment across the entire probe family. This is appropriate for condition-aware relation understanding because a model should not merely support the original scoped claim; it should also reject or abstain from unsupported condition changes.

\subsection{Why Disagreements Were Dropped}
\label{subsec:why_drop_disagreements}

All Stage 2 disagreements were removed from the final real-world datasets. This decision was made to prioritize precision and annotation reliability. Since the task is not exhaustive relation extraction, retaining disputed rows would introduce unnecessary ambiguity into the probe construction stage. A disputed tuple could produce probes that are unclear, weakly supported, or based on a condition that one validator did not accept.

Dropping disagreements also supports the benchmark's primary goal: evaluating whether models preserve conditional scope. The benchmark does not require every valid relation in a paragraph or post. It requires a set of validated scoped tuples for which the condition-bearing relation is sufficiently clear. Therefore, the final PubMed and Reddit sets contain only high-confidence Yes/Yes rows.

\subsection{Annotation Limitations}
\label{subsec:annotation_limitations}

The annotation protocol has several limitations.

First, the real-world annotations are selective rather than exhaustive. For PubMed and Reddit, the goal was to collect high-confidence condition-bearing relation tuples, not to annotate every possible valid relation in each paragraph or post. Therefore, the real-world datasets are best suited for targeted probing and transfer evaluation rather than exhaustive relation extraction benchmarking.

Second, the synthetic dataset is controlled and balanced, which makes it useful for diagnostic evaluation, but it may overestimate real-world performance. Synthetic examples are designed to cover known structures, condition types, relation types, and noise patterns. Real-world PubMed and Reddit examples contain more linguistic variation, less predictable structure, and more implicit context.

Third, Reddit annotation is especially difficult because informal posts often contain incomplete claims, pronouns, missing arguments, loose causal language, and conversational context. This explains why Reddit had a lower first-stage IOU score than PubMed.

Fourth, the probe construction process simplifies the range of ways conditions appear in natural discourse. Controlled probes are useful because they isolate the model's ability to preserve scope, but they may not capture every form of condition-sensitive reasoning required in practical relation extraction systems.

Finally, the benchmark does not fully separate extraction errors from reasoning errors in an end-to-end deployment setting. The tuple extraction task evaluates whether models can recover $E_1$, $R$, $E_2$, and $C$, while the probe task evaluates whether models preserve the condition once the tuple is available. In real systems, both capabilities are required.

\subsection{Summary}
\label{subsec:annotation_summary}

The annotation process produces scoped relational tuples of the form $(E_1, R, E_2 \mid C)$ across Synthetic, PubMed, and Reddit data. The Synthetic dataset provides controlled and balanced diagnostic examples. The PubMed dataset provides formal biomedical examples from prostate cancer genomics papers. The Reddit dataset provides informal user-generated examples from multiple subreddits.

The annotation process uses a two-stage design. Stage 1 identifies candidate scoped tuples. Stage 2 validates PubMed and Reddit candidates using independent annotators. Only rows with full agreement are retained. The resulting tuples are then converted into five controlled probes that test relation contradiction, condition replacement, condition removal, original support, and condition contradiction.

This annotation design directly supports the central goal of the benchmark: measuring whether models understand not only that a relation holds, but also when it holds.

\clearpage

\section{Experiment 1 Conditional Tuple Extraction Evaluation}
\label{app:exp1-evaluation}
Experiment~1 evaluates whether models can extract conditional causal relations from text. 
Unlike standard causal relation extraction, where the target is usually represented as a triple 
\((e_1,r,e_2)\), our task requires the model to recover the full scoped causal tuple

\[
\tau = (e_1, r, e_2, c),
\]

where \(e_1\) and \(e_2\) are the causal arguments, \(r\) is the relation, and \(c\) is the condition under which the relation holds. This formulation is important because the condition is not an optional modifier. It determines the scope of the causal claim. A model that extracts \((e_1,r,e_2)\) but omits \(c\) may convert a condition-specific claim into a broader claim that is not licensed by the paragraph.

\paragraph{Gold and predicted tuple sets.}
For each paragraph \(x_i\), let

\[
G_i = \{\tau_{i1}, \tau_{i2}, \ldots, \tau_{im_i}\}
\]

denote the set of annotated gold conditional causal tuples, and let

\[
\hat{G}_i = \{\hat{\tau}_{i1}, \hat{\tau}_{i2}, \ldots, \hat{\tau}_{in_i}\}
\]

denote the set of tuples predicted by a model. Each gold tuple is represented as

\[
\tau_{ij} = (e_{1ij}, r_{ij}, e_{2ij}, c_{ij}),
\]

and each predicted tuple is represented as

\[
\hat{\tau}_{ik} = (\hat{e}_{1ik}, \hat{r}_{ik}, \hat{e}_{2ik}, \hat{c}_{ik}).
\]

The goal is to measure how well \(\hat{G}_i\) covers the annotated conditional causal information in \(G_i\).

\paragraph{Field-level matching.}
To compare a predicted field with a gold field, we use a relaxed similarity function \(s(a,b)\). For Synthetic, where the generated gold annotations are complete and controlled, stricter matching is appropriate. For PubMed and Reddit, where surface forms may be more variable, we use relaxed field matching to avoid treating minor surface variation as complete failure. The relaxed score is defined as
\[
\begin{aligned}
s(a,b) = \max \big(&
s_{\mathrm{IoU}}(a,b),\;
s_{\mathrm{F1}}(a,b),\\
&
s_{\mathrm{char}}(a,b),\;
s_{\mathrm{contain}}(a,b)
\big).
\end{aligned}
\]

where \(s_{\mathrm{IoU}}\) is token Jaccard overlap, \(s_{\mathrm{F1}}\) is token-level F1, \(s_{\mathrm{char}}\) is character-level similarity, and \(s_{\mathrm{contain}}\) measures substring containment. A field is treated as matched when

\[
s(a,b) \geq \theta,
\]

where \(\theta=0.30\) for the relaxed PubMed and Reddit evaluation. This relaxed criterion is used only to decide whether a predicted field sufficiently recovers the annotated gold field; it does not change the definition of the target tuple.

\paragraph{Tuple matching.}
A predicted tuple \(\hat{\tau}_{ik}\) matches a gold tuple \(\tau_{ij}\) if all four fields match:
\[
\begin{aligned}
M_{\mathrm{tuple}}(\hat{\tau}_{ik},\tau_{ij})
= \mathbb{1}\Big[
& s(\hat{e}_{1ik},e_{1ij}) \geq \theta
\land s(\hat{r}_{ik},r_{ij}) \geq \theta \\
& \land s(\hat{e}_{2ik},e_{2ij}) \geq \theta
\land s(\hat{c}_{ik},c_{ij}) \geq \theta
\Big].
\end{aligned}
\]

We also define a standard relation extraction match that ignores the condition:

\[
\begin{aligned}
M_{\mathrm{RE}}(\hat{\tau}_{ik},\tau_{ij})
= \mathbb{1}\Big[
& s(\hat{e}_{1ik},e_{1ij}) \geq \theta
\land s(\hat{r}_{ik},r_{ij}) \geq \theta \\
& \land s(\hat{e}_{2ik},e_{2ij}) \geq \theta
\Big].
\end{aligned}
\]
This distinction allows us to measure the gap between ordinary causal relation extraction and full conditional causal extraction.

\paragraph{One-to-one alignment.}
For each paragraph, predicted tuples are greedily aligned to gold tuples under a one-to-one constraint. Once a gold tuple is matched to a prediction, it cannot be reused for another predicted tuple. This prevents a model from receiving repeated credit for the same gold annotation. Let \(H_{\mathrm{tuple}}\) be the total number of matched full tuples across the dataset, and let \(H_{\mathrm{RE}}\), \(H_{E_1}\), \(H_R\), \(H_{E_2}\), and \(H_C\) denote the corresponding matched counts for the relation triple and individual fields.

Let

\[
|G| = \sum_i |G_i|
\]

be the total number of annotated gold tuples, and

\[
|\hat{G}| = \sum_i |\hat{G}_i|
\]

be the total number of predicted tuples.

\paragraph{Full tuple precision, recall, and F1.}
For the complete conditional tuple, we compute

\[
P_{\mathrm{tuple}} = \frac{H_{\mathrm{tuple}}}{|\hat{G}|},
\]

\[
R_{\mathrm{tuple}} = \frac{H_{\mathrm{tuple}}}{|G|},
\]

and

\[
F_{1,\mathrm{tuple}} =
\frac{2P_{\mathrm{tuple}}R_{\mathrm{tuple}}}
{P_{\mathrm{tuple}}+R_{\mathrm{tuple}}}.
\]

For Synthetic, these metrics are interpreted in the standard way because the gold tuple set is complete by construction. For PubMed and Reddit, we emphasize recall-oriented gold coverage because the real-domain annotations are treated as targeted conditional-causality annotations rather than exhaustive paragraph-level inventories.

\paragraph{Component-level recall.}
We report recall for each tuple component:

\[
\begin{aligned}
R_{E_1} &= \frac{H_{E_1}}{|G|},
&
R_R &= \frac{H_R}{|G|},\\
R_{E_2} &= \frac{H_{E_2}}{|G|},
&
R_C &= \frac{H_C}{|G|}.
\end{aligned}
\]
These metrics identify which part of the conditional tuple is most difficult to recover. This is important because a model may identify entities while failing to recover the relation, or recover the relation while omitting the condition.

\paragraph{Relation extraction recall.}
To compare conditional extraction against standard causal relation extraction, we compute

\[
R_{\mathrm{RE}} = \frac{H_{\mathrm{RE}}}{|G|},
\]

where \(H_{\mathrm{RE}}\) counts matches for \((e_1,r,e_2)\) regardless of the condition. We then define the condition penalty as

\[
\Delta_{\mathrm{condition}}
=
R_{\mathrm{RE}} - R_{\mathrm{tuple}}.
\]

A large \(\Delta_{\mathrm{condition}}\) indicates that the model can recover the ordinary causal relation more often than the full conditional causal tuple. This directly measures the cost of requiring the condition slot.

\paragraph{Directional conditional alignment.}
We also compute two directional alignment measures. The first asks whether the relation triple is recovered when the condition is recovered:

\[
R(\mathrm{RE}\mid C)
=
\frac{H_{\mathrm{tuple}}}{H_C}.
\]

The second asks whether the condition is recovered when the relation triple is recovered:

\[
R(C\mid \mathrm{RE})
=
\frac{H_{\mathrm{tuple}}}{H_{\mathrm{RE}}}.
\]

These two quantities are not identical. \(R(\mathrm{RE}\mid C)\) measures whether recovered conditions are attached to the correct causal relation, while \(R(C\mid \mathrm{RE})\) measures whether successful relation extraction also preserves the condition. The latter is especially important because it tests whether success under standard causal relation extraction carries over to conditional causal extraction.

\paragraph{Paragraph Gold Coverage.}
For paragraph-level evaluation, we define Paragraph Gold Coverage (PGC). For each paragraph \(x_i\), the model receives credit only if it recovers all annotated gold tuples in that paragraph:

\[
\mathrm{PGC}
=
\frac{1}{N}
\sum_{i=1}^{N}
\mathbb{1}
\left[
G_i \subseteq \hat{G}_i
\right].
\]

This is stricter than tuple recall because a paragraph with multiple annotated tuples receives no paragraph-level credit unless all of them are recovered. For PubMed and Reddit, we use the term \textit{Paragraph Gold Coverage} rather than paragraph exact match because the real-domain annotations may not be exhaustive. The metric therefore means that all annotated tuples were recovered, not that every possible causal relation in the paragraph was extracted.

\paragraph{Why the evaluation differs across domains.}
The Synthetic dataset is fully controlled: the gold annotations contain the complete target tuples for each paragraph. Therefore, full precision, recall, F1, and paragraph-level exact recovery are all meaningful. PubMed and Reddit serve a different purpose. They evaluate whether the same extraction framework transfers to real biomedical and informal text. In these settings, surface forms are less controlled, and annotations are best interpreted as known gold targets rather than complete inventories of every possible relation. For this reason, the main real-domain metrics are recall-oriented gold coverage metrics:

\[
\begin{array}{lllll}
R_{\mathrm{tuple}}, &
R_{E_1}, &
R_R, &
R_{E_2}, &
R_C, \\[0.4em]
R_{\mathrm{RE}}, &
\Delta_{\mathrm{condition}}, &
R(\mathrm{RE}\mid C), &
R(C\mid \mathrm{RE}), &
\mathrm{PGC}.
\end{array}
\]

This distinction is important for a fair evaluation. If the real-domain gold set is not exhaustive, then automatic precision can unfairly penalize valid model predictions that were not annotated. Therefore, precision is reported for completeness when applicable, but real-domain conclusions are drawn mainly from gold coverage and condition-preservation metrics.

\paragraph{Evaluation rationale.}
This evaluation design is intended to isolate the core difficulty of conditional causal extraction. A model may perform well under standard relation extraction by recovering \((e_1,r,e_2)\), but still fail the conditional task if it omits \(c\). The comparison between \(R_{\mathrm{RE}}\) and \(R_{\mathrm{tuple}}\), together with \(\Delta_{\mathrm{condition}}\), directly quantifies this gap. The directional scores \(R(\mathrm{RE}\mid C)\) and \(R(C\mid \mathrm{RE})\) further show whether the model aligns conditions with the correct relations or merely extracts tuple components independently. These metrics therefore provide a more faithful evaluation of condition-sensitive causal extraction than ordinary tuple F1 alone.

\section{Extraction Results}
\subsection{Summary Synthetic Extraction Results}
\label{app:summary-synthetic-extraction}
\begin{table*}[t]
\centering
\small
\caption{Summary conditional tuple extraction results on the Synthetic test data. Gray cells indicate the highest score within each model section. Navy cells indicate the highest score overall in each metric column.}
\label{tab:exp1-non-finetuned-synthetic-test-results}
\resizebox{\textwidth}{!}{
\begin{tabular}{llrrrrrrrrrrr}
\toprule
\textbf{Model} & \textbf{Prompt/Type} &
$\boldsymbol{P_{\mathrm{tuple}}}$ &
$\boldsymbol{R_{\mathrm{tuple}}}$ &
$\boldsymbol{F_{1,\mathrm{tuple}}}$ &
$\boldsymbol{F_{1,E_1}}$ &
$\boldsymbol{F_{1,R}}$ &
$\boldsymbol{F_{1,E_2}}$ &
$\boldsymbol{F_{1,C}}$ &
$\boldsymbol{F_{1,\mathrm{RE}}}$ &
$\boldsymbol{\mathrm{EM}_{\mathrm{Paragraph}}}$ &
$\boldsymbol{F_1(\mathrm{RE}\mid C)}$ &
$\boldsymbol{F_1(C\mid \mathrm{RE})}$ \\
\midrule
\multicolumn{13}{l}{\textit{\textbf{Base Large Language Models}}} \\
\midrule
Qwen2.5-32B-Instruct & Two Shot + CoT
& \sectbest{0.7357} & \sectbest{0.6645} & \sectbest{0.6983}
& 0.9151 & \sectbest{0.8224} & 0.9053
& \sectbest{0.7798} & \sectbest{0.8004} & \sectbest{0.5222}
& \sectbest{0.8955} & 0.8724 \\

Kimi-K2 & Two Shot
& 0.5549 & 0.5618 & 0.5584
& \sectbest{0.9409} & 0.7681 & 0.8990
& 0.7130 & 0.7189 & 0.3511
& 0.7831 & 0.7767 \\

Llama-3.3-70B-Instruct & One Shot
& 0.5445 & 0.5371 & 0.5408
& 0.8485 & 0.7219 & 0.8966
& 0.7004 & 0.6790 & 0.3411
& 0.7721 & 0.7965 \\

Phi-4-mini-instruct & Two Shot
& 0.5547 & 0.4864 & 0.5183
& 0.8090 & 0.6795 & 0.7481
& 0.6991 & 0.5937 & 0.2811
& 0.7414 & \sectbest{0.8730} \\

Gemma-3-27b-it & Two Shot + CoT
& 0.6441 & 0.3240 & 0.4311
& 0.6251 & 0.5593 & 0.5848
& 0.5065 & 0.5133 & 0.1733
& 0.8511 & 0.8398 \\

Mistral-7B-Instruct-v0.3 & One Shot
& 0.3673 & 0.3146 & 0.3389
& 0.7495 & 0.6815 & 0.6622
& 0.5451 & 0.4684 & 0.2122
& 0.6217 & 0.7235 \\

GPT-5.4-mini & Two Shot + CoT
& 0.2414 & 0.2408 & 0.2411
& 0.9388 & 0.3255 & \sectbest{0.9375}
& 0.7341 & 0.3196 & 0.1187
& 0.3284 & 0.7543 \\

\midrule
\multicolumn{13}{l}{\textit{\textbf{Fine-tuned Large Language Models}}} \\
\midrule

Phi-4-mini-instruct & Two Shot
& \sectbest{0.6205} & \sectbest{0.5610} & \sectbest{0.5892}
& \sectbest{0.7929} & \sectbest{0.7347} & \sectbest{0.7544}
& \sectbest{0.7562} & \sectbest{0.6371} & \sectbest{0.3611}
& \sectbest{0.7792} & \sectbest{0.9248} \\

\midrule
\multicolumn{13}{l}{\textit{\textbf{Other Fine-tuned Language Models}}} \\
\midrule

FLAN-T5 & Seq2Seq LM
& \overallbest{0.9631} & \overallbest{0.9602} & \overallbest{0.9617}
& \overallbest{0.9938} & \overallbest{0.9792} & \overallbest{0.9934}
& \overallbest{0.9749} & \overallbest{0.9788} & \overallbest{0.9411}
& \overallbest{0.9864} & \overallbest{0.9825} \\

SpanBERT & Encoder LM
& 0.6652 & 0.9491 & 0.7822
& 0.8213 & 0.8016 & 0.8216
& 0.8082 & 0.8016 & 0.5700
& 0.9677 & 0.9758 \\

\bottomrule
\end{tabular}
}
\end{table*}

\subsection{Full Experiment 1 Extraction Results}
\label{app:exp1-extraction-results}
This section reports the complete conditional tuple extraction results across prompting settings and held-out Synthetic test results.
\begin{table*}[t]
\centering
\small
\caption{Results across models and prompting settings.}
\label{tab:all-model-results}
\resizebox{\textwidth}{!}{
\begin{tabular}{llllrrrrrrrrrrr}
\toprule
Model & Type & Prompt & Dataset
& $P_{\mathrm{tuple}}$
& $R_{\mathrm{tuple}}$
& $F_{1,\mathrm{tuple}}$
& $F_{1,E_1}$
& $F_{1,r}$
& $F_{1,E_2}$
& $F_{1,C}$
& $F_{1,\mathrm{RE}}$
& $\mathrm{EM}_{\mathrm{row}}$
& $F_1(\mathrm{RE}\mid C)$
& $F_1(C\mid \mathrm{RE})$ \\
\midrule
Mistral-7B-Instruct-v0.3 & LLM & Zero Shot & Synthetic-All & 0.1554 & 0.1293 & 0.1412 & 0.7888 & 0.2226 & 0.7120 & 0.6743 & 0.1769 & 0.0483 & 0.2093 & 0.7981 \\
Mistral-7B-Instruct-v0.3 & LLM & One Shot & Synthetic-All & 0.3989 & 0.3344 & 0.3639 & 0.7873 & 0.6931 & 0.6840 & 0.5569 & 0.5068 & 0.2307 & 0.6533 & 0.7180 \\
Mistral-7B-Instruct-v0.3 & LLM & Two Shot & Synthetic-All & 0.3704 & 0.2408 & 0.2919 & 0.7088 & 0.5590 & 0.5032 & 0.5362 & 0.3699 & 0.1787 & 0.5444 & 0.7890 \\
Mistral-7B-Instruct-v0.3 & LLM & Two Shot + CoT & Synthetic-All & 0.3704 & 0.2408 & 0.2919 & 0.7088 & 0.5590 & 0.5032 & 0.5362 & 0.3699 & 0.1787 & 0.5444 & 0.7890 \\
Mistral-7B-Instruct-v0.3 & LLM & CoT & Synthetic-All & 0.1973 & 0.1684 & 0.1817 & 0.7955 & 0.2583 & 0.7418 & 0.6853 & 0.2188 & 0.0687 & 0.2652 & 0.8305 \\
\midrule
Llama-3.3-70B-Instruct & LLM & Zero Shot & Synthetic-All & 0.1937 & 0.0500 & 0.0794 & 0.3661 & 0.1061 & 0.3982 & 0.3538 & 0.0905 & 0.0293 & 0.2245 & 0.8778 \\
Llama-3.3-70B-Instruct & LLM & One Shot & Synthetic-All & 0.5392 & 0.5271 & 0.5331 & 0.8591 & 0.7252 & 0.9002 & 0.6998 & 0.6834 & 0.3303 & 0.7618 & 0.7801 \\
Llama-3.3-70B-Instruct & LLM & Two Shot & Synthetic-All & 0.4337 & 0.2900 & 0.3476 & 0.4919 & 0.4328 & 0.5009 & 0.4299 & 0.4070 & 0.1897 & 0.8086 & 0.8540 \\
Llama-3.3-70B-Instruct & LLM & Two Shot + CoT & Synthetic-All & 0.4467 & 0.2941 & 0.3547 & 0.4906 & 0.4288 & 0.4932 & 0.4404 & 0.4008 & 0.2013 & 0.8054 & 0.8849 \\
Llama-3.3-70B-Instruct & LLM & CoT & Synthetic-All & 0.2611 & 0.2703 & 0.2656 & 0.8392 & 0.3642 & 0.9000 & 0.7408 & 0.3304 & 0.1520 & 0.3586 & 0.8039 \\
\midrule
Kimi-K2 & LLM & Zero Shot & Synthetic-All & 0.2019 & 0.1932 & 0.1975 & 0.9204 & 0.2393 & 0.9361 & 0.8006 & 0.2331 & 0.1060 & 0.2466 & 0.8470 \\
Kimi-K2 & LLM & One Shot & Synthetic-All & 0.5015 & 0.4979 & 0.4997 & 0.9411 & 0.6967 & 0.9382 & 0.6824 & 0.6840 & 0.3283 & 0.7322 & 0.7305 \\
Kimi-K2 & LLM & Two Shot & Synthetic-All & 0.5441 & 0.5463 & 0.5452 & 0.9464 & 0.7791 & 0.8864 & 0.7129 & 0.7173 & 0.3630 & 0.7647 & 0.7601 \\
Kimi-K2 & LLM & Two Shot + CoT & Synthetic-All & 0.5470 & 0.4161 & 0.4726 & 0.8226 & 0.7073 & 0.7709 & 0.5853 & 0.6564 & 0.2803 & 0.8075 & 0.7201 \\
Kimi-K2 & LLM & CoT & Synthetic-All & 0.2717 & 0.2125 & 0.2385 & 0.8311 & 0.3027 & 0.8356 & 0.6681 & 0.2993 & 0.1347 & 0.3570 & 0.7967 \\
\midrule
Phi-4-mini-instruct & LLM & Zero Shot & Synthetic-All & 0.1765 & 0.1287 & 0.1488 & 0.6824 & 0.2133 & 0.7096 & 0.5946 & 0.1838 & 0.0523 & 0.2503 & 0.8096 \\
Phi-4-mini-instruct & LLM & One Shot & Synthetic-All & 0.4753 & 0.4048 & 0.4372 & 0.7585 & 0.6484 & 0.7278 & 0.6152 & 0.5429 & 0.2213 & 0.7106 & 0.8053 \\
Phi-4-mini-instruct & LLM & Two Shot & Synthetic-All & 0.5443 & 0.4715 & 0.5053 & 0.8126 & 0.6781 & 0.7474 & 0.6968 & 0.5871 & 0.2810 & 0.7251 & 0.8606 \\
Phi-4-mini-instruct & LLM & Two Shot + CoT & Synthetic-All & 0.4230 & 0.2079 & 0.2787 & 0.5210 & 0.4266 & 0.4722 & 0.4430 & 0.3399 & 0.1273 & 0.6292 & 0.8201 \\
Phi-4-mini-instruct & LLM & CoT & Synthetic-All & 0.2802 & 0.2578 & 0.2685 & 0.7565 & 0.3632 & 0.7708 & 0.6480 & 0.3302 & 0.1087 & 0.4144 & 0.8132 \\
\midrule
Qwen2.5-32B-Instruct & LLM & Zero Shot & Synthetic-All & 0.3161 & 0.2825 & 0.2984 & 0.9024 & 0.3511 & 0.9105 & 0.8067 & 0.3420 & 0.1827 & 0.3699 & 0.8724 \\
Qwen2.5-32B-Instruct & LLM & One Shot & Synthetic-All & 0.6682 & 0.6103 & 0.6379 & 0.9194 & 0.8274 & 0.9220 & 0.7174 & 0.8092 & 0.4420 & 0.8893 & 0.7884 \\
Qwen2.5-32B-Instruct & LLM & Two Shot & Synthetic-All & 0.6757 & 0.6233 & 0.6484 & 0.9302 & 0.8341 & 0.8930 & 0.7572 & 0.7863 & 0.4693 & 0.8563 & 0.8247 \\
Qwen2.5-32B-Instruct & LLM & Two Shot + CoT & Synthetic-All & 0.7254 & 0.6585 & 0.6903 & 0.9275 & 0.8403 & 0.9050 & 0.7876 & 0.8041 & 0.5220 & 0.8764 & 0.8585 \\
Qwen2.5-32B-Instruct & LLM & CoT & Synthetic-All & 0.4673 & 0.4305 & 0.4482 & 0.8993 & 0.5142 & 0.9135 & 0.8295 & 0.4986 & 0.2890 & 0.5402 & 0.8989 \\
\midrule
Gemma-3-27b-it & LLM & Zero Shot & Synthetic-All & 0.0000 & 0.0000 & 0.0000 & 0.0000 & 0.0000 & 0.0000 & 0.0000 & 0.0000 & 0.0000 & 0.0000 & 0.0000 \\
Gemma-3-27b-it & LLM & One Shot & Synthetic-All & 0.5431 & 0.0722 & 0.1275 & 0.1991 & 0.1828 & 0.1971 & 0.1498 & 0.1739 & 0.0320 & 0.8513 & 0.7333 \\
Gemma-3-27b-it & LLM & Two Shot & Synthetic-All & 0.6405 & 0.0346 & 0.0657 & 0.0977 & 0.0907 & 0.0887 & 0.0782 & 0.0809 & 0.0247 & 0.8406 & 0.8127 \\
Gemma-3-27b-it & LLM & Two Shot + CoT & Synthetic-All & 0.6519 & 0.3167 & 0.4263 & 0.6239 & 0.5644 & 0.5814 & 0.5036 & 0.5194 & 0.1690 & 0.8465 & 0.8208 \\
Gemma-3-27b-it & LLM & CoT & Synthetic-All & 0.2751 & 0.0201 & 0.0374 & 0.1157 & 0.0521 & 0.1162 & 0.0857 & 0.0514 & 0.0063 & 0.4370 & 0.7290 \\
\midrule
Mistral-Small-3.1-24B-Instruct-2503 & LLM & Zero Shot & Synthetic-All & 0.0000 & 0.0000 & 0.0000 & 0.0000 & 0.0000 & 0.0000 & 0.0000 & 0.0000 & 0.0000 & 0.0000 & 0.0000 \\
Mistral-Small-3.1-24B-Instruct-2503 & LLM & One Shot & Synthetic-All & 0.0000 & 0.0000 & 0.0000 & 0.0000 & 0.0000 & 0.0000 & 0.0000 & 0.0000 & 0.0000 & 0.0000 & 0.0000 \\
Mistral-Small-3.1-24B-Instruct-2503 & LLM & Two Shot & Synthetic-All & 0.0000 & 0.0000 & 0.0000 & 0.0000 & 0.0000 & 0.0000 & 0.0000 & 0.0000 & 0.0000 & 0.0000 & 0.0000 \\
Mistral-Small-3.1-24B-Instruct-2503 & LLM & Two Shot + CoT & Synthetic-All & 0.0000 & 0.0000 & 0.0000 & 0.0000 & 0.0000 & 0.0000 & 0.0000 & 0.0000 & 0.0000 & 0.0000 & 0.0000 \\
Mistral-Small-3.1-24B-Instruct-2503 & LLM & CoT & Synthetic-All & 0.0000 & 0.0000 & 0.0000 & 0.0000 & 0.0000 & 0.0000 & 0.0000 & 0.0000 & 0.0000 & 0.0000 & 0.0000 \\
\bottomrule
\end{tabular}
}
\end{table*}

\begin{table*}[t]
\centering
\small
\caption{Synthetic test results across models and prompting settings.}
\label{tab:synthetic-test-full-results}
\resizebox{\textwidth}{!}{
\begin{tabular}{llllrrrrrrrrrrr}
\toprule
Model & Type & Prompt & Dataset 
& $P_{\mathrm{tuple}}$ 
& $R_{\mathrm{tuple}}$ 
& $F_{1,\mathrm{tuple}}$ 
& $F_{1,E_1}$ 
& $F_{1,r}$ 
& $F_{1,E_2}$ 
& $F_{1,C}$ 
& $F_{1,\mathrm{RE}}$ 
& $\mathrm{EM}_{\mathrm{para}}$ 
& $F_1(\mathrm{RE}\mid C)$ 
& $F_1(C\mid \mathrm{RE})$ \\
\midrule
Mistral-7B-Instruct-v0.3 & LLM & CoT & Synthetic-Test & 0.1756 & 0.1535 & 0.1638 & 0.7443 & 0.2407 & 0.7052 & 0.6788 & 0.1911 & 0.0611 & 0.2413 & 0.8571 \\
Mistral-7B-Instruct-v0.3 & LLM & One Shot & Synthetic-Test & 0.3673 & 0.3146 & 0.3389 & 0.7495 & 0.6815 & 0.6622 & 0.5451 & 0.4684 & 0.2122 & 0.6217 & 0.7235 \\
Mistral-7B-Instruct-v0.3 & LLM & Two Shot + CoT & Synthetic-Test & 0.3604 & 0.2404 & 0.2884 & 0.6709 & 0.5497 & 0.4986 & 0.5124 & 0.3626 & 0.1900 & 0.5629 & 0.7955 \\
Mistral-7B-Instruct-v0.3 & LLM & Two Shot & Synthetic-Test & 0.3604 & 0.2404 & 0.2884 & 0.6709 & 0.5497 & 0.4986 & 0.5124 & 0.3626 & 0.1900 & 0.5629 & 0.7955 \\
Mistral-7B-Instruct-v0.3 & LLM & Zero Shot & Synthetic-Test & 0.1359 & 0.1134 & 0.1236 & 0.7460 & 0.2110 & 0.6893 & 0.6656 & 0.1473 & 0.0456 & 0.1858 & 0.8391 \\
\midrule
Llama-3.3-70B-Instruct & LLM & CoT & Synthetic-Test & 0.2482 & 0.2621 & 0.2550 & 0.8201 & 0.3487 & 0.8885 & 0.7164 & 0.3151 & 0.1533 & 0.3559 & 0.8092 \\
Llama-3.3-70B-Instruct & LLM & One Shot & Synthetic-Test & 0.5445 & 0.5371 & 0.5408 & 0.8485 & 0.7219 & 0.8966 & 0.7004 & 0.6790 & 0.3411 & 0.7721 & 0.7965 \\
Llama-3.3-70B-Instruct & LLM & Two Shot + CoT & Synthetic-Test & 0.4481 & 0.2928 & 0.3542 & 0.4811 & 0.4212 & 0.4872 & 0.4305 & 0.3986 & 0.2044 & 0.8228 & 0.8887 \\
Llama-3.3-70B-Instruct & LLM & Two Shot & Synthetic-Test & 0.4211 & 0.2796 & 0.3361 & 0.4708 & 0.4185 & 0.4944 & 0.4139 & 0.3919 & 0.1889 & 0.8119 & 0.8575 \\
Llama-3.3-70B-Instruct & LLM & Zero Shot & Synthetic-Test & 0.1976 & 0.0558 & 0.0871 & 0.3961 & 0.1097 & 0.4294 & 0.3835 & 0.0964 & 0.0333 & 0.2270 & 0.9034 \\
\midrule
Kimi-K2 & LLM & CoT & Synthetic-Test & 0.2588 & 0.2029 & 0.2275 & 0.8229 & 0.2944 & 0.8287 & 0.6595 & 0.2901 & 0.1300 & 0.3449 & 0.7842 \\
Kimi-K2 & LLM & One Shot & Synthetic-Test & 0.4964 & 0.4979 & 0.4971 & 0.9372 & 0.6891 & 0.9368 & 0.6784 & 0.6733 & 0.3211 & 0.7327 & 0.7383 \\
Kimi-K2 & LLM & Two Shot + CoT & Synthetic-Test & 0.5554 & 0.4211 & 0.4790 & 0.8092 & 0.6861 & 0.7782 & 0.5867 & 0.6492 & 0.2856 & 0.8165 & 0.7379 \\
Kimi-K2 & LLM & Two Shot & Synthetic-Test & 0.5549 & 0.5618 & 0.5584 & 0.9409 & 0.7681 & 0.8990 & 0.7130 & 0.7189 & 0.3511 & 0.7831 & 0.7767 \\
Kimi-K2 & LLM & Zero Shot & Synthetic-Test & 0.1976 & 0.1888 & 0.1931 & 0.9133 & 0.2306 & 0.9294 & 0.7995 & 0.2254 & 0.1100 & 0.2415 & 0.8569 \\

\midrule
Phi-4-mini-instruct & LLM & CoT & Synthetic-Test & 0.2802 & 0.2511 & 0.2648 & 0.7473 & 0.3494 & 0.7662 & 0.6331 & 0.3206 & 0.1044 & 0.4183 & 0.8261 \\
Phi-4-mini-instruct & LLM & One Shot & Synthetic-Test & 0.4725 & 0.3998 & 0.4332 & 0.7472 & 0.6354 & 0.7176 & 0.6003 & 0.5334 & 0.2067 & 0.7215 & 0.8121 \\
Phi-4-mini-instruct & LLM & Two Shot + CoT & Synthetic-Test & 0.4158 & 0.1957 & 0.2661 & 0.4928 & 0.4110 & 0.4470 & 0.4203 & 0.3252 & 0.1356 & 0.6331 & 0.8182 \\
Phi-4-mini-instruct & LLM & Two Shot & Synthetic-Test & 0.5547 & 0.4864 & 0.5183 & 0.8090 & 0.6795 & 0.7481 & 0.6991 & 0.5937 & 0.2811 & 0.7414 & 0.8730 \\
Phi-4-mini-instruct & LLM & Zero Shot & Synthetic-Test & 0.1663 & 0.1206 & 0.1398 & 0.6759 & 0.1996 & 0.7031 & 0.5687 & 0.1754 & 0.0544 & 0.2459 & 0.7972 \\
\midrule

Qwen2.5-32B-Instruct & LLM & CoT & Synthetic-Test & 0.4660 & 0.4288 & 0.4466 & 0.8999 & 0.5048 & 0.9163 & 0.8266 & 0.4932 & 0.2856 & 0.5403 & 0.9055 \\
Qwen2.5-32B-Instruct & LLM & One Shot & Synthetic-Test & 0.6511 & 0.5951 & 0.6218 & 0.8984 & 0.8062 & 0.9100 & 0.7051 & 0.7835 & 0.4267 & 0.8819 & 0.7936 \\
Qwen2.5-32B-Instruct & LLM & Two Shot + CoT & Synthetic-Test & 0.7357 & 0.6645 & 0.6983 & 0.9151 & 0.8224 & 0.9053 & 0.7798 & 0.8004 & 0.5222 & 0.8955 & 0.8724 \\
Qwen2.5-32B-Instruct & LLM & Two Shot & Synthetic-Test & 0.6790 & 0.6275 & 0.6522 & 0.9167 & 0.8130 & 0.8910 & 0.7519 & 0.7758 & 0.4689 & 0.8674 & 0.8407 \\
Qwen2.5-32B-Instruct & LLM & Zero Shot & Synthetic-Test & 0.3015 & 0.2694 & 0.2846 & 0.8991 & 0.3332 & 0.9086 & 0.7875 & 0.3264 & 0.1744 & 0.3613 & 0.8717 \\

\midrule
Gemma-3-27b-it & LLM & CoT & Synthetic-Test & 0.2338 & 0.0153 & 0.0288 & 0.1024 & 0.0432 & 0.1032 & 0.0728 & 0.0424 & 0.0067 & 0.3956 & 0.6792 \\
Gemma-3-27b-it & LLM & One Shot & Synthetic-Test & 0.5058 & 0.0737 & 0.1287 & 0.2076 & 0.1890 & 0.2009 & 0.1473 & 0.1756 & 0.0367 & 0.8737 & 0.7331 \\
Gemma-3-27b-it & LLM & Two Shot + CoT & Synthetic-Test & 0.6441 & 0.3240 & 0.4311 & 0.6251 & 0.5593 & 0.5848 & 0.5065 & 0.5133 & 0.1733 & 0.8511 & 0.8398 \\
Gemma-3-27b-it & LLM & Two Shot & Synthetic-Test & 0.6780 & 0.0341 & 0.0649 & 0.0852 & 0.0820 & 0.0812 & 0.0714 & 0.0731 & 0.0244 & 0.9091 & 0.8889 \\
Gemma-3-27b-it & LLM & Zero Shot & Synthetic-Test & 0.0000 & 0.0000 & 0.0000 & 0.0000 & 0.0000 & 0.0000 & 0.0000 & 0.0000 & 0.0000 & 0.0000 & 0.0000 \\
\midrule
Mistral-Small-3.1-24B-Instruct-2503 & LLM & CoT & Synthetic-Test & 0.0000 & 0.0000 & 0.0000 & 0.0000 & 0.0000 & 0.0000 & 0.0000 & 0.0000 & 0.0000 & 0.0000 & 0.0000 \\
Mistral-Small-3.1-24B-Instruct-2503 & LLM & One Shot & Synthetic-Test & 0.0000 & 0.0000 & 0.0000 & 0.0000 & 0.0000 & 0.0000 & 0.0000 & 0.0000 & 0.0000 & 0.0000 & 0.0000 \\
Mistral-Small-3.1-24B-Instruct-2503 & LLM & Two Shot + CoT & Synthetic-Test & 0.0000 & 0.0000 & 0.0000 & 0.0000 & 0.0000 & 0.0000 & 0.0000 & 0.0000 & 0.0000 & 0.0000 & 0.0000 \\
Mistral-Small-3.1-24B-Instruct-2503 & LLM & Two Shot & Synthetic-Test & 0.0000 & 0.0000 & 0.0000 & 0.0000 & 0.0000 & 0.0000 & 0.0000 & 0.0000 & 0.0000 & 0.0000 & 0.0000 \\
Mistral-Small-3.1-24B-Instruct-2503 & LLM & Zero Shot & Synthetic-Test & 0.0000 & 0.0000 & 0.0000 & 0.0000 & 0.0000 & 0.0000 & 0.0000 & 0.0000 & 0.0000 & 0.0000 & 0.0000 \\

\bottomrule
\end{tabular}
}
\end{table*}
\begin{table*}[t]
\centering
\small
\caption{Master best-prompt-per-model comparison across datasets.}
\label{tab:exp2-master-best-prompt2}
\resizebox{\textwidth}{!}{%
\begin{tabular}{lllrrrrrrrrrrrrrr}
\toprule
Dataset & Model & Prompt & Acc. & Macro F1 & W-F1 & Sup. F1 & NS F1 & NEE F1 & P1 Acc. & P2 Acc. & P3 Acc. & P4 Acc. & P5 Acc. & OGR & SPS & Rows \\
\midrule
PubMed & Kimi-K2 & Two Shot + CoT & 0.5416 & 0.5310 & 0.5292 & 0.5396 & 0.6311 & 0.4222 & 0.5559 & 0.4324 & 0.1453 & \textbf{0.9294} & 0.6514 & 0.3589 & 0.0306 & 6961 \\
Reddit & Kimi-K2 & Two Shot + CoT & 0.5448 & 0.5340 & 0.5151 & 0.6257 & 0.6029 & 0.3735 & 0.6583 & 0.3594 & 0.1583 & 0.8603 & 0.6974 & 0.1976 & 0.0247 & 7883 \\
Synthetic & Kimi-K2 & Two Shot + CoT & 0.5439 & 0.5388 & 0.5289 & 0.5879 & 0.5749 & 0.4534 & 0.7591 & 0.5357 & 0.1149 & 0.8778 & 0.4328 & 0.2951 & 0.0225 & 37268 \\
PubMed & Llama-3.3-70B-Instruct & Two Shot + CoT & 0.6558 & 0.6539 & 0.6714 & 0.5678 & 0.7188 & 0.6749 & 0.5695 & 0.5863 & 0.5268 & 0.9199 & 0.6807 & 0.2990 & 0.1972 & 6961 \\
Synthetic & Llama-3.3-70B-Instruct & Two Shot + CoT & \textbf{0.7433} & \textbf{0.7389} & \textbf{0.7432} & \textbf{0.7175} & 0.7080 & \textbf{0.7914} & 0.7634 & \textbf{0.8930} & 0.6702 & 0.9024 & 0.4881 & 0.1355 & \textbf{0.2304} & 37268 \\
PubMed & Phi-4-mini-instruct & Zero Shot & 0.6525 & 0.6440 & 0.6652 & 0.5400 & 0.6851 & 0.7070 & 0.5086 & 0.6535 & \textbf{0.6779} & 0.7707 & 0.6535 & 0.2209 & 0.1689 & 6961 \\
Reddit & Phi-4-mini-instruct & Zero Shot & 0.6492 & 0.6455 & 0.6513 & 0.6173 & 0.6614 & 0.6578 & 0.6236 & 0.4994 & 0.6356 & 0.7765 & 0.7150 & 0.1412 & 0.1235 & 7883 \\
Synthetic & Phi-4-mini-instruct & Zero Shot & 0.5927 & 0.5338 & 0.5763 & 0.3216 & 0.6668 & 0.6129 & 0.6942 & 0.7247 & 0.6734 & 0.2310 & 0.6394 & \textbf{0.0364} & 0.0451 & 37268 \\
PubMed & Qwen2.5-32B-Instruct & Two Shot + CoT & 0.6381 & 0.6356 & 0.6512 & 0.5590 & 0.7078 & 0.6399 & 0.5480 & 0.5412 & 0.4689 & 0.9170 & 0.7201 & 0.3061 & 0.1536 & 6961 \\
Reddit & Qwen2.5-32B-Instruct & Zero Shot & 0.6222 & 0.6180 & 0.6027 & 0.6919 & 0.6552 & 0.5069 & 0.7119 & 0.3323 & 0.3739 & 0.8389 & 0.8607 & 0.1019 & 0.0819 & 7883 \\
Synthetic & Qwen2.5-32B-Instruct & Two Shot + CoT & 0.6837 & 0.6740 & 0.6785 & 0.6519 & \textbf{0.7338} & 0.6363 & \textbf{0.7815} & 0.6833 & 0.3783 & 0.7022 & \textbf{0.8732} & 0.0917 & 0.1191 & 37268 \\
\bottomrule
\end{tabular}%
}
\end{table*}

\section{Experiment 2 Probe Evaluation Metrics and Extended Results}
\label{app:exp2-evaluation}
Accuracy is computed as

\[
\mathrm{Acc} = \frac{1}{N}\sum_{i=1}^{N}\mathbf{1}[\hat{y}_i = y_i],
\]

where \(N\) is the number of probe instances. Macro F1 averages F1 equally over the three labels:

\[
\mathrm{MacroF1} = \frac{1}{3}\sum_{\ell \in \mathcal{Y}} F_1(\ell),
\]

where

\[
\mathcal{Y} =
\left\{
\begin{array}{l}
\text{Supported},\\
\text{Not Supported},\\
\text{Not Enough Evidence}
\end{array}
\right\}.
\]

Weighted F1 adjusts this average by the number of examples belonging to each label:

\[
\mathrm{WeightedF1} = \sum_{\ell \in \mathcal{Y}} \frac{n_\ell}{N}F_1(\ell).
\]

We also report probe-level accuracy for each probe type:

\[
\mathrm{P_kAcc} = \frac{1}{N_k}\sum_{i=1}^{N_k}\mathbf{1}[\hat{y}_{ik}=y_{ik}],
\]

where \(k \in \{1,2,3,4,5\}\). These scores show which type of perturbation is easiest or hardest for the model.

To directly measure overgeneralization, we define the Overgeneralization Rate (OGR). OGR measures how often the model predicts \textit{Supported} for probes where the condition has been replaced, removed, or contradicted. These are probes \(p_{i2}\), \(p_{i3}\), and \(p_{i5}\):

{\scriptsize
\[
\mathrm{OGR}
=
\frac{
\begin{array}{l}
\sum_i \mathbf{1}[\hat{y}_{i2}=\text{Supported}]
+
\sum_i \mathbf{1}[\hat{y}_{i3}=\text{Supported}]
\\
\quad+
\sum_i \mathbf{1}[\hat{y}_{i5}=\text{Supported}]
\end{array}
}{
N_2 + N_3 + N_5
}.
\]
}
Lower OGR is better. A high OGR means that the model accepts a causal claim even when the condition has been changed or removed.

Finally, we define the Scope Preservation Score (SPS). SPS measures whether the model preserves the full expected behavior across the probe set for the same underlying tuple. A tuple receives credit only if all five probes are classified correctly:

\[
\begin{aligned}
\mathrm{SPS}
&=
\frac{1}{M}\sum_{i=1}^{M}
\mathbf{1}\big[
\hat{y}_{i1}=y_{i1}
\land
\hat{y}_{i2}=y_{i2} \\
&\qquad\land
\hat{y}_{i3}=y_{i3}
\land
\hat{y}_{i4}=y_{i4} \\
&\qquad\land
\hat{y}_{i5}=y_{i5}
\big].
\end{aligned}
\]

where \(M\) is the number of complete tuple-level probe groups. SPS is stricter than ordinary accuracy because a model must get the supported claim, the relation contradiction, the condition replacement, the condition removal, and the condition contradiction correct for the same tuple. Thus, SPS captures whether the model understands the full scope pattern, not just isolated labels.

\begin{table*}[htbp]
\centering
\scriptsize
\setlength{\tabcolsep}{2pt}
\renewcommand{\arraystretch}{1.05}
\caption{PubMed results (Exp2b).}
\label{tab:pubmed_results}
\resizebox{\linewidth}{!}{%
\begin{tabular}{llrrrrrrrrrrrrrrrrrr}
\toprule
Model & Prompt & Rows & Acc. & Macro P & Macro R & Macro F1 & Wtd. P & Wtd. R & Wtd. F1 & Supp. F1 & Not Supp. F1 & NEE F1 & P1 Acc. & P2 Acc. & P3 Acc. & P4 Acc. & P5 Acc. & OGR & SPS \\
\midrule
Kimi-K2 & Zero Shot & 6961 & 0.4916 & 0.5346 & 0.5623 & 0.4779 & 0.5713 & 0.4916 & 0.4713 & 0.5103 & 0.5939 & 0.3294 & 0.5301 & 0.3923 & 0.0601 & 0.9090 & 0.5734 & 0.4032 & 0.0102 \\
Kimi-K2 & One Shot & 6961 & 0.4999 & 0.5869 & 0.5743 & 0.4725 & 0.6330 & 0.4999 & 0.4626 & 0.5212 & 0.6146 & 0.2818 & 0.5595 & 0.2770 & 0.0666 & 0.9389 & 0.6650 & 0.4042 & 0.0138 \\
Kimi-K2 & Two Shot & 6961 & 0.5146 & 0.5759 & 0.5861 & 0.4994 & 0.6171 & 0.5146 & 0.4923 & 0.5344 & 0.6026 & 0.3612 & 0.5552 & 0.4023 & 0.0780 & 0.9367 & 0.6077 & 0.3746 & 0.0160 \\
Kimi-K2 & Two Shot + CoT & 6961 & 0.5416 & 0.6085 & 0.6073 & 0.5310 & 0.6551 & 0.5416 & 0.5292 & 0.5396 & 0.6311 & 0.4222 & 0.5559 & 0.4324 & 0.1453 & 0.9294 & 0.6514 & 0.3589 & 0.0306 \\
Llama-3.3-70B-Instruct & Zero Shot & 6961 & 0.6495 & 0.6923 & 0.6948 & 0.6469 & 0.7481 & 0.6495 & 0.6615 & 0.5752 & 0.6996 & 0.6659 & 0.5616 & 0.5791 & 0.5032 & 0.9170 & 0.6908 & 0.2801 & 0.1739 \\
Llama-3.3-70B-Instruct & One Shot & 6961 & 0.6195 & 0.6844 & 0.6706 & 0.6184 & 0.7457 & 0.6195 & 0.6339 & 0.5424 & 0.6971 & 0.6157 & 0.5616 & 0.4968 & 0.4660 & 0.9214 & 0.6564 & 0.3457 & 0.1492 \\
Llama-3.3-70B-Instruct & Two Shot & 6961 & 0.6470 & 0.7027 & 0.6956 & 0.6459 & 0.7636 & 0.6470 & 0.6626 & 0.5643 & 0.7082 & 0.6652 & 0.5573 & 0.5934 & 0.4882 & 0.9338 & 0.6671 & 0.3133 & 0.1892 \\
Llama-3.3-70B-Instruct & Two Shot + CoT & 6961 & 0.6558 & 0.7045 & 0.7005 & 0.6539 & 0.7644 & 0.6558 & 0.6714 & 0.5678 & 0.7188 & 0.6749 & 0.5695 & 0.5863 & 0.5268 & 0.9199 & 0.6807 & 0.2990 & 0.1972 \\
Phi-4-mini-instruct & Zero Shot & 6961 & 0.6525 & 0.6679 & 0.6725 & 0.6440 & 0.7193 & 0.6525 & 0.6652 & 0.5400 & 0.6851 & 0.7070 & 0.5086 & 0.6535 & 0.6779 & 0.7707 & 0.6535 & 0.2209 & 0.1689 \\
Phi-4-mini-instruct & One Shot & 6961 & 0.4824 & 0.6554 & 0.5614 & 0.4439 & 0.7222 & 0.4824 & 0.4351 & 0.4873 & 0.6335 & 0.2110 & 0.5279 & 0.1303 & 0.1059 & 0.9483 & 0.7072 & 0.4803 & 0.0160 \\
Phi-4-mini-instruct & Two Shot & 6961 & 0.5558 & 0.6769 & 0.6231 & 0.5500 & 0.7425 & 0.5558 & 0.5567 & 0.5175 & 0.6396 & 0.4931 & 0.5050 & 0.3228 & 0.3379 & 0.9527 & 0.6671 & 0.4032 & 0.0793 \\
Phi-4-mini-instruct & Two Shot + CoT & 6961 & 0.5814 & 0.6715 & 0.6378 & 0.5710 & 0.7270 & 0.5814 & 0.5741 & 0.5558 & 0.6553 & 0.5020 & 0.5602 & 0.3307 & 0.3479 & 0.9141 & 0.7595 & 0.3066 & 0.0844 \\
Qwen2.5-32B-Instruct & Zero Shot & 6961 & 0.5742 & 0.6375 & 0.6282 & 0.5643 & 0.6752 & 0.5742 & 0.5568 & 0.6009 & 0.6222 & 0.4697 & 0.5681 & 0.2963 & 0.3443 & 0.8930 & 0.7745 & 0.2164 & 0.0575 \\
Qwen2.5-32B-Instruct & One Shot & 6961 & 0.5838 & 0.6548 & 0.6415 & 0.5743 & 0.7077 & 0.5838 & 0.5784 & 0.5537 & 0.6701 & 0.4990 & 0.5552 & 0.3651 & 0.3314 & 0.9243 & 0.7487 & 0.3219 & 0.0764 \\
Qwen2.5-32B-Instruct & Two Shot & 6961 & 0.5868 & 0.6543 & 0.6456 & 0.5790 & 0.7056 & 0.5868 & 0.5823 & 0.5625 & 0.6610 & 0.5134 & 0.5408 & 0.3873 & 0.3386 & 0.9338 & 0.7394 & 0.3097 & 0.0830 \\
Qwen2.5-32B-Instruct & Two Shot + CoT & 6961 & 0.6381 & 0.6921 & 0.6854 & 0.6356 & 0.7512 & 0.6381 & 0.6512 & 0.5590 & 0.7078 & 0.6399 & 0.5480 & 0.5412 & 0.4689 & 0.9170 & 0.7201 & 0.3061 & 0.1536 \\
\bottomrule
\end{tabular}%
}
\end{table*}

\begin{table*}[htbp]
\centering
\scriptsize
\setlength{\tabcolsep}{2pt}
\renewcommand{\arraystretch}{1.05}
\caption{Reddit results (Exp2c).}
\label{tab:reddit_results}
\resizebox{\linewidth}{!}{%
\begin{tabular}{llrrrrrrrrrrrrrrrrrr}
\toprule
Model & Prompt & Rows & Acc. & Macro P & Macro R & Macro F1 & Wtd. P & Wtd. R & Wtd. F1 & Supp. F1 & Not Supp. F1 & NEE F1 & P1 Acc. & P2 Acc. & P3 Acc. & P4 Acc. & P5 Acc. & OGR & SPS \\
\midrule
Kimi-K2 & Zero Shot & 7883 & 0.5299 & 0.5464 & 0.5847 & 0.5170 & 0.5630 & 0.5299 & 0.4994 & 0.6020 & 0.5974 & 0.3517 & 0.6343 & 0.3701 & 0.1185 & 0.8493 & 0.6866 & 0.2196 & 0.0188 \\
Kimi-K2 & One Shot & 7883 & 0.5103 & 0.5495 & 0.5714 & 0.4737 & 0.5678 & 0.5103 & 0.4471 & 0.6019 & 0.6047 & 0.2143 & 0.6803 & 0.1791 & 0.0763 & 0.8661 & 0.7604 & 0.2327 & 0.0065 \\
Kimi-K2 & Two Shot & 7883 & 0.5327 & 0.5527 & 0.5906 & 0.5218 & 0.5672 & 0.5327 & 0.5012 & 0.6211 & 0.5888 & 0.3554 & 0.6349 & 0.3625 & 0.1293 & 0.8700 & 0.6765 & 0.2070 & 0.0247 \\
Kimi-K2 & Two Shot + CoT & 7883 & 0.5448 & 0.5684 & 0.5990 & 0.5340 & 0.5844 & 0.5448 & 0.5151 & 0.6257 & 0.6029 & 0.3735 & 0.6583 & 0.3594 & 0.1583 & 0.8603 & 0.6974 & 0.1976 & 0.0247 \\
Llama-3.3-70B-Instruct & Two Shot & 7883 & 0.6897 & 0.7007 & 0.7239 & 0.6871 & 0.7349 & 0.6897 & 0.6909 & 0.6688 & 0.7138 & 0.6786 & 0.6728 & 0.6280 & 0.4918 & 0.8889 & 0.7730 & 0.1578 & 0.1806 \\
Llama-3.3-70B-Instruct & Two Shot + CoT & 7883 & 0.6999 & 0.7022 & 0.7220 & 0.6964 & 0.7289 & 0.6999 & 0.7002 & 0.6778 & 0.7173 & 0.6940 & 0.7011 & 0.6696 & 0.5088 & 0.8291 & 0.7945 & 0.1171 & 0.1988 \\
Phi-4-mini-instruct & Zero Shot & 7883 & 0.6492 & 0.6494 & 0.6711 & 0.6455 & 0.6778 & 0.6492 & 0.6513 & 0.6173 & 0.6614 & 0.6578 & 0.6236 & 0.4994 & 0.6356 & 0.7765 & 0.7150 & 0.1412 & 0.1235 \\
Phi-4-mini-instruct & One Shot & 7883 & 0.4886 & 0.6109 & 0.5553 & 0.4570 & 0.6614 & 0.4886 & 0.4445 & 0.5176 & 0.5871 & 0.2663 & 0.5858 & 0.1740 & 0.1375 & 0.8772 & 0.6803 & 0.3539 & 0.0169 \\
Phi-4-mini-instruct & Two Shot & 7883 & 0.5569 & 0.6479 & 0.6161 & 0.5533 & 0.7054 & 0.5569 & 0.5591 & 0.5255 & 0.6026 & 0.5319 & 0.5561 & 0.4155 & 0.3260 & 0.9019 & 0.5952 & 0.3510 & 0.0669 \\
Phi-4-mini-instruct & Two Shot + CoT & 7883 & 0.5772 & 0.6412 & 0.6146 & 0.5679 & 0.6803 & 0.5772 & 0.5661 & 0.5767 & 0.6218 & 0.5054 & 0.6532 & 0.3600 & 0.3392 & 0.7953 & 0.7446 & 0.2022 & 0.0695 \\
Qwen2.5-32B-Instruct & Zero Shot & 7883 & 0.6222 & 0.6828 & 0.6594 & 0.6180 & 0.7023 & 0.6222 & 0.6027 & 0.6919 & 0.6552 & 0.5069 & 0.7119 & 0.3323 & 0.3739 & 0.8389 & 0.8607 & 0.1019 & 0.0819 \\
Qwen2.5-32B-Instruct & One Shot & 7883 & 0.5848 & 0.6661 & 0.6276 & 0.5676 & 0.6921 & 0.5848 & 0.5493 & 0.6559 & 0.6443 & 0.4027 & 0.7068 & 0.2074 & 0.3090 & 0.8343 & 0.8739 & 0.1364 & 0.0513 \\
Qwen2.5-32B-Instruct & Two Shot & 7883 & 0.6024 & 0.6712 & 0.6486 & 0.5919 & 0.6985 & 0.6024 & 0.5764 & 0.6665 & 0.6482 & 0.4609 & 0.6904 & 0.2850 & 0.3317 & 0.8713 & 0.8417 & 0.1410 & 0.0669 \\
Qwen2.5-32B-Instruct & Two Shot + CoT & 7883 & 0.6849 & 0.6974 & 0.7111 & 0.6810 & 0.7279 & 0.6849 & 0.6844 & 0.6641 & 0.7068 & 0.6720 & 0.6885 & 0.5631 & 0.5290 & 0.8376 & 0.8108 & 0.1305 & 0.1689 \\
\bottomrule
\end{tabular}%
}
\end{table*}

\begin{table*}[htbp]
\centering
\scriptsize
\setlength{\tabcolsep}{2pt}
\renewcommand{\arraystretch}{1.05}
\caption{Synthetic results (Exp2a).}
\label{tab:synthetic_results}
\resizebox{\linewidth}{!}{%
\begin{tabular}{llrrrrrrrrrrrrrrrrrr}
\toprule
Model & Prompt & Rows & Acc. & Macro P & Macro R & Macro F1 & Wtd. P & Wtd. R & Wtd. F1 & Supp. F1 & Not Supp. F1 & NEE F1 & P1 Acc. & P2 Acc. & P3 Acc. & P4 Acc. & P5 Acc. & OGR & SPS \\
\midrule
Kimi-K2 & One Shot & 37268 & 0.4955 & 0.5800 & 0.5643 & 0.4679 & 0.6136 & 0.4955 & 0.4480 & 0.5674 & 0.5665 & 0.2698 & 0.7661 & 0.2799 & 0.0440 & 0.9077 & 0.4807 & 0.3550 & 0.0039 \\
Kimi-K2 & Two Shot & 37268 & 0.4916 & 0.5425 & 0.5558 & 0.4776 & 0.5739 & 0.4916 & 0.4660 & 0.5356 & 0.5557 & 0.3417 & 0.7471 & 0.3881 & 0.0649 & 0.8756 & 0.3833 & 0.3859 & 0.0058 \\
Kimi-K2 & Two Shot + CoT & 37268 & 0.5439 & 0.5819 & 0.5997 & 0.5388 & 0.6100 & 0.5439 & 0.5289 & 0.5879 & 0.5749 & 0.4534 & 0.7591 & 0.5357 & 0.1149 & 0.8778 & 0.4328 & 0.2951 & 0.0225 \\
Llama-3.3-70B-Instruct & Zero Shot & 37268 & 0.6350 & 0.6358 & 0.6599 & 0.6304 & 0.6701 & 0.6350 & 0.6398 & 0.5836 & 0.6624 & 0.6452 & 0.7559 & 0.7516 & 0.4194 & 0.7838 & 0.4649 & 0.2357 & 0.0896 \\
Llama-3.3-70B-Instruct & One Shot & 37268 & 0.6479 & 0.6777 & 0.6920 & 0.6454 & 0.7261 & 0.6479 & 0.6563 & 0.5907 & 0.6370 & 0.7085 & 0.7387 & 0.7236 & 0.5268 & 0.9122 & 0.3387 & 0.3176 & 0.1046 \\
Llama-3.3-70B-Instruct & Two Shot & 37268 & 0.6821 & 0.6942 & 0.7238 & 0.6802 & 0.7332 & 0.6821 & 0.6862 & 0.6504 & 0.6484 & 0.7419 & 0.7513 & 0.7617 & 0.5528 & 0.9317 & 0.4134 & 0.2372 & 0.1339 \\
Llama-3.3-70B-Instruct & Two Shot + CoT & 37268 & 0.7433 & 0.7374 & 0.7699 & 0.7389 & 0.7658 & 0.7433 & 0.7432 & 0.7175 & 0.7080 & 0.7914 & 0.7634 & 0.8930 & 0.6702 & 0.9024 & 0.4881 & 0.1355 & 0.2304 \\
Phi-4-mini-instruct & Zero Shot & 37268 & 0.5927 & 0.5805 & 0.5323 & 0.5338 & 0.5908 & 0.5927 & 0.5763 & 0.3216 & 0.6668 & 0.6129 & 0.6942 & 0.7247 & 0.6734 & 0.2310 & 0.6394 & 0.0364 & 0.0451 \\
Phi-4-mini-instruct & One Shot & 37268 & 0.4797 & 0.5898 & 0.5214 & 0.4732 & 0.6369 & 0.4797 & 0.4723 & 0.4777 & 0.5091 & 0.4326 & 0.6082 & 0.3503 & 0.2110 & 0.7291 & 0.5005 & 0.3124 & 0.0188 \\
Phi-4-mini-instruct & Two Shot & 37268 & 0.5021 & 0.5766 & 0.5391 & 0.5040 & 0.6283 & 0.5021 & 0.5164 & 0.4425 & 0.5041 & 0.5655 & 0.5496 & 0.5122 & 0.3229 & 0.7232 & 0.4035 & 0.3716 & 0.0312 \\
Phi-4-mini-instruct & Two Shot + CoT & 37268 & 0.4970 & 0.6146 & 0.5107 & 0.4661 & 0.6449 & 0.4970 & 0.4564 & 0.5148 & 0.5738 & 0.3099 & 0.7840 & 0.2386 & 0.1340 & 0.5790 & 0.7499 & 0.1543 & 0.0160 \\
Qwen2.5-32B-Instruct & Zero Shot & 37268 & 0.6012 & 0.6672 & 0.6077 & 0.5838 & 0.6841 & 0.6012 & 0.5785 & 0.6101 & 0.6633 & 0.4779 & 0.8133 & 0.4073 & 0.2497 & 0.6404 & 0.8953 & 0.0963 & 0.0397 \\
Qwen2.5-32B-Instruct & One Shot & 37268 & 0.6685 & 0.6806 & 0.6607 & 0.6520 & 0.6997 & 0.6685 & 0.6618 & 0.6028 & 0.7244 & 0.6288 & 0.7936 & 0.5970 & 0.4173 & 0.6217 & 0.9128 & 0.0947 & 0.1022 \\
Qwen2.5-32B-Instruct & Two Shot & 37268 & 0.6407 & 0.6563 & 0.6460 & 0.6289 & 0.6741 & 0.6407 & 0.6315 & 0.6154 & 0.6971 & 0.5740 & 0.7772 & 0.5510 & 0.3488 & 0.6725 & 0.8541 & 0.1097 & 0.0821 \\
Qwen2.5-32B-Instruct & Two Shot + CoT & 37268 & 0.6837 & 0.6873 & 0.6868 & 0.6740 & 0.7031 & 0.6837 & 0.6785 & 0.6519 & 0.7338 & 0.6363 & 0.7815 & 0.6833 & 0.3783 & 0.7022 & 0.8732 & 0.0917 & 0.1191 \\
\bottomrule
\end{tabular}%
}
\end{table*}

\newpage
\begin{table*}[t]
\centering
\small
\caption{Master model comparison across datasets and model classes.}
\label{tab:exp2-master-comparison-with-class}
\resizebox{\textwidth}{!}{%
\begin{tabular}{llllrrrrrrrrrrrrr}
\toprule
\textbf{Dataset} & \textbf{Model} & \textbf{Class} & \textbf{Prompt} &
$\boldsymbol{\mathrm{Acc.}}$ &
$\boldsymbol{\mathrm{Macro\ F1}}$ &
$\boldsymbol{\mathrm{W\mbox{-}F1}}$ &
$\boldsymbol{\mathrm{Sup.\ F1}}$ &
$\boldsymbol{\mathrm{NS\ F1}}$ &
$\boldsymbol{\mathrm{NEE\ F1}}$ &
$\boldsymbol{\mathrm{P1\ Acc.}}$ &
$\boldsymbol{\mathrm{P2\ Acc.}}$ &
$\boldsymbol{\mathrm{P3\ Acc.}}$ &
$\boldsymbol{\mathrm{P4\ Acc.}}$ &
$\boldsymbol{\mathrm{P5\ Acc.}}$ &
$\boldsymbol{\mathrm{OGR}}$ &
$\boldsymbol{\mathrm{SPS}}$ \\
\midrule

\multicolumn{17}{l}{\textit{\textbf{Base Large Language Models}}} \\
\midrule
PubMed & Kimi-K2 & Base LLM & Two Shot + CoT & 0.5416 & 0.5310 & 0.5292 & 0.5396 & 0.6311 & 0.4222 & 0.5559 & 0.4324 & 0.1453 & 0.9294 & 0.6514 & 0.3589 & 0.0306 \\
PubMed & Llama-3.3-70B-Instruct & Base LLM & Two Shot + CoT & 0.6558 & 0.6539 & 0.6714 & 0.5678 & 0.7188 & 0.6749 & 0.5695 & 0.5863 & 0.5268 & 0.9199 & 0.6807 & 0.2990 & 0.1972 \\
PubMed & Phi-4-mini-instruct & Base LLM & Zero Shot & 0.6525 & 0.6440 & 0.6652 & 0.5400 & 0.6851 & 0.7070 & 0.5086 & 0.6535 & 0.6779 & 0.7707 & 0.6535 & 0.2209 & 0.1689 \\
PubMed & Qwen2.5-32B-Instruct & Base LLM & Two Shot + CoT & 0.6381 & 0.6356 & 0.6512 & 0.5590 & 0.7078 & 0.6399 & 0.5480 & 0.5412 & 0.4689 & 0.9170 & 0.7201 & 0.3061 & 0.1536 \\

\midrule
Reddit & Kimi-K2 & Base LLM & Two Shot + CoT & 0.5448 & 0.5340 & 0.5151 & 0.6257 & 0.6029 & 0.3735 & 0.6583 & 0.3594 & 0.1583 & 0.8603 & 0.6974 & 0.1976 & 0.0247 \\
Reddit & Llama-3.3-70B-Instruct & Base LLM & Two Shot + CoT & 0.6999 & 0.6964 & 0.7002 & 0.6778 & 0.7173 & 0.6940 & 0.7011 & 0.6696 & 0.5088 & 0.8291 & 0.7945 & 0.1171 & 0.1988 \\
Reddit & Phi-4-mini-instruct & Base LLM & Zero Shot & 0.6492 & 0.6455 & 0.6513 & 0.6173 & 0.6614 & 0.6578 & 0.6236 & 0.4994 & 0.6356 & 0.7765 & 0.7150 & 0.1412 & 0.1235 \\
Reddit & Qwen2.5-32B-Instruct & Base LLM & Zero Shot & 0.6222 & 0.6180 & 0.6027 & 0.6919 & 0.6552 & 0.5069 & 0.7119 & 0.3323 & 0.3739 & 0.8389 & 0.8607 & 0.1019 & 0.0819 \\

\midrule
Synthetic-Test & Kimi-K2 & Base LLM & Two Shot + CoT & 0.5481 & 0.5427 & 0.5326 & 0.5928 & 0.5820 & 0.4532 & 0.7621 & 0.5354 & 0.1154 & 0.8839 & 0.4445 & 0.2924 & 0.0227 \\
Synthetic-Test & Llama-3.3-70B-Instruct & Base LLM & Two Shot + CoT & 0.7417 & 0.7374 & 0.7421 & 0.7142 & 0.7134 & 0.7847 & 0.7693 & 0.8851 & 0.6606 & 0.8969 & 0.4971 & 0.1366 & 0.2365 \\
Synthetic-Test & Phi-4-mini-instruct & Base LLM & Zero Shot & 0.5930 & 0.5353 & 0.5774 & 0.3253 & 0.6651 & 0.6155 & 0.6927 & 0.7327 & 0.6739 & 0.2366 & 0.6285 & 0.0383 & 0.0486 \\
Synthetic-Test & Qwen2.5-32B-Instruct & Base LLM & Two Shot + CoT & 0.6890 & 0.6791 & 0.6841 & 0.6543 & 0.7393 & 0.6438 & 0.7915 & 0.7029 & 0.3715 & 0.7076 & 0.8717 & 0.0928 & 0.1216 \\

\midrule
All Synthetic & Kimi-K2 & Base LLM & Two Shot + CoT & 0.5439 & 0.5388 & 0.5289 & 0.5879 & 0.5749 & 0.4534 & 0.7591 & 0.5357 & 0.1149 & 0.8778 & 0.4328 & 0.2951 & 0.0225 \\
All Synthetic & Llama-3.3-70B-Instruct & Base LLM & Two Shot + CoT & 0.7433 & 0.7389 & 0.7432 & 0.7175 & 0.7080 & 0.7914 & 0.7634 & 0.8930 & 0.6702 & 0.9024 & 0.4881 & 0.1355 & 0.2304 \\
All Synthetic & Phi-4-mini-instruct & Base LLM & Zero Shot & 0.5927 & 0.5338 & 0.5763 & 0.3216 & 0.6668 & 0.6129 & 0.6942 & 0.7247 & 0.6734 & 0.2310 & 0.6394 & 0.0364 & 0.0451 \\
All Synthetic & Qwen2.5-32B-Instruct & Base LLM & Two Shot + CoT & 0.6837 & 0.6740 & 0.6785 & 0.6519 & 0.7338 & 0.6363 & 0.7815 & 0.6833 & 0.3783 & 0.7022 & 0.8732 & 0.0917 & 0.1191 \\

\midrule
\multicolumn{17}{l}{\textit{\textbf{Fine-tuned Large Language Models}}} \\
\midrule
PubMed & Phi-4-mini-instruct & FT-LLM (P) & Zero Shot & 0.7539 & 0.7577 & 0.7549 & 0.7712 & 0.7440 & 0.7579 & 0.7901 & 0.5333 & 0.7180 & 0.7627 & 0.9656 & 0.0057 & 0.2620 \\
PubMed & Phi-4-mini-instruct & FT-LLM (P+T) & Zero Shot & 0.4051 & 0.2064 & 0.2392 & 0.0454 & 0.5739 & 0.0000 & 0.9971 & 0.0000 & 0.0000 & 0.0233 & 0.9993 & 0.0002 & 0.0000 \\

Reddit & Phi-4-mini-instruct & FT-LLM (P) & Zero Shot & 0.5889 & 0.5944 & 0.5835 & 0.6473 & 0.6302 & 0.5058 & 0.7125 & 0.2276 & 0.4615 & 0.6959 & 0.8499 & 0.0679 & 0.0757 \\
Reddit & Phi-4-mini-instruct & FT-LLM (P+T) & Zero Shot & 0.3967 & 0.2132 & 0.2464 & 0.0529 & 0.5712 & 0.0156 & 0.9634 & 0.0057 & 0.0101 & 0.0273 & 0.9660 & 0.0006 & 0.0000 \\

Synthetic-Test & Phi-4-mini-instruct & FT-LLM (P) & Zero Shot & 0.8948 & 0.8959 & 0.8959 & 0.8962 & 0.8751 & 0.9165 & 0.8445 & 0.7737 & 0.9225 & 0.9402 & 0.9933 & 0.0009 & 0.5728 \\
Synthetic-Test & Phi-4-mini-instruct & FT-LLM (P+T) & Zero Shot & 0.4056 & 0.2080 & 0.2404 & 0.0462 & 0.5738 & 0.0040 & 1.0000 & 0.0000 & 0.0040 & 0.0237 & 0.9996 & 0.0000 & 0.0000 \\

\midrule
\multicolumn{17}{l}{\textit{\textbf{Other Fine-tuned Language Models}}} \\
\midrule
PubMed & BERT & Encoder (P) & -- & 0.8917 & 0.8830 & 0.8914 & 0.8419 & 0.8865 & 0.9208 & 0.9176 & 0.8812 & 0.9155 & 0.7868 & 0.9556 & 0.0131 & 0.6521 \\
PubMed & PubMedBERT & Encoder (P) & -- & 0.9392 & 0.9359 & 0.9390 & 0.9208 & 0.9307 & 0.9561 & 0.9169 & 0.9778 & 0.9807 & 0.9090 & 0.9112 & 0.0084 & 0.7666 \\
PubMed & BioBERT & Encoder (P) & -- & 0.8914 & 0.8845 & 0.8913 & 0.8513 & 0.8833 & 0.9189 & 0.9248 & 0.8933 & 0.9155 & 0.8042 & 0.9177 & 0.0167 & 0.6722 \\
PubMed & SciBERT & Encoder (P) & -- & 0.9358 & 0.9311 & 0.9356 & 0.9088 & 0.9266 & 0.9578 & 0.9377 & 0.9549 & 0.9628 & 0.8741 & 0.9485 & 0.0067 & 0.7717 \\
PubMed & BERT & Encoder (P+T) & -- & 0.8536 & 0.8440 & 0.8535 & 0.7970 & 0.8521 & 0.8827 & 0.8725 & 0.8533 & 0.8712 & 0.7700 & 0.8998 & 0.0344 & 0.6084 \\
PubMed & PubMedBERT & Encoder (P+T) & -- & 0.9206 & 0.9175 & 0.9204 & 0.9034 & 0.9132 & 0.9360 & 0.8947 & 0.9413 & 0.9599 & 0.9119 & 0.8948 & 0.0198 & 0.7373 \\
PubMed & BioBERT & Encoder (P+T) & -- & 0.8576 & 0.8506 & 0.8579 & 0.8144 & 0.8535 & 0.8838 & 0.8918 & 0.8475 & 0.8561 & 0.7999 & 0.8919 & 0.0365 & 0.6550 \\
PubMed & SciBERT & Encoder (P+T) & -- & 0.9180 & 0.9134 & 0.9178 & 0.8915 & 0.9114 & 0.9372 & 0.9019 & 0.9442 & 0.9535 & 0.8908 & 0.8991 & 0.0205 & 0.7466 \\

\midrule
Reddit & BERT & Encoder (P) & -- & 0.5722 & 0.4848 & 0.5258 & 0.2870 & 0.6526 & 0.5149 & 0.9868 & 0.2989 & 0.4023 & 0.1715 & 0.9899 & 0.0057 & 0.1362 \\
Reddit & PubMedBERT & Encoder (P) & -- & 0.6019 & 0.5361 & 0.5683 & 0.3805 & 0.6687 & 0.5590 & 0.9823 & 0.3739 & 0.4073 & 0.2430 & 0.9924 & 0.0069 & 0.2144 \\
Reddit & BioBERT & Encoder (P) & -- & 0.6041 & 0.5307 & 0.5696 & 0.3432 & 0.6695 & 0.5795 & 0.9792 & 0.3903 & 0.4439 & 0.2118 & 0.9836 & 0.0055 & 0.1879 \\
Reddit & SciBERT & Encoder (P) & -- & 0.6411 & 0.5911 & 0.6186 & 0.4585 & 0.6914 & 0.6235 & 0.9874 & 0.4199 & 0.4931 & 0.3054 & 0.9899 & 0.0061 & 0.2560 \\
Reddit & BERT & Encoder (P+T) & -- & 0.5707 & 0.5025 & 0.5421 & 0.3110 & 0.6395 & 0.5568 & 0.8682 & 0.4168 & 0.4779 & 0.2040 & 0.8758 & 0.0259 & 0.1494 \\
Reddit & PubMedBERT & Encoder (P+T) & -- & 0.5715 & 0.5019 & 0.5311 & 0.3612 & 0.6523 & 0.4923 & 0.9716 & 0.3090 & 0.3480 & 0.2404 & 0.9786 & 0.0216 & 0.1942 \\
Reddit & BioBERT & Encoder (P+T) & -- & 0.5838 & 0.5134 & 0.5461 & 0.3554 & 0.6566 & 0.5282 & 0.9685 & 0.3506 & 0.3947 & 0.2196 & 0.9748 & 0.0036 & 0.1917 \\
Reddit & SciBERT & Encoder (P+T) & -- & 0.6160 & 0.5603 & 0.5884 & 0.4242 & 0.6763 & 0.5803 & 0.9735 & 0.3928 & 0.4395 & 0.2807 & 0.9836 & 0.0097 & 0.2465 \\

\midrule
Synthetic-Test & BERT & Encoder (P) & -- & 0.9907 & 0.9892 & 0.9907 & 0.9817 & 0.9887 & 0.9973 & 0.9835 & 0.9942 & 0.9982 & 0.9804 & 0.9973 & 0.0003 & 0.9546 \\
Synthetic-Test & PubMedBERT & Encoder (P) & -- & 0.9933 & 0.9924 & 0.9933 & 0.9876 & 0.9916 & 0.9979 & 0.9804 & 0.9964 & 0.9991 & 0.9951 & 0.9955 & 0.0003 & 0.9666 \\
Synthetic-Test & BioBERT & Encoder (P) & -- & 0.9949 & 0.9942 & 0.9949 & 0.9909 & 0.9936 & 0.9982 & 0.9871 & 0.9978 & 0.9996 & 0.9942 & 0.9960 & 0.0001 & 0.9746 \\
Synthetic-Test & SciBERT & Encoder (P) & -- & 0.9932 & 0.9923 & 0.9932 & 0.9876 & 0.9916 & 0.9977 & 0.9840 & 0.9955 & 0.9982 & 0.9929 & 0.9955 & 0.0007 & 0.9661 \\
Synthetic-Test & BERT & Encoder (P+T) & -- & 0.9910 & 0.9897 & 0.9910 & 0.9833 & 0.9890 & 0.9969 & 0.9822 & 0.9964 & 0.9982 & 0.9830 & 0.9951 & 0.0003 & 0.9559 \\
Synthetic-Test & PubMedBERT & Encoder (P+T) & -- & 0.9935 & 0.9925 & 0.9935 & 0.9876 & 0.9921 & 0.9979 & 0.9808 & 0.9969 & 0.9987 & 0.9955 & 0.9955 & 0.0007 & 0.9693 \\
Synthetic-Test & BioBERT & Encoder (P+T) & -- & 0.9941 & 0.9934 & 0.9941 & 0.9900 & 0.9929 & 0.9974 & 0.9862 & 0.9978 & 0.9991 & 0.9938 & 0.9938 & 0.0001 & 0.9710 \\
Synthetic-Test & SciBERT & Encoder (P+T) & -- & 0.9936 & 0.9927 & 0.9936 & 0.9885 & 0.9922 & 0.9976 & 0.9853 & 0.9978 & 0.9987 & 0.9938 & 0.9924 & 0.0009 & 0.9693 \\

\bottomrule
\end{tabular}%
}
\end{table*}

\section{Fine-Tuning, Cross-Domain Transfer, and In-Domain Transfer Effects}
\label{app:finetuning_transfer_effects}

This subsection provides a detailed appendix-level analysis of how fine-tuning affects model performance under cross-domain and in-domain transfer. The PubMed comparison is reported in Table~\ref{tab:pubmed_transfer_base_ft}, while the Reddit comparison is reported in Table~\ref{tab:reddit_transfer_base_ft}. For PubMed, Synthetic-to-PubMed transfer is denoted as S$\rightarrow$P, while PubMed-to-PubMed transfer is denoted as P$\rightarrow$P. For Reddit, Synthetic-to-Reddit transfer is denoted as S$\rightarrow$R, while Reddit-to-Reddit transfer is denoted as R$\rightarrow$R. A positive delta indicates that fine-tuning or transfer improves the corresponding score, while a negative delta indicates a decrease in performance. The discussion focuses mainly on $R_{\mathrm{tuple}}$ and $F_{1,\mathrm{tuple}}$, because $R_{\mathrm{tuple}}$ measures the ability of the model to recover complete scoped tuples, while $F_{1,\mathrm{tuple}}$ captures the balance between tuple precision and tuple recall.

\begin{table*}[t]
\centering
\scriptsize
\resizebox{\textwidth}{!}{%
\begin{tabular}{lllcccccccccc}
\toprule
\textbf{Type} & \textbf{Model} & \textbf{Prompt/Type} &
\textbf{Base $R_{\mathrm{tuple}}$} &
\textbf{Base $F_{1,\mathrm{tuple}}$} &
\textbf{S$\rightarrow$P $R_{\mathrm{tuple}}$} &
\textbf{S$\rightarrow$P $F_{1,\mathrm{tuple}}$} &
\textbf{$\Delta R$ S$\rightarrow$P--Base} &
\textbf{P$\rightarrow$P $R_{\mathrm{tuple}}$} &
\textbf{P$\rightarrow$P $F_{1,\mathrm{tuple}}$} &
\textbf{$\Delta R$ P$\rightarrow$P--Base} &
\textbf{$\Delta R$ P$\rightarrow$P--S$\rightarrow$P} &
\textbf{$\Delta F_1$ P$\rightarrow$P--S$\rightarrow$P} \\
\midrule
FT-LLM & Llama-3.3-70B-Instruct & One Shot &
0.2263 & 0.1429 & 0.2220 & 0.1453 & -0.0043 & 0.1794 & 0.1565 & -0.0469 & -0.0426 & +0.0112 \\

FT-LLM & Phi-4-mini-instruct & Two Shot &
0.0643 & 0.0643 & 0.0568 & 0.0628 & -0.0075 & 0.1279 & 0.0633 & +0.0636 & +0.0711 & +0.0005 \\

FT-LLM & Qwen2.5-32B-Instruct & Two Shot + CoT &
0.0456 & 0.0531 & 0.1776 & 0.1195 & +0.1320 & 0.1847 & 0.1374 & +0.1391 & +0.0071 & +0.0179 \\

FT-LM & FLAN-T5-Base & Seq2Seq LM &
-- & -- & 0.0558 & 0.0475 & -- & 0.3440 & 0.2276 & -- & +0.2882 & +0.1801 \\

FT-LM & SpanBERT & Encoder LM &
-- & -- & 0.0000 & 0.0000 & -- & 0.2819 & 0.0147 & -- & +0.2819 & +0.0147 \\
\bottomrule
\end{tabular}
}
\caption{Base, Synthetic-to-PubMed, and PubMed-to-PubMed comparison on the PubMed test set. S$\rightarrow$P denotes synthetic fine-tuning evaluated on PubMed, while P$\rightarrow$P denotes PubMed fine-tuning evaluated on PubMed.}
\label{tab:pubmed_transfer_base_ft}
\end{table*}

\begin{table*}[t]
\centering
\scriptsize
\resizebox{\textwidth}{!}{%
\begin{tabular}{lllcccccccccc}
\toprule
\textbf{Type} & \textbf{Model} & \textbf{Prompt/Type} &
\textbf{Base $R_{\mathrm{tuple}}$} &
\textbf{Base $F_{1,\mathrm{tuple}}$} &
\textbf{S$\rightarrow$R $R_{\mathrm{tuple}}$} &
\textbf{S$\rightarrow$R $F_{1,\mathrm{tuple}}$} &
\textbf{$\Delta R$ S$\rightarrow$R--Base} &
\textbf{R$\rightarrow$R $R_{\mathrm{tuple}}$} &
\textbf{R$\rightarrow$R $F_{1,\mathrm{tuple}}$} &
\textbf{$\Delta R$ R$\rightarrow$R--Base} &
\textbf{$\Delta R$ R$\rightarrow$R--S$\rightarrow$R} &
\textbf{$\Delta F_1$ R$\rightarrow$R--S$\rightarrow$R} \\
\midrule
FT-LLM & Llama-3.3-70B-Instruct & One Shot &
0.0354 & 0.0234 & 0.0463 & 0.0416 & +0.0109 & 0.0030 & 0.0057 & -0.0324 & -0.0433 & -0.0359 \\

FT-LLM & Phi-4-mini-instruct & Two Shot &
0.0117 & 0.0123 & 0.0105 & 0.0145 & -0.0012 & 0.0284 & 0.0131 & +0.0167 & +0.0179 & -0.0014 \\

FT-LLM & Qwen2.5-32B-Instruct & Two Shot + CoT &
0.0058 & 0.0072 & 0.0269 & 0.0356 & +0.0211 & 0.0105 & 0.0167 & +0.0047 & -0.0164 & -0.0189 \\

FT-LM & FLAN-T5-Base & Seq2Seq LM &
-- & -- & 0.0000 & 0.0000 & -- & 0.1267 & 0.0816 & -- & +0.1267 & +0.0816 \\

FT-LM & SpanBERT & Encoder LM &
-- & -- & 0.0000 & 0.0000 & -- & 0.0722 & 0.0037 & -- & +0.0722 & +0.0037 \\
\bottomrule
\end{tabular}
}
\caption{Base, Synthetic-to-Reddit, and Reddit-to-Reddit comparison on the Reddit test set. S$\rightarrow$R denotes synthetic fine-tuning evaluated on Reddit, while R$\rightarrow$R denotes Reddit fine-tuning evaluated on Reddit.}
\label{tab:reddit_transfer_base_ft}
\end{table*}

\subsubsection{PubMed: Synthetic-to-PubMed versus PubMed-to-PubMed Transfer}
\label{app:pubmed_transfer_analysis}

As shown in Table~\ref{tab:pubmed_transfer_base_ft}, the PubMed results show that in-domain transfer is generally beneficial, but the effect is not uniform across model families or individual models. For the FT-LLMs, the average base $R_{\mathrm{tuple}}$ is 0.1121. Under S$\rightarrow$P transfer, the average $R_{\mathrm{tuple}}$ increases to 0.1521, giving an average gain of +0.0401 over the base models. Under P$\rightarrow$P transfer, the average $R_{\mathrm{tuple}}$ increases further to 0.1640, giving an average gain of +0.0519 over the base models. Therefore, for FT-LLMs on PubMed, both synthetic fine-tuning and PubMed fine-tuning improve tuple recall on average, but the additional advantage of P$\rightarrow$P over S$\rightarrow$P is modest, with a mean gain of only +0.0119 in $R_{\mathrm{tuple}}$.

The same pattern is also visible in $F_{1,\mathrm{tuple}}$. The average base $F_{1,\mathrm{tuple}}$ for the FT-LLMs is 0.0868. Under S$\rightarrow$P transfer, this rises to 0.1092, producing an average gain of +0.0224. Under P$\rightarrow$P transfer, it rises again to 0.1191, producing an average gain of +0.0323 over the base setting. The difference between P$\rightarrow$P and S$\rightarrow$P in $F_{1,\mathrm{tuple}}$ is +0.0099. This means that PubMed in-domain fine-tuning gives a slightly better balance of precision and recall than synthetic cross-transfer, although the improvement is not large for the decoder-based FT-LLMs.

At the individual model level, Llama-3.3-70B-Instruct behaves differently from the other FT-LLMs. Its base $R_{\mathrm{tuple}}$ is 0.2263. Under S$\rightarrow$P transfer, its $R_{\mathrm{tuple}}$ becomes 0.2220, which is a small decrease of -0.0043. This means synthetic fine-tuning does not improve Llama's PubMed tuple recall; it slightly reduces it. However, its $F_{1,\mathrm{tuple}}$ increases from 0.1429 to 0.1453, a gain of +0.0024. This suggests that although Llama retrieves slightly fewer complete tuples after synthetic fine-tuning, the predictions become marginally better balanced in terms of precision and recall. Under P$\rightarrow$P transfer, Llama's $R_{\mathrm{tuple}}$ drops further to 0.1794. This is a decrease of -0.0469 relative to the base model and -0.0426 relative to S$\rightarrow$P. However, its $F_{1,\mathrm{tuple}}$ increases to 0.1565, which is +0.0136 above the base setting and +0.0112 above S$\rightarrow$P. This is an important result: PubMed fine-tuning reduces Llama's complete tuple recall, but improves tuple-level F1. In other words, Llama becomes less recall-oriented but somewhat more balanced after P$\rightarrow$P fine-tuning.

Phi-4-mini-instruct shows a different pattern. Its base $R_{\mathrm{tuple}}$ is 0.0643. Under S$\rightarrow$P transfer, it decreases to 0.0568, a delta of -0.0075. Its $F_{1,\mathrm{tuple}}$ also decreases slightly from 0.0643 to 0.0628, a delta of -0.0015. Therefore, synthetic fine-tuning does not help Phi on PubMed; it slightly hurts both tuple recall and tuple F1. Under P$\rightarrow$P transfer, however, Phi's $R_{\mathrm{tuple}}$ rises sharply to 0.1279. This is a gain of +0.0636 over the base model and +0.0711 over S$\rightarrow$P. The $F_{1,\mathrm{tuple}}$ under P$\rightarrow$P is 0.0633, which is slightly below the base value by -0.0010 but slightly above S$\rightarrow$P by +0.0005. This means that PubMed fine-tuning substantially improves Phi's ability to recover complete tuples, but this recall gain does not translate into a strong F1 gain. The likely reason is that the model retrieves more correct tuples but may also produce additional incorrect or weakly matched predictions, limiting the F1 improvement.

Qwen2.5-32B-Instruct benefits the most consistently from fine-tuning on PubMed. Its base $R_{\mathrm{tuple}}$ is 0.0456. Under S$\rightarrow$P transfer, it increases to 0.1776, producing a large gain of +0.1320. Its $F_{1,\mathrm{tuple}}$ also increases from 0.0531 to 0.1195, a gain of +0.0664. This shows that synthetic fine-tuning gives Qwen a much stronger extraction signal for PubMed than its base instruction-following ability alone. Under P$\rightarrow$P transfer, Qwen improves slightly further: $R_{\mathrm{tuple}}$ increases to 0.1847, which is +0.1391 above the base model and +0.0071 above S$\rightarrow$P. Its $F_{1,\mathrm{tuple}}$ increases to 0.1374, which is +0.0843 above the base model and +0.0179 above S$\rightarrow$P. Unlike Llama and Phi, Qwen improves in both recall and F1 under both transfer settings. This indicates that Qwen is highly responsive to fine-tuning and that PubMed-specific fine-tuning gives it an additional advantage beyond synthetic supervision.

The FT-LMs show a much stronger dependence on in-domain PubMed supervision than the FT-LLMs. FLAN-T5-Base obtains an $R_{\mathrm{tuple}}$ of 0.0558 under S$\rightarrow$P and 0.3440 under P$\rightarrow$P, giving a very large gain of +0.2882. Its $F_{1,\mathrm{tuple}}$ also increases from 0.0475 to 0.2276, a gain of +0.1801. SpanBERT shows a similar pattern: it obtains 0.0000 in both $R_{\mathrm{tuple}}$ and $F_{1,\mathrm{tuple}}$ under S$\rightarrow$P, but improves to 0.2819 in $R_{\mathrm{tuple}}$ and 0.0147 in $F_{1,\mathrm{tuple}}$ under P$\rightarrow$P. The corresponding deltas are +0.2819 for $R_{\mathrm{tuple}}$ and +0.0147 for $F_{1,\mathrm{tuple}}$. These results show that the smaller or more traditional supervised models do not transfer well from synthetic data to PubMed, but they benefit strongly from PubMed-specific training.

Overall, PubMed shows three important trends. First, in-domain transfer is generally better than synthetic cross-transfer, especially for FLAN-T5-Base, SpanBERT, and Phi. Second, Qwen benefits from both synthetic and PubMed fine-tuning, suggesting that it can absorb the extraction structure from synthetic data and refine it further using in-domain biomedical examples. Third, Llama behaves differently: fine-tuning reduces $R_{\mathrm{tuple}}$ but improves $F_{1,\mathrm{tuple}}$, indicating a possible shift from broad recall toward more selective prediction. Thus, PubMed in-domain fine-tuning is useful, but its benefit depends on the model's starting ability, model architecture, and sensitivity to the fine-tuning distribution.

\subsubsection{Reddit: Synthetic-to-Reddit versus Reddit-to-Reddit Transfer}
\label{app:reddit_transfer_analysis}

As shown in Table~\ref{tab:reddit_transfer_base_ft}, the Reddit results are less stable than the PubMed results. Unlike PubMed, where P$\rightarrow$P generally improves performance, Reddit in-domain transfer does not consistently help the FT-LLMs. For the FT-LLMs, the average base $R_{\mathrm{tuple}}$ is 0.0176. Under S$\rightarrow$R transfer, the average $R_{\mathrm{tuple}}$ increases to 0.0279, giving a gain of +0.0103 over the base models. Under R$\rightarrow$R transfer, however, the average $R_{\mathrm{tuple}}$ falls to 0.0140, which is -0.0037 below the base setting and -0.0139 below S$\rightarrow$R. This shows that, for the FT-LLMs, Reddit in-domain fine-tuning is not only weaker than synthetic cross-transfer on average, but can also reduce performance below the base model level.

The same pattern appears more clearly in $F_{1,\mathrm{tuple}}$. The FT-LLM base average $F_{1,\mathrm{tuple}}$ is 0.0143. Under S$\rightarrow$R transfer, this increases to 0.0306, giving an average gain of +0.0163. Under R$\rightarrow$R transfer, however, the average $F_{1,\mathrm{tuple}}$ decreases to 0.0118, which is -0.0025 below the base setting and -0.0187 below S$\rightarrow$R. Therefore, synthetic fine-tuning provides a stronger and more consistent signal for Reddit FT-LLMs than Reddit fine-tuning itself. This suggests that the synthetic data may provide clearer extraction patterns, while Reddit in-domain data may introduce noise, informal phrasing, implicit causality, or inconsistent tuple boundaries that make fine-tuning less stable.

Llama-3.3-70B-Instruct illustrates this instability most strongly. Its base $R_{\mathrm{tuple}}$ on Reddit is 0.0354, with a base $F_{1,\mathrm{tuple}}$ of 0.0234. Under S$\rightarrow$R transfer, $R_{\mathrm{tuple}}$ improves to 0.0463, a gain of +0.0109. Its $F_{1,\mathrm{tuple}}$ also increases to 0.0416, a gain of +0.0182. This means synthetic fine-tuning helps Llama on Reddit, even though the target domain is not synthetic. However, under R$\rightarrow$R transfer, Llama's $R_{\mathrm{tuple}}$ collapses to 0.0030. This is -0.0324 below the base model and -0.0433 below S$\rightarrow$R. Its $F_{1,\mathrm{tuple}}$ also drops to 0.0057, which is -0.0177 below the base setting and -0.0359 below S$\rightarrow$R. This is one of the strongest negative transfer effects in the results. It indicates that Reddit in-domain fine-tuning severely harms Llama's ability to recover complete scoped tuples, even though synthetic fine-tuning improves it.

Phi-4-mini-instruct shows a more nuanced pattern. Its base $R_{\mathrm{tuple}}$ is 0.0117 and its base $F_{1,\mathrm{tuple}}$ is 0.0123. Under S$\rightarrow$R transfer, its $R_{\mathrm{tuple}}$ slightly decreases to 0.0105, a delta of -0.0012. However, its $F_{1,\mathrm{tuple}}$ increases to 0.0145, a delta of +0.0022. This means synthetic fine-tuning slightly reduces complete tuple recall but slightly improves the precision-recall balance. Under R$\rightarrow$R transfer, Phi's $R_{\mathrm{tuple}}$ increases to 0.0284, which is +0.0167 above the base model and +0.0179 above S$\rightarrow$R. However, its $F_{1,\mathrm{tuple}}$ is 0.0131, which is only +0.0008 above the base model and -0.0014 below S$\rightarrow$R. This means Reddit fine-tuning helps Phi retrieve more complete tuples, but the improvement is mostly recall-oriented and does not produce a meaningful F1 gain. Phi is therefore the only FT-LLM where R$\rightarrow$R clearly improves $R_{\mathrm{tuple}}$ over S$\rightarrow$R, but the F1 result remains weak.

Qwen2.5-32B-Instruct performs better with S$\rightarrow$R than with R$\rightarrow$R. Its base $R_{\mathrm{tuple}}$ is 0.0058 and its base $F_{1,\mathrm{tuple}}$ is 0.0072. Under S$\rightarrow$R transfer, $R_{\mathrm{tuple}}$ rises to 0.0269, giving a gain of +0.0211. Its $F_{1,\mathrm{tuple}}$ also rises to 0.0356, a gain of +0.0284. This shows that Qwen benefits substantially from synthetic fine-tuning when evaluated on Reddit. Under R$\rightarrow$R transfer, however, Qwen's $R_{\mathrm{tuple}}$ is only 0.0105. This is still +0.0047 above the base model, but it is -0.0164 below S$\rightarrow$R. Its $F_{1,\mathrm{tuple}}$ is 0.0167, which is +0.0095 above the base model but -0.0189 below S$\rightarrow$R. Thus, Reddit fine-tuning helps Qwen relative to its base version, but it is clearly worse than synthetic fine-tuning. This suggests that Qwen can use fine-tuning effectively, but the cleaner synthetic signal transfers better to Reddit than noisy Reddit in-domain supervision.

The FT-LMs show the opposite pattern from the FT-LLMs. FLAN-T5-Base obtains 0.0000 for both $R_{\mathrm{tuple}}$ and $F_{1,\mathrm{tuple}}$ under S$\rightarrow$R, but improves to 0.1267 in $R_{\mathrm{tuple}}$ and 0.0816 in $F_{1,\mathrm{tuple}}$ under R$\rightarrow$R. The corresponding deltas are +0.1267 and +0.0816. SpanBERT also obtains 0.0000 under S$\rightarrow$R, but improves to 0.0722 in $R_{\mathrm{tuple}}$ and 0.0037 in $F_{1,\mathrm{tuple}}$ under R$\rightarrow$R. The deltas are +0.0722 and +0.0037. These results indicate that the supervised FT-LMs require in-domain Reddit training to perform at all. Unlike the FT-LLMs, they do not appear to benefit from synthetic-to-Reddit transfer.

The Reddit results therefore show a split between model families. For FT-LLMs, S$\rightarrow$R is generally stronger than R$\rightarrow$R. Llama improves under S$\rightarrow$R but collapses under R$\rightarrow$R. Qwen improves under both settings relative to the base model, but the S$\rightarrow$R gains are much larger. Phi is the exception, since R$\rightarrow$R improves its $R_{\mathrm{tuple}}$ more than S$\rightarrow$R, although its $F_{1,\mathrm{tuple}}$ remains nearly unchanged. For FT-LMs, however, R$\rightarrow$R is clearly better than S$\rightarrow$R, because both FLAN-T5-Base and SpanBERT fail under synthetic cross-transfer but improve under Reddit-specific training.

Overall, Reddit demonstrates that domain match alone is not sufficient to guarantee better fine-tuning outcomes. In-domain Reddit fine-tuning helps the supervised FT-LMs, but it does not consistently help the instruction-tuned FT-LLMs. The stronger performance of S$\rightarrow$R for Llama and Qwen suggests that clean synthetic supervision may teach the tuple extraction format more effectively than noisy in-domain Reddit examples. Reddit likely contains more informal language, implicit claims, fragmented statements, sarcasm, and less stable causal structure than PubMed. As a result, Reddit in-domain fine-tuning may expose models to target-domain vocabulary, but it may also weaken the consistency of the extraction behavior. This explains why R$\rightarrow$R can improve some recall values while still reducing $F_{1,\mathrm{tuple}}$ or producing unstable model-level effects.

\section{Embedding-Based Data Validation}
\label{app:data_validation}

This appendix reports the embedding-based validation used to examine whether the three source groups in CondRelBench occupy distinguishable regions in semantic space. The purpose of this analysis is to test whether PubMed, Reddit, and Synthetic examples differ enough in their distributional structure to justify treating them as distinct evaluation settings. This matters because CondRelBench combines a controlled synthetic source, a formal biomedical source, and an informal user-generated source. If the three sources were distributionally similar, then cross-domain evaluation would be less informative. If they are distributionally distinct, then transfer experiments are useful for testing whether models learn general condition-aware relation behavior or mainly fit the surface distribution of the source on which they are trained.

\subsection{Embedding Model and PCA Projection}
\label{app:data_validation_pca}

Each paragraph was embedded using \texttt{sentence-transformers/all-MiniLM-L6-v2}. Let $x_i$ denote a paragraph and let $\mathbf{e}_i=f(x_i)$ denote its sentence embedding. Each paragraph is associated with a source label $s_i \in \{\text{PubMed},\text{Reddit},\text{Synthetic}\}$.

\begin{figure}[t]
    \centering
    \includegraphics[width=0.78\linewidth]{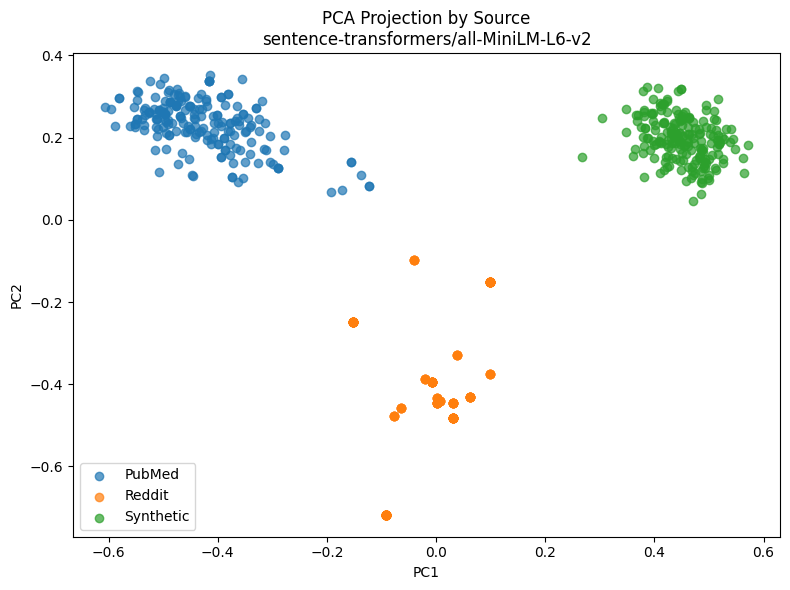}
    \caption{PCA projection of paragraph embeddings by source using \texttt{sentence-transformers/all-MiniLM-L6-v2}. PubMed, Reddit, and Synthetic examples occupy visibly different regions of the embedding space. Reddit is more dispersed, Synthetic is more compact, and PubMed forms a separate biomedical cluster.}
    \label{fig:pca_source_projection}
\end{figure}

To visually inspect the source distributions, we projected the paragraph embeddings into two dimensions using principal component analysis (PCA). The PCA projection is shown in Figure~\ref{fig:pca_source_projection}. The figure shows that the three source groups occupy visibly different regions of the embedding space. PubMed examples form a cluster toward the left of the projection, Synthetic examples form a compact cluster toward the right, and Reddit examples appear in a lower region with fewer points and greater dispersion. This visualization is consistent with the expected source differences: Synthetic is controlled, PubMed is formal biomedical text, and Reddit is informal user-generated text.

The PCA plot is used as an exploratory visualization rather than a statistical test. Since PCA compresses the original embedding space into two dimensions, it cannot by itself establish distributional separation. We therefore complement the visualization with three quantitative checks: within-source centroid distance, non-parametric tests over centroid-distance distributions, and maximum mean discrepancy tests over full embedding distributions.

\subsection{Within-Source Centroid Distance}
\label{app:data_validation_centroid}

To measure how compact or internally varied each source group is, we computed the centroid of each source in embedding space. For each source $s$, the centroid is defined as:

\[
\boldsymbol{\mu}_s=
\frac{1}{|D_s|}
\sum_{x_i \in D_s}
\mathbf{e}_i,
\]

where $D_s$ is the set of paragraphs belonging to source $s$. We then measured the cosine distance between each paragraph embedding and the centroid of its own source:

\[
d_i = 1 - \cos(\mathbf{e}_i, \boldsymbol{\mu}_{s_i}).
\]

This produces a source-specific compactness measure. A lower mean distance indicates that examples from that source are more tightly clustered around their own centroid. A higher mean distance indicates greater internal variation.

\begin{table}[t]
\centering
\small
\begin{tabular}{lr}
\hline
\textbf{Source} & \textbf{Mean distance to own centroid} \\
\hline
Synthetic & 0.287 \\
PubMed & 0.363 \\
Reddit & 0.476 \\
\hline
\end{tabular}
\caption{Mean cosine distance from each paragraph embedding to its own source centroid. Lower values indicate a more compact source distribution, while higher values indicate greater internal variation.}
\label{tab:centroid_distance}
\end{table}

The results in Table~\ref{tab:centroid_distance} show that Synthetic has the smallest mean distance to its own centroid, with a mean distance of $0.287$. This is expected because the synthetic dataset was generated under controlled conditions, with systematic variation over entities, relations, conditions, and perturbation patterns. The synthetic examples are therefore more regular than naturally occurring text.

PubMed has a mean distance of $0.363$, placing it between Synthetic and Reddit. This indicates that PubMed is more varied than Synthetic, but still relatively structured. This is also expected because PubMed paragraphs come from formal biomedical writing, where style, terminology, and discourse structure are more standardized than in social media text.

Reddit has the largest mean distance, $0.476$, indicating the greatest within-source variation. This reflects the nature of Reddit as informal user-generated text. Reddit posts are less standardized, vary more strongly in tone and structure, and may express conditions less explicitly or less consistently than biomedical or synthetic text. This greater internal variation also helps explain why Reddit is the most difficult source in the downstream experiments.

\subsection{Kruskal--Wallis Test Across Source Distance Distributions}
\label{app:data_validation_kruskal}

The centroid-distance results suggest that the sources differ in internal compactness. To test whether these differences are statistically significant, we applied a Kruskal--Wallis test to the source-wise distance distributions. We used the Kruskal--Wallis test because it does not assume that the distance values are normally distributed.

The null hypothesis is that the three source groups come from the same distance distribution:

{\scriptsize
\[
\begin{aligned}
H_0:\quad
d_{\text{PubMed}},\;
d_{\text{Reddit}},\;
d_{\text{Synthetic}}
&\sim \text{same distribution}.
\end{aligned}
\]
}
The alternative hypothesis is that at least one source group has a different distance distribution:

\[
H_1:
\exists s_a, s_b
\text{ such that }
d_{s_a}
\text{ and }
d_{s_b}
\text{ differ in distribution.}
\]

The Kruskal--Wallis test returned:

\[
H = 285.51,
\qquad
p = 1.01 \times 10^{-62}.
\]

This result strongly rejects the null hypothesis. The extremely small $p$-value shows that the three source groups do not have the same within-source distance distribution. In practical terms, the sources differ in how tightly or loosely their paragraph embeddings cluster around their own source centroids.

\subsection{Pairwise Mann--Whitney Tests with Holm Correction}
\label{app:data_validation_mann_whitney}

After the Kruskal--Wallis test showed that the three source distance distributions differ overall, we conducted pairwise Mann--Whitney tests to identify whether each pair of sources differs significantly. The tested source pairs were:

\begin{itemize}
    \item PubMed--Reddit
    \item PubMed--Synthetic
    \item Reddit--Synthetic
\end{itemize}

The pairwise tests were corrected using Holm correction to control for multiple comparisons. All pairwise Mann--Whitney comparisons were significant after Holm correction. This means that the observed differences are not driven by only one source being different from the others. Instead, each source differs from each other source in its within-source distance profile.

This result supports three separate conclusions. First, Synthetic is significantly more compact than PubMed and Reddit. Second, Reddit is significantly more internally varied than both PubMed and Synthetic. Third, PubMed occupies an intermediate but still distinct position. Therefore, the three datasets should not be treated as interchangeable samples from a single distribution.

\subsection{Maximum Mean Discrepancy Test}
\label{app:data_validation_mmd}

The centroid-distance analysis measures how far examples are from their own source centroid. However, source centroids alone do not capture all distributional differences. Two sources can have similar centroid distances but still differ in the shape, density, or higher-order structure of their embedding distributions. To compare the full embedding distributions, we also applied a maximum mean discrepancy (MMD) permutation test.

For two embedding samples $X=\{x_1,\ldots,x_m\}$ and $Y=\{y_1,\ldots,y_n\}$, the squared MMD statistic is:
{\small
\[
\begin{aligned}
\mathrm{MMD}^2(X,Y)
&=
\frac{1}{m^2}
\sum_{i,j}
k(x_i,x_j)
+
\frac{1}{n^2}
\sum_{i,j}
k(y_i,y_j) \\
&\quad
-
\frac{2}{mn}
\sum_{i,j}
k(x_i,y_j),
\end{aligned}
\]
}
where $k(\cdot,\cdot)$ is a kernel function. MMD measures whether two samples appear to come from the same distribution in the embedding space. A larger discrepancy indicates greater distributional difference between the two source groups.

We applied MMD permutation tests to all three source pairs:

\begin{itemize}
    \item PubMed--Reddit
    \item PubMed--Synthetic
    \item Reddit--Synthetic
\end{itemize}

The MMD tests showed significant distributional differences for all three pairwise comparisons. After Holm correction, the adjusted $p$-value was:

\[
p_{\text{Holm}} = 0.002997
\]

for each source-pair comparison. This confirms that the full embedding distributions differ significantly, not just their centroid-distance summaries.

\begin{table*}[t]
\centering
\small
\begin{tabular}{ll}
\hline
\textbf{Comparison} & \textbf{MMD result} \\
\hline
PubMed--Reddit & Significant after Holm correction, $p_{\text{Holm}}=0.002997$ \\
PubMed--Synthetic & Significant after Holm correction, $p_{\text{Holm}}=0.002997$ \\
Reddit--Synthetic & Significant after Holm correction, $p_{\text{Holm}}=0.002997$ \\
\hline
\end{tabular}
\caption{Pairwise MMD permutation tests over embedding distributions. All source pairs show significant distributional differences after Holm correction.}
\label{tab:mmd_results}
\end{table*}

\subsection{Interpretation of the Distributional Validation}
\label{app:data_validation_interpretation}

Across the visualization and statistical tests, the results consistently show that the three source groups are distributionally distinct. The PCA projection in Figure~\ref{fig:pca_source_projection} gives an intuitive view of this separation. PubMed, Reddit, and Synthetic occupy visibly different regions in the projected embedding space. Synthetic forms a compact cluster, PubMed forms a separate formal biomedical cluster, and Reddit appears more dispersed. This matches the expected properties of the sources: Synthetic is controlled, PubMed is formal and technical, and Reddit is informal and noisier.

The centroid-distance analysis quantifies this pattern. Synthetic has the lowest mean distance to its source centroid, which supports its role as a controlled diagnostic subset. Reddit has the highest mean distance, showing that it contains greater internal variation. PubMed falls between them, reflecting a more standardized but still naturally occurring biomedical distribution.

The Kruskal--Wallis result confirms that these distance distributions differ significantly across the three sources. The pairwise Mann--Whitney tests further show that each source differs from the others, rather than one source alone driving the overall result. Finally, the MMD permutation tests show that the full embedding distributions are significantly different for every source pair.

Together, these results support the claim that CondRelBench is not merely a mixture of three differently named datasets. It contains three distributionally distinct source groups: a controlled synthetic subset, a formal biomedical subset, and an informal user-generated subset.

\subsection{Implications for Cross-Domain Fine-Tuning}
\label{app:data_validation_cross_finetuning}

The distributional validation motivates our cross-domain fine-tuning experiments, especially the synthetic-to-real transfer setting. Since Synthetic, PubMed, and Reddit occupy different regions of embedding space, training on one source should not be expected to transfer perfectly to another. This is particularly important for the synthetic-to-real setup. The synthetic subset is useful because it provides controlled coverage of relation structures, condition structures, and perturbation behavior. However, because it is also the most compact source group, it may not fully represent the linguistic variation found in PubMed and Reddit.

This means that synthetic-to-real fine-tuning is a deliberately difficult test. We do not expect synthetic training to fully solve PubMed or Reddit performance. Instead, the experiment asks a narrower and more diagnostic question: does training on controlled condition-scoped examples teach any transferable behavior that helps models recognize scoped relations in naturally occurring text?

The distributional results suggest that transfer from Synthetic to PubMed may be more plausible than transfer from Synthetic to Reddit. PubMed is formal and more structured than Reddit, even though it remains distributionally different from Synthetic. Reddit, by contrast, has the largest within-source variation and is likely to contain more informal, implicit, and noisy condition expressions. Therefore, if synthetic-to-real transfer is weak on Reddit, that does not invalidate the synthetic benchmark. It instead reflects the larger domain gap identified by the embedding analysis.

The validation also explains why in-domain fine-tuning remains important. If the three sources are significantly different, then models may need source-specific exposure to adapt to the language patterns of each domain. PubMed may require adaptation to biomedical terminology and formal scientific phrasing. Reddit may require adaptation to informal discourse, implicit conditions, personal narratives, and less standardized relation expression. Synthetic data can teach controlled scope behavior, but it cannot be assumed to cover all real-domain surface forms.

Therefore, the cross-domain fine-tuning experiments should be interpreted cautiously. Strong synthetic-to-real transfer would suggest that the model learned a generalizable condition-aware behavior. Weak synthetic-to-real transfer would be expected given the measured distributional gap, especially for Reddit. In both cases, the experiment is informative because it tests whether condition-aware relation understanding transfers beyond the controlled diagnostic source.

\section{Probe Validation}
\label{app:probe_validation}

This appendix describes the validation process used after probe construction. For each annotated scoped tuple $(E_1, R, E_2 \mid C)$, five probe classes were expected: relation contradiction, condition replacement, condition removal, supported original tuple, and condition contradiction. The expected number of probes for each class is therefore equal to the number of annotated tuples in that dataset. The actual number may be lower because problematic probes were removed during validation.

\subsection{Validation Procedure}

Probe validation was conducted by a group of annotators who reviewed the generated probes collectively. The group discussed each questionable probe and removed probes that were judged to be problematic. A probe was considered problematic if it failed to preserve the intended intervention for its probe class, introduced unsupported information, changed more than one intended component, produced an ambiguous label, or made the answer depend on information outside the source text.

In this context, a problematic probe refers to any generated probe that could not be confidently assigned its intended gold label under the benchmark design. This includes cases where a relation-contradiction probe did not clearly contradict the relation, a condition-replacement probe accidentally remained supported by the source text, a condition-removal probe still contained a scope-bearing condition, a supported-original probe altered the original tuple, or a condition-contradiction probe did not clearly contradict the original condition. Such probes were removed rather than corrected in order to preserve a clean evaluation set.

\subsection{Expected and Actual Probe Counts}

Let $N_d$ denote the number of annotated scoped tuples in dataset $d$. Since each tuple is expected to generate five probes, the expected number of probes for each probe class is:

\[
E_{d,k} = N_d,
\]

where $k \in \{P1, P2, P3, P4, P5\}$. The expected total number of probes for dataset $d$ is:

\[
E_d = 5N_d.
\]

The actual number of probes after validation is denoted as $A_{d,k}$ for each probe class and $A_d$ for the dataset total. The number of removed probes is:

\[
R_{d,k} = E_{d,k} - A_{d,k},
\]

and the retention rate is:

\[
\mathrm{Retention}_{d,k} =
\frac{A_{d,k}}{E_{d,k}} \times 100.
\]

\begin{table}[t]
\centering
\small
\resizebox{\columnwidth}{!}{%
\begin{tabular}{llrrrrr}
\hline
\textbf{Dataset} & \textbf{Probe Class} & \textbf{Expected} & \textbf{Actual} & \textbf{Removed} & \textbf{Retention} & \textbf{Gold Label} \\
\hline
Synthetic & P1: Relation Contradiction & 7,765 & 7,455 & 310 & 96.01\% & Not Supported \\
Synthetic & P2: Condition Replacement & 7,765 & 7,456 & 309 & 96.02\% & Not Enough Evidence \\
Synthetic & P3: Condition Removal & 7,765 & 7,458 & 307 & 96.05\% & Not Enough Evidence \\
Synthetic & P4: Supported Original Tuple & 7,765 & 7,441 & 324 & 95.83\% & Supported \\
Synthetic & P5: Condition Contradiction & 7,765 & 7,458 & 307 & 96.05\% & Not Supported \\
\hline
PubMed & P1: Relation Contradiction & 1,397 & 1,396 & 1 & 99.93\% & Not Supported \\
PubMed & P2: Condition Replacement & 1,397 & 1,397 & 0 & 100.00\% & Not Enough Evidence \\
PubMed & P3: Condition Removal & 1,397 & 1,397 & 0 & 100.00\% & Not Enough Evidence \\
PubMed & P4: Supported Original Tuple & 1,397 & 1,374 & 23 & 98.35\% & Supported \\
PubMed & P5: Condition Contradiction & 1,397 & 1,397 & 0 & 100.00\% & Not Supported \\
\hline
Reddit & P1: Relation Contradiction & 1,586 & 1,586 & 0 & 100.00\% & Not Supported \\
Reddit & P2: Condition Replacement & 1,586 & 1,586 & 0 & 100.00\% & Not Enough Evidence \\
Reddit & P3: Condition Removal & 1,586 & 1,586 & 0 & 100.00\% & Not Enough Evidence \\
Reddit & P4: Supported Original Tuple & 1,586 & 1,539 & 47 & 97.04\% & Supported \\
Reddit & P5: Condition Contradiction & 1,586 & 1,586 & 0 & 100.00\% & Not Supported \\
\hline
\end{tabular}%
}
\caption{Expected and actual probe counts by dataset and probe class after group validation. Expected counts are based on one probe per class for each annotated scoped tuple. Actual counts reflect the retained probes after problematic probes were removed.}
\label{tab:probe_validation_by_class}
\end{table}

\subsection{Dataset-Level Probe Retention}

Table~\ref{tab:probe_validation_summary} summarizes the expected and actual number of probes for each dataset. The expected total is computed as five times the number of annotated scoped tuples. The actual totals reflect the final retained probe set after problematic probes were removed.

\begin{table}[t]
\centering
\small
\resizebox{\columnwidth}{!}{%
\begin{tabular}{lrrrrr}
\hline
\textbf{Dataset} & \textbf{Annotated Tuples} & \textbf{Expected Probes} & \textbf{Actual Probes} & \textbf{Removed Probes} & \textbf{Retention} \\
\hline
Synthetic & 7,765 & 38,825 & 37,268 & 1,557 & 95.99\% \\
PubMed & 1,397 & 6,985 & 6,961 & 24 & 99.66\% \\
Reddit & 1,586 & 7,930 & 7,883 & 47 & 99.41\% \\
\hline
\end{tabular}%
}
\caption{Dataset-level probe retention after validation. Expected probes are computed as five probes per annotated scoped tuple. Actual probes are the retained probes after problematic probes were removed.}
\label{tab:probe_validation_summary}
\end{table}

\subsection{Validation Metrics}

We report the following validation metrics:

\begin{itemize}
    \item \textbf{Expected probes}: the number of probes that should be generated for a dataset or probe class before validation. For each probe class, this equals the number of annotated scoped tuples.
    \item \textbf{Actual probes}: the number of probes retained after validation.
    \item \textbf{Removed probes}: the number of generated probes removed during validation because they were judged problematic.
    \item \textbf{Retention rate}: the percentage of expected probes retained after validation.
    \item \textbf{Removal rate}: the percentage of expected probes removed after validation.
\end{itemize}

The removal rate is defined as:

\[
\mathrm{Removal}_{d,k} =
\frac{R_{d,k}}{E_{d,k}} \times 100.
\]

For dataset-level validation, the same metric is computed using total expected and actual probes:

{
\[
\begin{aligned}
\mathrm{Retention}_{d}
&= \frac{A_d}{E_d} \times 100,\\
\mathrm{Removal}_{d}
&= \frac{E_d - A_d}{E_d} \times 100.
\end{aligned}
\]
}
\begin{table*}[t]
\centering
\tiny
\begin{tabular}{p{2.5cm}p{11.5cm}}
\toprule
\textbf{Field} & \textbf{Content} \\
\midrule
\textbf{Input paragraph} &
The write-up clearly distinguishes between different settings before specifying the outcome. A particular claim in the text is that Intervention FF and Drug Y boost Marker Delta specifically for setting upsilon. The surrounding wording avoids a generalized interpretation. The paragraph briefly mentions that Profile O has a slight link to Variable D within surface state orange; this remains insignificant from the main data. That additional comment serves only as noise for the extractor. \\
\midrule
\textbf{Probe statement} &
Clinical note: In setting upsilon, the team tested Intervention FF with Marker Delta as the readout. The record emphasizes that after administration, Marker Delta levels declined relative to pre-intervention baseline, and the change was described as a reduction rather than a boost. The clinician notes no sustained elevation and characterizes the effect as inhibitory, citing repeated measures at follow-up timepoints. \\
\midrule
\textbf{Gold label} &
\textit{Not Supported} \\
\bottomrule
\end{tabular}
\caption{Example of the model-facing input and gold label for a relation-contradiction probe.}
\label{tab:contextual_probe_example}
\end{table*}

\subsection{Interpretation}

The probe validation process ensures that the final benchmark evaluates condition-aware relation understanding rather than artifacts of probe generation. Because each probe class tests a specific perturbation, retaining a problematic probe would weaken the interpretation of model performance. For example, if a condition-replacement probe is still supported by the original paragraph, then a model should not be penalized for predicting \textit{Supported}. Similarly, if a condition-removal probe still contains the original condition implicitly, then it no longer tests whether the model overgeneralizes beyond the stated condition.

Removing problematic probes therefore improves the validity of the probe evaluation. The retained probes are those for which the group could confidently preserve the intended probe operation and gold label. This makes the final probe set cleaner and better aligned with the benchmark objective: testing whether models preserve the condition under which a relation holds.

\subsection{Contextual Probe Input, Output, and Gold Label}
\label{app:contextual_probe_io_gold}

Each probe instance is evaluated as a contextual inference problem rather than as a standalone sentence classification task. The model receives two inputs: the source paragraph and a probe statement. The expected output is one label from the three-way label space: \textit{Supported}, \textit{Not Supported}, or \textit{Not Enough Evidence}. The gold label is assigned by comparing the probe statement against the scoped relation licensed by the paragraph.

In this example, the paragraph licenses the scoped tuple
{\scriptsize
\[
(\text{Intervention FF},\ \text{boost},\ \text{Marker Delta}\mid \text{setting upsilon}).
\]}
The probe preserves the same general setting, \textit{setting upsilon}, but reverses the relation by stating that Marker Delta declined and that the effect was inhibitory. The probe is therefore not a case of missing evidence. It is a direct contradiction of the relation expressed in the paragraph. The correct gold label is consequently \textit{Not Supported}.

This design also explains why the probe is not written as a simple isolated sentence. A simple sentence such as ``Intervention FF reduces Marker Delta'' would test only surface-level contradiction. It would not test whether a model can track the original condition, ignore distractor content, preserve the relevant entities, and compare the probe relation against the scoped claim in the paragraph. The longer probe format forces the model to evaluate the claim in context: it must identify the relation-bearing tuple in the paragraph, determine the condition under which it holds, and then decide whether the probe preserves, removes, replaces, or contradicts that scoped relation.

\section{Responsible Research, Artifacts, and AI-Assistance Disclosure}
\label{app:responsible_research_artifacts}

This section documents the research artifacts, intended use, privacy safeguards, annotation process, computational experiments, and AI-assistant use associated with this work. It is included to make the responsible research assumptions behind CondRelBench explicit.

\subsection{Artifacts Used and Created}
\label{app:artifacts_used_created}

This work creates CondRelBench, a mixed-domain benchmark for condition-scoped relation understanding. The benchmark contains Synthetic, PubMed, and Reddit subsets, together with condition-scoped tuples, probe claims, gold labels, metadata, and evaluation scripts. The benchmark is intended for research on condition-aware relation extraction, scoped claim verification, model evaluation, and overgeneralization analysis. It is not intended for clinical, legal, financial, or other high-stakes decision-making.

The work also uses existing artifacts, including pretrained language models, tokenizer implementations, biomedical articles, public Reddit posts, and standard Python-based machine learning and evaluation packages. Where existing artifacts are used, we cite the original creators or providers in the main paper, appendix, references, metadata, or release documentation. For PubMed-derived examples, source article metadata is retained to support attribution. For Reddit-derived examples, source-level documentation is maintained in the dataset metadata while applying the privacy filtering described below.

\subsection{Licensing, Terms of Use, and Intended Use}
\label{app:artifact_license_terms}

The released benchmark artifacts are intended for all research and evaluation with attribution. Dataset metadata records the source domain, dataset split, tuple information, probe type, gold label, and source attribution fields where applicable. Any public release of the benchmark should preserve source attribution and should not override the access terms or licenses of the source materials.

The synthetic portion is created by the authors and may be distributed with the benchmark release under the license specified in the accompanying release documentation. PubMed-derived examples are released with citation metadata so that users can identify and credit the original scientific sources. Reddit-derived examples are included only after privacy filtering and are intended for research on language understanding, not for user profiling, identity inference, or redistribution outside the stated research context.

\subsection{Privacy and Data Filtering}
\label{app:privacy_filtering}

The PubMed subset is based on scientific text from published biomedical papers. The Reddit subset is based on publicly available posts, but additional privacy filtering was applied because Reddit posts may contain personal or identifying details. We removed any Reddit document that mentioned a name of a person, family member, pet, or similar identifying detail. Retained Reddit rows did not require further anonymization after this filtering step. The goal of this filtering was to preserve the linguistic properties of informal discourse while reducing the risk of exposing private identity or personal relationships.

The benchmark should therefore be used only for research on condition-aware relation understanding. It should not be used to identify, profile, contact, or make decisions about individuals.

\subsection{Annotation and Human Involvement}
\label{app:human_annotation_disclosure}

Human annotation was used to identify and validate condition-scoped tuples of the form $(E_1, R, E_2 \mid C)$. Annotators were given instructions describing the tuple fields, condition definition, span-matching rules, acceptance criteria, and rejection criteria. The annotation process followed a two-stage design. Stage 1 identified candidate tuples by marking the relation arguments, relation span, and condition. Stage 2 validated whether each candidate tuple was acceptable.

For PubMed and Reddit, validation subsets were independently checked by annotators using token-level overlap criteria, and candidate tuples were accepted only when the relation, arguments, and condition were sufficiently supported by the source text. Disagreements or invalid rows were removed rather than forced into the dataset. This design prioritizes high-confidence scoped relations over exhaustive annotation of every possible relation in each document.

Annotators were selected based on relevant training or familiarity with biomedical and linguistic annotation tasks. Annotation instructions and validation criteria are documented in the appendix so that the annotation decisions can be inspected and reproduced.

\subsection{Computational Experiments and Reproducibility}
\label{app:compute_reproducibility}

The experiments include prompted model evaluation, fine-tuning, cross-domain transfer evaluation, tuple extraction evaluation, probe classification, and ablation experiments. The models are grouped into prompted large language models, fine-tuned large language models, and smaller fine-tuned encoder or encoder-decoder models. Experimental results are reported using tuple-level recall, tuple-level F1, component recall, relation-condition recovery metrics, accuracy, Macro F1, label-level F1, probe-type accuracy, Overgeneralization Rate (OGR), and Scope Preservation Score (SPS).

The experimental setup, model settings, prompting settings, training configuration, evaluation scripts, and relevant hyperparameters are documented in the appendix and accompanying code release. Where applicable, the release also records model identifiers, package versions, preprocessing choices, and evaluation thresholds. These details are provided to make the reported results reproducible and to clarify whether each result comes from a prompted, fine-tuned, transfer, or in-domain setting.

\subsection{Use of AI Assistants}
\label{app:ai_assistant_use}

AI assistants were used to support parts of the research workflow, including drafting, editing, code debugging, LaTeX/table formatting, and the generation of initial probe candidates. The AI-generated probe candidates were not accepted automatically. They were manually reviewed and filtered to remove malformed probes, wrong-component edits, weak alternate conditions, and cases where the intended condition perturbation was not properly expressed.

AI assistants were also used to help refine the manuscript language, appendix wording, and presentation of experimental results. 

\subsection{Potential Risks}
\label{app:potential_risks}

The main risk of this work is misuse or overinterpretation of condition-scoped relations. A model may correctly identify a relation but apply it outside the condition under which the source text supports it. This can produce unsupported or misleading conclusions, especially in scientific, biomedical, clinical, or public-discourse settings. CondRelBench is designed to expose this failure mode, not to certify models for deployment.

There is also a risk that benchmark users may treat the dataset as an exhaustive relation extraction resource. This would be inappropriate because the real-world PubMed and Reddit annotations are selective rather than exhaustive. The benchmark is intended for targeted probing of condition awareness, scope preservation, and overgeneralization behavior. Users should not treat missing tuples as proof that no valid relation exists in a source document.

\end{document}